\documentclass[10pt,twocolumn]{article}
\usepackage[margin=0.75in,bottom=0.5in,footskip=18pt,columnsep=0.25in,headheight=26pt,headsep=8pt]{geometry}
\usepackage[T1]{fontenc}
\usepackage{amsmath,amssymb,amsthm,mathtools}
\usepackage{mathpazo}           %
\usepackage{tgpagella,tgheros,microtype}
\usepackage{graphicx,booktabs,longtable,array}
\usepackage[numbers,sort&compress]{natbib}
\usepackage{xcolor}
\usepackage{tikz}
\usepackage{tcolorbox}
\usepackage{enumitem}
\tcbuselibrary{skins,breakable}
\usepackage{titlesec,fancyhdr}
\usepackage{titletoc}
\usepackage{tabularx}
\usepackage{array}
\usepackage{booktabs}
\titlecontents{section}[2.2em]{\addvspace{5pt}\sffamily\bfseries}{\contentslabel{2.2em}}{\hspace*{-2.2em}}{\titlerule*[0.7pc]{.}\contentspage}
\titlecontents{subsection}[4.6em]{\addvspace{1pt}\small}{\contentslabel{2.4em}}{\hspace*{-2.4em}}{\titlerule*[0.7pc]{.}\contentspage}
\usepackage[font=footnotesize,labelfont=bf,labelsep=period]{caption}
\makeatletter%
\newcommand\FloatBarrier{\par\begingroup\let\@elt\relax\edef\@tempa{\@botlist\@deferlist\@dbldeferlist}\ifx\@tempa\@empty\else\clearpage\fi\endgroup}
\makeatother
\usetikzlibrary{arrows.meta,calc,positioning,shapes.geometric}
\usepackage[colorlinks=true,linkcolor=blue!60!black,citecolor=blue!60!black,urlcolor=blue!60!black]{hyperref}
\usepackage[capitalize,noabbrev]{cleveref}

\definecolor{good}{HTML}{1F77B4}\definecolor{bad}{HTML}{D62728}\definecolor{found}{HTML}{9467BD}\definecolor{aligned}{HTML}{2CA02C}
\definecolor{rule}{HTML}{8EB9D9}

\newcommand{\selvlm}{at most 54.5}

\NewMarkClass{roadmap}%
\titleformat{\section}{\sffamily\bfseries\Large\raggedright}{\thesection}{0.6em}{\expanded{\noexpand\InsertMark{roadmap}{\arabic{section}}}}
\titleformat{\subsection}{\sffamily\bfseries\large\raggedright}{\thesubsection}{0.6em}{}
\titlespacing*{\section}{0pt}{10pt plus 3pt minus 2pt}{5pt plus 1pt}
\titlespacing*{\subsection}{0pt}{8pt plus 2pt minus 2pt}{4pt plus 1pt}
\newif\ifappendixpart
\newcommand{\roadmap}[3]{\hyperref[#3]{\ifnum0\LastMark{roadmap}=#1\textbf{Sec #1: #2}\else Sec #1: #2\fi}}
\newcommand{\brownlogo}[1]{\includegraphics[height=#1]{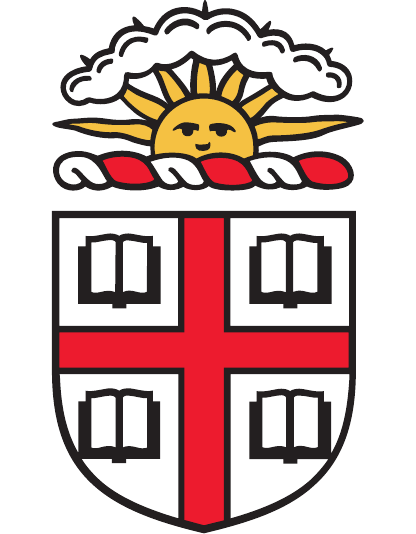}\hspace{.18em}\raisebox{.24\height}{\includegraphics[height=.42\dimexpr#1\relax]{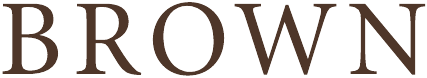}}}
\newcommand{\footlogos}{\brownlogo{9pt}\hspace{1.4em}\includegraphics[height=10.5pt]{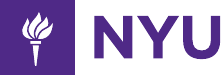}\hspace{1.4em}\includegraphics[height=8.6pt]{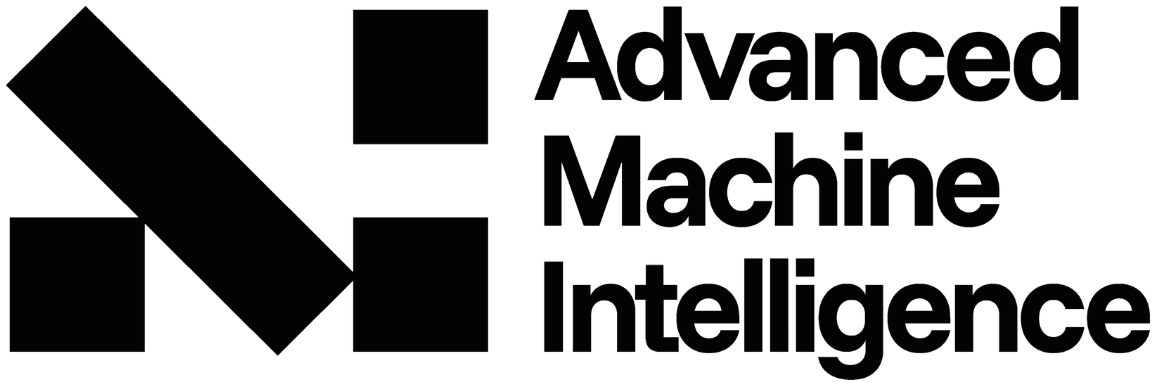}}
\fancypagestyle{plain}{\fancyhf{}\fancyfoot[L]{\footnotesize Correspondence: \href{mailto:rbalestr@brown.edu}{\texttt{rbalestr@brown.edu}}}\fancyfoot[R]{\small\thepage}}

\makeatletter%
\tcbset{colbreak/.code={\if@twocolumn\if@firstcolumn\pgfkeysalso{breakable}\fi\fi}}
\makeatother
\tcbset{result/.style={enhanced,colbreak,arc=2pt,boxrule=0.6pt,left=4pt,right=4pt,top=2pt,bottom=3pt,toptitle=1pt,bottomtitle=1pt,
  fonttitle=\bfseries,coltitle=white,before skip=6pt,after skip=6pt}}
\newcounter{theorem}\newcounter{proposition}
\NewDocumentEnvironment{theorem}{o}{\refstepcounter{theorem}%
  \begin{tcolorbox}[result,colframe=red!45!black,colback=red!2,title={Theorem~\thetheorem\IfValueT{#1}{: #1}}]}{\end{tcolorbox}}
\NewDocumentEnvironment{proposition}{o}{\refstepcounter{proposition}%
  \begin{tcolorbox}[result,colframe=blue!50!black,colback=blue!2,title={Proposition~\theproposition\IfValueT{#1}{: #1}}]}{\end{tcolorbox}}
\NewDocumentEnvironment{fullstatement}{m}{%
  \begin{tcolorbox}[result,colframe=black!55,colback=black!2,title={#1, full statement}]}{\end{tcolorbox}}

\newcommand{\R}{\mathbb{R}}
\newcommand{\E}{\mathbb{E}}
\newcommand{\Z}{\mathcal{Z}}
\newcommand{\X}{\mathcal{X}}
\newcommand{\Oset}{\mathrm{O}(k)}
\newcommand{\mQ}{\boldsymbol{Q}}\newcommand{\mK}{\boldsymbol{K}}\newcommand{\mC}{\boldsymbol{C}}\newcommand{\mI}{\boldsymbol{I}}\newcommand{\mU}{\boldsymbol{U}}\newcommand{\mM}{\boldsymbol{M}}\newcommand{\mY}{\boldsymbol{Y}}\newcommand{\mR}{\boldsymbol{R}}\newcommand{\mX}{\boldsymbol{X}}\newcommand{\mE}{\boldsymbol{E}}\newcommand{\mZ}{\boldsymbol{Z}}
\newcommand{\vPhi}{\boldsymbol{\Phi}}\newcommand{\vphi}{\boldsymbol{\phi}}\newcommand{\mF}{\boldsymbol{F}}
\newcommand{\projectlink}[3]{\raisebox{-.18\height}{\includegraphics[height=9.5pt]{#1}}\,\href{#2}{#3}}
\newcommand{\projectlinks}{\projectlink{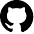}{https://github.com/galilai-group/platonic-reasoning}{Code}\hspace{1.6em}\projectlink{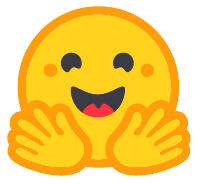}{https://huggingface.co/spaces/galilai-group/platonic-reasoning}{Demo}\hspace{1.6em}\projectlink{figs/icon_hf.pdf}{https://huggingface.co/galilai-group}{Models and data}}
\title{\bfseries Platonic Reasoning: Open-World Multimodal Reasoning\\Without Training on Paired Data\vspace{-0.5em}}
\author{Randall Balestriero$^{1,3}$\qquad Yann LeCun$^{2,3}$\\[2pt]
$^1$\,\raisebox{-.22\height}{\brownlogo{21pt}}\hspace{2.4em}$^2$\,\raisebox{-.22\height}{\includegraphics[height=19pt]{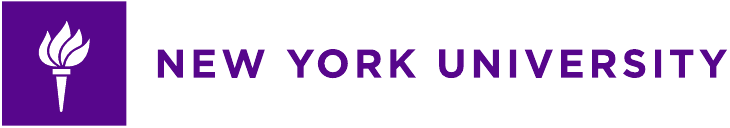}}\hspace{2.4em}$^3$\,\raisebox{-.2\height}{\includegraphics[height=18pt]{figs/logo_ami.png}}\\[5pt]
{\small\projectlinks}}
\date{}

\begin{document}
\twocolumn[{%
\vspace*{-3.8em}
\maketitle
\vspace{-3.05em}
\thispagestyle{plain}
\begin{tcolorbox}[enhanced,colback=rule!18,boxrule=0pt,frame hidden,sharp corners,left=8pt,right=8pt,top=3.5pt,bottom=3pt]
Science records the world in many modalities beyond language, from brain activity to physical simulations, and self-supervised encoders learn rich representations of them. Understanding what these encoders learn requires reasoning in language, but multimodal reasoning relies on large amounts of paired data that most modalities will never have. We introduce \emph{Platonic Reasoning}, which lets a language model reason about a modality without training any network on paired data. Platonic Reasoning needs four ingredients: a frozen encoder of the modality, the {\em anchor}, unpaired text about the modality, a generalist language model, and a few pairs. We prove that the anchor bounds how well any text encoder can align with it, and that every anchor admits a \emph{Platonic JEPA} that reaches this bound, trained on text alone with \emph{Platonic views}. \emph{Platonic View Search} (PVS) finds these views automatically: our 109M-parameter Platonic JEPA, trained on Wikipedia alone, aligns with DINOv2 better than language models of up to 284B parameters. One linear map aligns the anchor with the Platonic JEPA and enables retrieval from text. Two recipes turn this alignment into reasoning: a language model fine-tuned on text alone to decode Platonic JEPA embeddings, or in-context learning with no training. With only 10 pairs, Platonic Reasoning answers blind open questions about physical simulations far better than the latest open-weight vision--language models (VLMs), Qwen3.8 and Gemma~4, even when these VLMs are fine-tuned on the same pairs, and the gap grows when the anchor chooses which samples to label. Platonic Reasoning also leads on the dynamics of single runs and captions sleep EEG and simulations better than the best fine-tuned VLM, opening a path to reason in language about the modalities of science that will never be paired with text.

\par\vspace{2.5pt}
\centering
\definecolor{pj}{HTML}{9467BD}\definecolor{tr}{HTML}{E07B24}
\resizebox{\linewidth}{!}{%
\begin{tikzpicture}[x=1cm, y=1cm, font=\small,
  head/.style={fill=pj, text=white, rounded corners=3pt, inner sep=4pt, font=\sffamily\bfseries\large, anchor=north west},
  step/.style={font=\sffamily\bfseries, text=pj!85!black, anchor=north west},
  sub/.style={font=\sffamily\footnotesize, text=black!60, anchor=north west},
  enc/.style={draw=#1, fill=#1!13, rounded corners=4pt, line width=1pt, align=center, inner sep=3pt, minimum height=0.85cm, text width=2.3cm, font=\small},
  sc/.style={draw=#1, fill=#1!12, rounded corners=4pt, line width=1pt, align=center, inner sep=3pt, minimum height=0.9cm, text width=4.1cm, font=\small},
  view/.style={draw=black!30, fill=black!3, rounded corners=3pt, align=left, inner sep=3pt, text width=2.6cm, font=\scriptsize, anchor=north},
  arr/.style={-{Stealth[length=6pt,width=5pt]}, line width=1.1pt, black!55},
  lab/.style={font=\sffamily\scriptsize\bfseries, text=black!60, anchor=south west},
  lock/.pic={\fill[black!80, rounded corners=0.7pt] (-0.14,-0.13) rectangle (0.14,0.07); \draw[black!80, line width=1.1pt] (-0.085,0.07) -- (-0.085,0.14) arc(180:0:0.085) -- (0.085,0.07);}]
\draw[pj!45, line width=0.9pt, rounded corners=7pt] (0,0) rectangle (18,-7.5);
\node[head] at (0.2,-0.18) {Platonic JEPA Training};
\draw[black!25, dashed, line width=0.7pt] (12.7,-0.85) -- (12.7,-7.3);
\node[anchor=north west, font=\sffamily\bfseries\normalsize, text=pj!85!black] (st1) at (5.6,-0.3) {1\enspace Search};
\node[sub, anchor=west] at ([xshift=5pt]st1.east) {a few view samplers, one small JEPA each, scored on a few pairs};
\node[anchor=north, inner sep=0] (photo) at (1.7,-1.05) {\includegraphics[height=1.5cm]{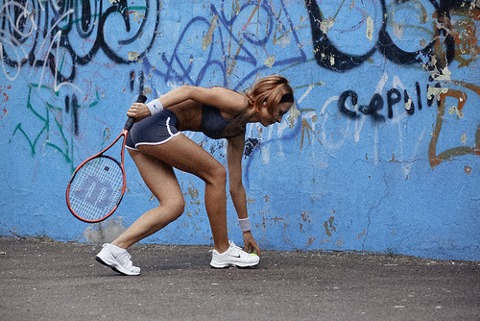}};
\node[enc=good] (g) at (1.7,-3.55) {\textbf{anchor} $g$\\[-1pt]{\scriptsize DINOv2, frozen}};
\pic at ([xshift=-1pt,yshift=-1pt]g.north east) {lock};
\node[anchor=north, inner sep=0] (ta) at (1.7,-4.3) {\includegraphics[height=2.0cm]{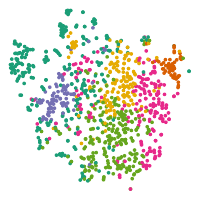}};
\node[font=\sffamily\scriptsize\bfseries, text=good] at (1.7,-6.65) {anchor embedding};
\draw[arr] (photo.south) -- (g.north); \draw[arr] (g.south) -- (ta.north);
\node[anchor=north west, draw=black!40, rounded corners=3pt, align=center, inner sep=4pt, text width=8.25cm, font=\footnotesize] (cap) at (3.7,-1.05) {its caption: ``A tennis player stops to pick up the ball.''};
\draw[black!20, line width=0.6pt] (3.35,-1.05) -- (3.35,-6.9);
\foreach \c/\x/\nm/\v/\r/\f in {1/5.0/bad views/{[MASK] [MASK] pick up the ball.}/.16/bad,
                                2/7.95/paraphrase/{A tennis player pauses to retrieve the ball.}/.22/paraphrase,
                                3/10.9/imagine/{A tennis player in a bright yellow shirt\dots}/.34/imagine} {
  \node[view] (v\c) at (\x,-1.95) {\textbf{\nm}: \v};
  \draw[arr] (cap.south -| \x,0) -- (v\c.north);
  \node[enc=found, text width=1.8cm] (j\c) at (\x,-3.55) {\textbf{JEPA}\\[-1pt]{\scriptsize\textcolor{tr}{trainable}}};
  \draw[arr] (v\c.south) -- (j\c.north);
  \node[anchor=north, inner sep=0] (t\c) at (\x,-4.3) {\includegraphics[height=2.0cm]{figs/teaser/tsne_\f.pdf}};
  \draw[arr] (j\c.south) -- (t\c.north);
  \node[font=\sffamily\scriptsize\bfseries] at (\x,-6.65) {reward \r};
}
\draw[pj, line width=1.6pt, rounded corners=6pt] (9.45,-1.7) rectangle (12.35,-6.95);
\node[font=\sffamily\scriptsize\bfseries, text=pj] (bk) at (10.75,-7.2) {best: kept};
\pic at ([xshift=7pt,yshift=-1pt]bk.east) {lock};
\node[anchor=west, font=\scriptsize, text=black!60] at (3.5,-7.2) {\textcolor{pj}{$\bullet$}\,photo found by its caption through one rotation};
\node[anchor=north west, font=\sffamily\bfseries\normalsize, text=pj!85!black] at (12.95,-0.8) {2\enspace Scale};
\node[sub, text width=4.8cm, align=left] at (12.95,-1.3) {the best sampler, more data, a bigger model, then one rotation};
\node[sc=found] (s1) at (15.3,-2.75) {\textbf{imagine} sampler\\[-1pt]{\scriptsize every text of the corpus}};
\pic at ([xshift=-1pt,yshift=-1pt]s1.north east) {lock};
\node[sc=pj] (s2) at (15.3,-4.15) {\textbf{Platonic JEPA}\\[-1pt]{\scriptsize one modality only, at scale}};
\node[sc=aligned] (s3) at (15.3,-5.55) {\textbf{one rotation} $\hat{\boldsymbol Q}$\\[-1pt]{\scriptsize fitted on $n$ pairs, no training}};
\node[sc=good] (s4) at (15.3,-6.9) {\textbf{aligned with} $g$};
\draw[arr] (s1) -- (s2); \draw[arr] (s2) -- (s3); \draw[arr] (s3) -- (s4);
\begin{scope}[yshift=0.3cm]
\node[head] at (0.2,-7.95) {Deploy};
\node[sub] at (2.0,-8.1) {read modalities with little paired text (examples: one map fitted on 10 pairs; retrieval: 100)};
\node[lab] at (0.25,-9.05) {CAPTIONING};
\foreach \x in {3.18, 6.22} \draw[black!25, dashed, line width=0.6pt] (\x,-9.35) -- (\x,-12.4);
\tikzset{dcap/.style={anchor=north, text width=2.85cm, align=left, font={\footnotesize\hyphenpenalty=10000\exhyphenpenalty=10000}}}
\foreach \x/\f in {1.65/sim, 4.7/eeg, 7.75/wrist10}
  \node[anchor=north, inner sep=0] at (\x,-9.35) {\includegraphics[height=1.85cm,width=2.85cm,keepaspectratio]{figs/teaser/\f.png}};
\node[dcap] at (1.65,-11.3) {\textcolor{pj}{\textbf{Simulation}: Rayleigh-B\'enard convection, Ra $=10^{10}$, Pr $=2$.}\,\textcolor{aligned}{\checkmark}};
\node[dcap] at (4.7,-11.3) {\textcolor{pj}{\textbf{Sleep EEG}: Stage N3 sleep.}\,\textcolor{aligned}{\checkmark}};
\node[dcap] at (7.75,-11.3) {\textcolor{pj}{\textbf{Wrist motion}: walking.}\,\textcolor{aligned}{\checkmark}};
\draw[black!20, line width=0.6pt] (9.25,-9.05) -- (9.25,-12.55);
\node[lab] at (9.4,-9.05) {RETRIEVAL};
\node[anchor=north west, text width=2.55cm, align=left, font={\footnotesize\hyphenpenalty=10000}, draw=pj, rounded corners=3pt, line width=0.8pt, inner sep=3pt] at (9.45,-9.35) {\textcolor{pj}{\textbf{query} \textcolor{black!55}{(1{,}612 calls)}: ``The sound of a Lyric Cicada (\emph{Neotibicen lyricen}).''}};
\foreach \j/\x in {1/10.15, 2/11.55} {
  \node[anchor=north, inner sep=0] at (\x,-11.0) {\includegraphics[width=1.25cm,height=1.2cm]{figs/teaser/call\j.png}};
  \node[anchor=north, font=\footnotesize] at (\x,-12.23) {\textcolor{aligned}{\checkmark}\,top-\j};
}
\draw[black!20, line width=0.6pt] (12.3,-9.05) -- (12.3,-12.55);
\node[lab] at (12.45,-9.05) {VS.\ FINE-TUNED VLMs, BASELINES};
\node[anchor=north west, inner sep=0] at (12.45,-9.3) {\includegraphics{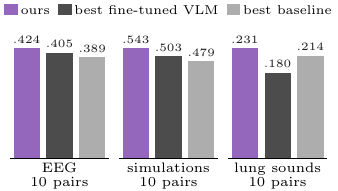}};
\node[anchor=north, font=\footnotesize\itshape, text=black!60] at (9.0,-12.65) {one Platonic JEPA per domain, the same recipe on every domain; VLMs fine-tuned on the same pairs; means over 3 draws of the pairs (1 for Qwen3.8 on EEG)};
\end{scope}
\end{tikzpicture}}
\captionsetup{hypcap=false}\captionof{figure}{\textbf{Platonic JEPA}, the core of Platonic Reasoning: search views, scale the best, align by one map. The views are written from the caption alone, so the imagined view's unstated details, e.g., colors, come from the language model, not from the photo (\cref{sec:flickr}).}
\label{fig:teaser}
\end{tcolorbox}
}]

\clearpage
\section{Introduction}\label{sec:intro}

Science now records far more data than anyone can parse, e.g., simulations of physical systems, electroencephalography (EEG) or wearable sensors. Self-supervised encoders and world models learn from such data at scale. Their representations already capture much of the dynamics of those systems \citep{oquab2024dinov2,mccabe2025walrus,wang2025cbramod}. Yet, this knowledge only becomes useful once people can query it, e.g., ask what a simulation shows or why a recording looks abnormal. People reason in language, and most of what is known about a modality is written in text. Language is thus the natural interface to what an encoder has learned. Connecting an encoder to language, however, requires multimodal models trained on paired data at web scale \citep{radford2021clip}. Most scientific modalities only have hundreds of pairs with text, far too few to train such a model. These modalities do have strong self-supervised encoders and abundant unpaired data. We call the modality that has the frozen encoder the {\em source} modality, e.g., physical simulations, EEG or photographs. We call the modality we align it with the {\em target} modality, which is text throughout this paper. We ask the following question: {\em can a language model reason about a source modality without training on paired data?}

Answering this question first requires aligning the source and target modalities. The Platonic Representation Hypothesis (PRH) suggests that such an alignment may emerge without pairs, because models trained on different modalities converge towards a shared representation of the world \citep{huh2024platonic}. This convergence was recently proven and improved for Joint-Embedding Predictive Architectures (JEPAs) \citep{lecun2022path} trained on each modality alone \citep{balestriero2026prh}. The convergence requires good positive views on both sides. Positive views are pairs of inputs known to share their content, and they alone determine what a JEPA learns \citep{balestriero2022contrastive,haochen2021provable}. Good views are only known for images, where years of practice converged to crops and color changes \citep{oquab2024dinov2,balestriero2025lejepa}. This practice does not transfer to text. Masking or cropping a sentence keeps its wording, which is exactly the nuisance a JEPA should ignore. Latent prediction of masked text was also found to fail \citep{dinh2026paradox}, unless the texts come naturally paired, as in LLM-JEPA \citep{huang2025llmjepa}. For EEG or physical simulations, no good views are known at all. We ask a second question: {\em can the PRH be exploited when the views are unknown, and does the resulting alignment suffice to reason about the source modality?}

\input{incontext_fig}

We answer this question with an \emph{anchor}, a frozen encoder of the source modality trained in any way. First, we prove that the anchor sets a ceiling on how well any encoder of the target modality can align with it, for any method and any number of pairs (\cref{thm:ceiling}). Second, every anchor admits \emph{Platonic views}, and the JEPA trained on them, the \emph{Platonic JEPA}, reaches the ceiling (\cref{thm:every}). Third, the alignment of a view's JEPA is maximal exactly at Platonic views (\cref{thm:search}). Moreover, with enough dimensions, a Platonic JEPA followed by one linear map predicts the anchor as well as any encoder of the target modality, however it is trained and whatever nonlinear head it uses. Hence, {\bf positive views can be searched for whenever a frozen encoder of the source modality is available}. \emph{Platonic View Search} (PVS) does so with a language model that rewrites each text with a few generic prompts, e.g., ``imagine the scene behind this text and describe it anew''. PVS trains one small JEPA per prompt and keeps the best aligned one (\cref{fig:teaser}). On Flickr30k, PVS finds that imagined views are better positives than the other human captions of the same photograph (\cref{sec:flickr}). On Wikipedia, PVS trains text JEPAs that align with DINOv2 better than every language model we tested (\cref{sec:scale}).

\emph{Platonic Reasoning} uses this alignment to reason (\cref{fig:incontext}). Platonic Reasoning needs four ingredients: the anchor, a Platonic JEPA trained on unpaired text of the domain, a generalist language model, and 10 pairs that fit one linear map from the anchor to the Platonic JEPA. The anchor, unpaired text and a generalist language model already exist for most sciences. A few hundred held-out pairs also choose the views and settings, but no network trains on them (\cref{sec:domains}). We give two recipes. The first fine-tunes an open language model on text alone to decode noisy Platonic JEPA embeddings. The second retrieves the corpus descriptions nearest to the mapped input and puts them in the prompt of any language model, with no training; we test three open-weight models. With 10 pairs, Platonic Reasoning answers blind open questions about physical simulations far better than the latest open-weight vision--language models (VLMs), Qwen3.8-27B \citep{qwen2026qwen38} and Gemma-4-31B \citep{gemma2026gemma4}, even when these VLMs are fine-tuned on the same pairs (66.2 against 51.3; \cref{tab:openqa}). The gap remains with more pairs and against the newest open VLM, Qwen3.8-Flash-Next (\cref{tab:openqa_detail}). Two checks show that the in-context recipe uses the input. First, the top retrieved description names the right physical system for most simulations, far more often than chance (\cref{tab:factcheck}). Second, replacing the retrieved descriptions with five random ones drops the score from 66.2 to 36.4. Platonic Reasoning is also ahead on open questions about the dynamics of single runs, and narrowly on label questions about sleep EEG (\cref{tab:openqa,tab:labelqa}). Models that already read a modality, e.g., photographs or sounds, stay ahead on it. Captioning by retrieval is also much cheaper than these VLMs, and fast on a CPU for most modalities (\cref{tab:cost}). We summarize our contributions below.
\begin{enumerate}[leftmargin=*,itemsep=1pt,topsep=2pt]
\item {\bf A Platonic Representation Theorem for imperfect encoders and sampling.}~A frozen encoder of the source modality bounds how well any encoder of the target modality can align with it, even when neither encoder captures the whole world. JEPAs trained on Platonic views reach this bound, and every encoder admits such views. For these JEPAs, and only for them, one linear map predicts the anchor as well as any function can. A rotation suffices only when the anchor carries no nuisance (\cref{thm:ceiling,thm:every,thm:search,prop:closed}).
\item {\bf Platonic JEPAs via Platonic View Search.}~We find positive views automatically. We score each candidate view generator by the alignment of the JEPA it trains. This score is maximal at Platonic views (\cref{thm:search}). PVS needs a few hundred pairs, and no network is trained on them (\cref{sec:pvs}).
\item {\bf Text-only JEPAs that align with images better than language models.}~Platonic JEPAs trained on text alone align with DINOv2 better than JEPAs trained on human captions of the same photographs. A 109M-parameter Platonic JEPA trained on Wikipedia aligns better than every language model we tested, up to 284B parameters (\cref{sec:flickr,sec:scale}).
\item {\bf Platonic Reasoning for science.}~With only 10 pairs, a decoder trained on text alone, or a language model in context, can read a Platonic JEPA. Both readers answer short and open questions about a modality, and the same map retrieves and captions it. On simulations and sleep EEG, Platonic Reasoning beats the latest open-weight VLMs fine-tuned on the same pairs and a model that never sees the input, and captioning by retrieval costs a fraction of theirs. We also give a guide on which method to use, and when to fine-tune a VLM instead (\cref{tab:guide}; \cref{sec:exp,sec:reasoning,sec:phase}).
\end{enumerate}

\section{Every Anchor Has Platonic Views}\label{sec:theory}

This section is modality-agnostic and considers a frozen encoder of a source modality, the anchor, and a JEPA trained on a target modality alone, e.g., text in the rest of the paper. We first bound how well any encoder of the target modality can align with the anchor (\cref{sec:ceiling}). We then show that views reaching this bound exist for every anchor (\cref{sec:exist}). Finally, we show that the search's reward is maximal exactly at these views. The reward is the alignment of the JEPA that a view law trains. Both this reward and the linear map that aligns the two encoders can be estimated from pairs (\cref{sec:find}).

\subsection{The Anchor Sets the Ceiling}\label{sec:ceiling}

\begin{figure*}[t]
\centering
\includegraphics{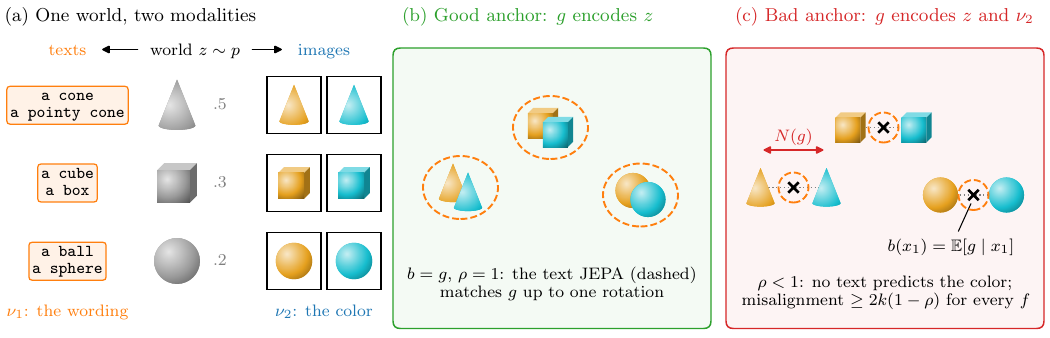}
\caption{\textbf{The anchor sets the ceiling (\cref{thm:ceiling}).} (a) Three states $z\sim p$: a cone, a cube and a sphere. Texts render them with a wording $\nu_1$, images with a color $\nu_2$. (b) A good anchor $g$ encodes the state alone: a text predicts all of it ($b=g$, $\rho=1$), and a text JEPA matches it up to one rotation. (c) A bad anchor also encodes the color, which these texts never mention. A text encoder can at best reach the predictable part $b(x_1)=\E[g(x_2)\mid x_1]$ (\texttimes). The color split, the anchor's nuisance $N(g)$, is a misalignment that no method and no number of pairs can remove.}
\label{fig:ceiling}
\end{figure*}

\paragraph{Setting.} Two modalities sense one world. A \emph{paired observation} $(x_1,x_2)$ follows a joint law whose marginals are the laws of each modality's unpaired data. The \emph{anchor} $g:\X_2\to\R^d$ is a frozen encoder of the source modality, indexed by 2, trained in any way. The target modality, indexed by 1, has encoders $f:\X_1\to\R^k$. All encoders are centered and whitened. A \emph{rotation} is a matrix with orthonormal rows or columns, reflections included, e.g., $\mQ\in\mathrm O(k,d)$. The \emph{misalignment} of $f$ with $g$ under $\mQ$ is $\E\|f(x_1)-\mQ g(x_2)\|^2$ over paired observations. The \emph{predictable part} of the anchor is $b(x_1)=\E[g(x_2)\mid x_1]$, i.e., what the target modality can know of the anchor. The covariance of $b$, $\mC=\mU\boldsymbol\Lambda\mU^\top$, has rank $s$ and eigenvalues $\Lambda_{11}\ge\dots\ge\Lambda_{ss}>0$. The whitened directions of $b$ are $h=\boldsymbol\Lambda^{-1/2}\mU^\top b\in\R^s$. These directions are ordered by $\Lambda_{ii}^{1/2}$, the largest correlation that any function of $x_1$ reaches with the anchor's $i$-th predictable direction. We write $\rho=\frac1k\sum_{i\le r}\Lambda_{ii}^{1/2}\in[0,1]$, with $r=\min(k,s)$, for the mean of the $k$ largest of these correlations, i.e., $\mathrm{tr}\,\mC^{1/2}/k$ if $k\ge s$. When eigenvalues tie, $h$ is unique only up to rotations within them, and a span \emph{contains} $h_{1:r}$ if it does for some such choice. The top $k$ directions are \emph{unique} if $k=s$ or $\Lambda_{kk}>\Lambda_{k+1,k+1}$. A JEPA of the target modality trained with second-order anti-collapse on a \emph{view law} $W$, i.e., a law of positive pairs, learns the spectral embedding $f_W$ of $W$, namely the top $k$ non-constant eigenfunctions of its Markov operator \citep{balestriero2022contrastive,haochen2021provable}.

\paragraph{World model.} For JEPAs whose positives are two observations of one scene, cross-modal alignment is a theorem \citep{balestriero2026prh}. We use its world model to interpret our results. A finite world has states $z\in\Z$, drawn as $z\sim p$. Modality $m$ renders a state injectively as $x_m=\gamma_m(z,\nu_m)$, with a nuisance $\nu_m\sim q_m$ of its own, e.g., the words of a caption or the viewpoint of a photograph. The variables $z$, $\nu_1$ and $\nu_2$ are independent. The split between world and nuisance is therefore not fixed in advance: the state holds what the two modalities share, and each nuisance what only its own modality sees. A color or viewpoint that captions name, or a watermark whose stock-photo site they name, is thus part of the world. An encoder's \emph{world part} is $a_f(z)=\E[f(x)\mid z]$, with covariance $\mC_f$. Its \emph{nuisance energy} is $N(f)=\E\|f(x)-a_f(z)\|^2=\mathrm{tr}(\mI-\mC_f)$. \emph{Good views} of a world kernel $\pi$, a joint law of two states, pair renderings of two states $(z,z')\sim\pi$ with fresh nuisances. The JEPA of good views learns $\vPhi_\pi(z)$, the top non-constant eigenfunctions of $\pi$'s Markov operator on functions of the state \citep[Prop.~2]{balestriero2026prh}.

\paragraph{Assumptions.} \textbf{(A1)} $x_1$ is finite, with one law $P$ for its unpaired data, positive pairs and paired observations. \textbf{(A2)} View laws $W$ are symmetric with marginal $P$. \textbf{(A3)} Their Markov operators have a gap after their $k$-th non-constant eigenvalue, $\mu_k>\max(\mu_{k+1},0)$. \textbf{(A4)} A JEPA is the exact population minimizer: $f_W$ whitens the top-$k$ eigenspace, unique up to rotation. \textbf{(A5)} Encoders are whitened in population, which in practice means on abundant unpaired data. The encoders are independent of the $n$ i.i.d.\ pairs. \Cref{thm:every} builds its views from $b$, i.e., from pairs, so it proves that these views exist but does not give a sampler. Each result below is restated in full, with every condition, next to its proof in \cref{app:proofs}.

\begin{theorem}[The Anchor Sets the Ceiling]\label{thm:ceiling}
Assume (A5). (i) Every whitened encoder $f$ of the target modality, however it is trained and on however many pairs, satisfies
\[
\max_{\mQ\in\mathrm O(k,d)}\E\langle f(x_1),\mQ g(x_2)\rangle\le k\rho,
\]
with equality if and only if the span of $f$ contains the top directions $h_{1:r}$; for $k=d$, the misalignment is thus at least $2k(1-\rho)$. If $k\le s$ and the top $k$ directions are unique, these encoders are exactly the rotations of $h_{1:k}$.\par
(ii) In the world model, $b=a_g$, and the misalignment splits into the two nuisance energies and the mismatch of the world parts,
\begin{equation}\label{eq:split}
\begin{split}
\E\|f(x_1)-\mQ g(x_2)\|^2={}&N(f)+N(\mQ g)\\
&+\E\|a_f(z)-\mQ a_g(z)\|^2,
\end{split}
\end{equation}
so it vanishes if and only if $f(x_1)=\mQ g(x_2)=\mQ a_g(z)$, i.e., $f$ and $\mQ g$ encode the same function of the state, and nothing else. (Proof in \cref{prf:ceiling}.)
\end{theorem}

In short, the anchor, not the method, sets how well the target modality can be aligned with it (\cref{fig:ceiling}). No number of pairs and no supervision recovers the part of the anchor that the target modality cannot predict. Part (i) is classical: it sums canonical correlations \citep{hotelling1936relations} through orthogonal Procrustes \citep{schonemann1966generalized} and bounds them by those of all functions of $x_1$. We restate it to name the ceiling. When two modalities sense one world, the unpredictable part is exactly the anchor's nuisance, since by definition the other modality does not see that nuisance. A zero misalignment thus certifies that neither encoder spends a direction on nuisance. This certificate is the representation analogue of co-training's \emph{agreement implies correctness} \citep{blum1998cotraining,dasgupta2001pac} and of multi-view redundancy \citep{tosh2021contrastive}. The anchor need not be good. For example, DINOv2's positives are crops of one photograph, i.e., bad views. Yet the encoders that align best with DINOv2 span its top whitened world directions, and nothing else if $k\le s$. When the target modality only sees part of the world, e.g., when a caption describes part of a photograph, (ii) no longer applies, and the ceiling drops through $b$.

\subsection{Aligning a JEPA to Any Anchor via Platonic Views}\label{sec:exist}

We now ask how a JEPA of the target modality can reach this ceiling without training on pairs. The Platonic Representation Theorem obtains alignment from good views on both sides, but good views are unknown for most modalities. Moreover, most of our anchors are not JEPAs but reconstruct masked inputs, predict the next step of a simulation, or solve pretext tasks (\cref{sec:domains}). We show next that the anchor itself defines the views we need.

We call any JEPA of the target modality that reaches the ceiling of \cref{thm:ceiling} a \emph{Platonic JEPA}. We call any view law whose JEPA is a Platonic JEPA \emph{Platonic views} of $g$. Our construction pairs two inputs of the target modality more often when the anchor's predictable part makes them more alike. The construction uses the kernel $\omega_g(x,x')=\lambda\,h_{1:r}(x)^\top h_{1:r}(x')+\lambda'e(x)^\top e(x')$, where $e$ completes $h_{1:r}$ with $k-r$ centered, whitened functions of $x_1$ orthogonal to $h$ (none if $k\le s$). The weights $\lambda,\lambda'>0$ satisfy $\lambda\max\|h_{1:r}\|^2+\lambda'\max\|e\|^2\le1$, which keeps the view law $W_g$ of \cref{thm:every} non-negative.

\begin{theorem}[Every Anchor Has Platonic Views]\label{thm:every}
Assume (A1)--(A5) and $|\X_1|>k$. (i) The view law $W_g(x,x')=P(x)P(x')\big(1+\omega_g(x,x')\big)$ trains a JEPA that reaches the ceiling $k\rho$, i.e., every anchor has Platonic views.\par
(ii) If $k\le s$ and the top $k$ directions are unique, every Platonic JEPA is a rotation of $h_{1:k}$, so that all Platonic views train the same JEPA.\par
(iii) In the world model with more than $k$ states, $W_g$ can be chosen to be an imagination (\cref{prop:imagine}), and if the anchor is a JEPA trained on good views of a world kernel $\pi$, good views of $\pi$ in the target modality are Platonic. (Proof in \cref{prf:every}.)
\end{theorem}

Hence, every frozen encoder, JEPA or not, defines views of the target modality. The JEPA of these views learns all that the target modality can know of the encoder, up to its width $k$, and nothing else when $k\le s$. When the anchor is a good-view JEPA, good views of its world kernel are Platonic, as in the Platonic Representation Theorem. For other anchors, good views of $\pi$ reach the ceiling only if $\vPhi_\pi$ spans $h_{1:r}$. Good views of $\pi$ do so for the bad-view JEPA of \cref{fig:anchors}, whose world part lies in the top directions of $\pi$, but not for the three other anchors. $W_g$ needs $b$, i.e., pairs, and its gap $\min(\lambda,\lambda')$ is at most $1/\max\|h_{1:r}\|^2\le1/r$. However, any view law whose JEPA spans $h_{1:r}$ is as good. The construction amounts to the views of InfoMin, which share only what a task needs \citep{tian2020infomin}, with the anchor as the task. Not every imagination is Platonic, only one whose kernel's top eigenfunctions span $h_{1:r}$. A product corruption, which corrupts the state and the nuisance independently, is Platonic if the anchor's world part spans its top $s\le k$ state directions and the corruption keeps less of the nuisance than of the weakest of these directions (\cref{prf:nuisance}).

\begin{proposition}[Imagining Gives Good Views]\label{prop:imagine}
Assume (A1)--(A5) in the world model. An \emph{imagination} draws a new state $z'\sim\kappa(\cdot\mid z)$ from an observation of $z$, ignoring its nuisance, and renders $z'$ afresh. If $\pi_\kappa(z,z')=p(z)\kappa(z'\mid z)$ is symmetric, an imagination gives good views of $\pi_\kappa$, whose JEPA learns a function of the state alone. A \emph{corruption}, e.g., a crop, edits the observation instead. A corruption gives \emph{bad views}, whose JEPA spends a direction on nuisance, once it keeps more of the nuisance than of the state's $k$-th direction. (Proof in \cref{prf:imagine}.)
\end{proposition}

An imagination is thus a world model used as a view sampler, while a corruption only edits its input. An imagination does not recover the observed state: it draws a new one, and its JEPA becomes invariant to whatever $\kappa$ resamples. For text, we ask a language model to imagine the scene behind a caption and describe it anew. The details that the caption leaves open, e.g., colors or the background, are then drawn from the often stereotypical prior of the language model, not from the photograph (\cref{fig:teaser}). The predictable part $b(x_1)=\E[g(x_2)\mid x_1]$ averages over these details, so a JEPA loses nothing of $b$ by ignoring them. Resampling an attribute that the caption states and the anchor encodes would instead lower the alignment. The language model is an imagination only approximately, because it reads the wording and writes in its own style. A model asked to paraphrase keeps more of the text. The reward tests which of the two is Platonic for a given anchor. The spread $\kappa$ of the imagination sets the world kernel. An exact posterior, $\kappa(z'\mid z)=\mathbf 1[z'=z]$, carries no nuisance but has no gap, so the JEPA is free among functions of the state.

\subsection{How to Search for Platonic Views}\label{sec:find}

We score a view law of the target modality by the alignment of the JEPA it trains,
\begin{equation}\label{eq:reward}
\begin{split}
R(W)&=\max_{\mQ\in\mathrm O(k,d)}\E\big\langle f_W(x_1),\mQ g(x_2)\big\rangle\\
&=\|\mM_W\|_*,\quad\mM_W=\E[f_W(x_1)g(x_2)^\top],
\end{split}
\end{equation}
so that the smallest misalignment of $f_W$ with $g$ is $k+\min(k,d)-2R(W)$. On $n$ i.i.d.\ pairs, $\hat\mM_W$ is the sample mean, $\hat R_n(W)=\|\hat\mM_W\|_*$, and $V_W=n\,\E\|\hat\mM_W-\mM_W\|_F^2=\E[\|f_W\|^2\|g\|^2]-\|\mM_W\|_F^2$ is a fourth moment of the encoders' outputs; for a Platonic JEPA $f^\star$ we write $\mM$ and $V$.

\begin{theorem}[Platonic Views Maximize the Reward, and One Linear Map Aligns]\label{thm:search}
Assume (A1)--(A5). (i) $R(W)\le k\rho$ for every view law $W$, with equality if and only if the span of $f_W$ contains $h_{1:r}$, i.e., $f_W$ is a Platonic JEPA.\par
(ii) For a Platonic JEPA $f^\star$, the least-squares linear map to the anchor is $\mM^\top=(\mM^\top\mM)^{1/2}\mQ^{\star\top}$, i.e., a rotation $\mQ^\star$ that maximizes the alignment followed by a scaling, and if $k\ge s$ it predicts the anchor as well as any function of $x_1$ can, since $\mM^\top f^\star(x_1)=\E[g(x_2)\mid x_1]$. Conversely, the least-squares linear map of a whitened encoder does so only if its span contains $h_{1:s}$, and, if $k=d$, a rotation does so if and only if, in addition, $g$ is a function of $x_1$, i.e., $\mC=\mI$.\par
(iii) If $k=d=s$, the rotation $\hat\mQ$ that maximizes the alignment on the $n$ pairs satisfies $\E\|\hat\mQ-\mQ^\star\|_F\le2\sqrt{V/(n\Lambda_{kk})}$.\par
(iv) The reward estimated on $n$ pairs has $\E\big(\hat R_n(W)-R(W)\big)^2\le\bar kV_W/n$, with $\bar k=\min(k,d)$; hence, if one of the candidates $W_1,\dots,W_m$ is Platonic and every non-Platonic one trails it by at least $\Delta>0$, the maximizer of $\hat R_n$ is Platonic with probability at least $1-4\bar k\sum_iV_{W_i}/(n\Delta^2)$. (Proof in \cref{prf:search}.)
\end{theorem}

By \cref{thm:ceiling}, the reward is thus maximal exactly at the Platonic JEPAs. The guarantees hold for the population reward under (A1)--(A5). In practice, we compute a held-out, rank-based proxy (\cref{sec:impl}). If $k\ge s$, the Platonic JEPA followed by $\mM^\top$, a rotation scaled by $(\mM^\top\mM)^{1/2}$, predicts the anchor as well as any encoder of the target modality could. Aligning a modality with an anchor thus reduces to finding Platonic views and one linear map. The remaining error, $\E\|g-\E[g\mid x_1]\|^2=\mathrm{tr}(\mI-\mC)$ ($=N(g)$ in the world model), belongs to the anchor. For $k=d$, a rotation alone adds $\mathrm{tr}\big((\mI-\mC^{1/2})^2\big)$ to this error, which gives the floor $2k(1-\rho)$. This extra cost vanishes only when the anchor is a function of $x_1$, i.e., $\mC=\mI$, which is exactly the regime of the Platonic Representation Theorem, where neither encoder keeps nuisance. An imperfect anchor needs one linear map, and no more. Part (iii) is the polar-factor bound of \citet{li1995polar} with a weaker constant, and (iv) is Chebyshev's inequality with a union bound. Both certify an $n^{-1/2}$ rate, but with worst-case constants ($V_W$ is of order $kd$). These bounds do not predict that a few hundred pairs suffice to rank our candidates in practice (\cref{fig:synthetic}). Nuisance kept by a view sampler costs reward exactly when it pushes one of the anchor's top directions out of the JEPA's top $k$ (\cref{prf:nuisance}), as \cref{fig:synthetic} validates.

The search matters only when a JEPA chooses which $k$ functions of $x_1$ to keep, i.e., if $|\X_1|>k+1$. Here we count distinct strings, so a label written in $t$ templates counts as $t$ values.

\begin{proposition}[A Closed Vocabulary Leaves Only the Rotation]\label{prop:closed}
Assume (A1) and (A5), and let $x_1$ take at most $k+1$ values. Every encoder whose values are affinely independent, whitened on their span, reaches the ceiling, and all such encoders coincide up to a rotation; with exactly $k+1$ values, these are all whitened encoders, so that every view law has the same reward. Distinct values do not suffice, as three collinear values span a single dimension. (Proof in \cref{prf:closed}.)
\end{proposition}

In short, on a closed vocabulary, every encoder with affinely independent values is Platonic, for the law on which it is whitened and the pairs are drawn. Views thus have nothing to find, and only the rotation is left to learn. On open text, $x_1$ takes more values than any encoder has dimensions, and the encoder's views decide what it keeps (\cref{thm:every,thm:search}). \Cref{sec:phase} measures these two regimes.

\section{Platonic View Search on Natural Images and Text}\label{sec:pvs}

We first give the algorithm for any source and target modalities (\cref{sec:impl}). We then validate PVS on photographs and their human captions, where good views are known \citep{balestriero2026prh} (\cref{sec:flickr}). Finally, we scale PVS to Wikipedia, where good views are unknown, and build a captioner on the resulting JEPA (\cref{sec:scale}).

\subsection{The Algorithm}\label{sec:impl}

PVS needs three ingredients. The first is the \emph{anchor}, a frozen encoder of the source modality, JEPA or not (\cref{thm:every}). We use DINOv2 ViT-B \citep{oquab2024dinov2} for images, and a self-supervised encoder trained without text for every other source modality (\cref{app:domains}). The second is a set of candidate view samplers. For text, a language model (Qwen2.5-7B-Instruct \citep{qwen2024qwen25}) rewrites each text on its own with one of five generic prompts: imagine (another observer describes the same item), paraphrase, literal, report and caption (\texttt{platonic/generate.py}). The prompts are the same in every domain, except for the modality's name. Spans of one text, the bad views of \citet{balestriero2026prh}, serve as a reference. The third ingredient is a few hundred validation pairs, held out from every test. No network is trained on them.

\begin{figure}[t]
\centering
\includegraphics{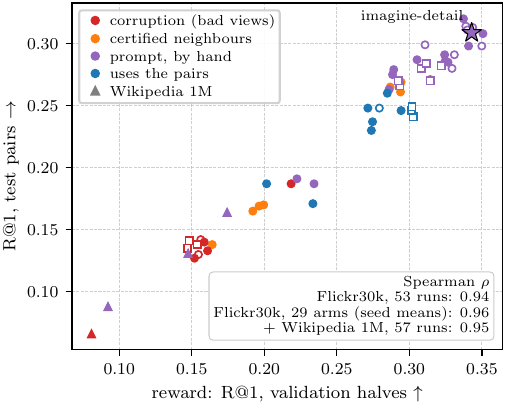}
\caption{\textbf{The reward predicts the test.} Reward on the Flickr30k validation halves against test R@1, for every candidate view generator of the search, on Flickr30k and on Wikipedia. Collapsed seeds are excluded (\texttt{paper/analysis.py search}). The legend gives the Spearman correlation of the reward with the test. We observe that \emph{a few hundred validation pairs rank views as the test pairs do}.}
\label{fig:reward_vs_test}
\end{figure}

\paragraph{The search.} For each candidate, the language model rewrites the \emph{corpus}, i.e., the unpaired texts of the target modality, in one pass, and we train a small JEPA on the rewrites. Every JEPA uses the same recipe, the BERT-S LeJEPA of \citet{balestriero2026prh}, so that only the positives change. We score a candidate by retrieval. We fit a rotation from the anchor to the JEPA on half of the validation pairs, and measure the R@1 from the source modality to the distinct texts on the other half. This reward is a held-out, rank-based proxy for the alignment $R$ of \cref{eq:reward}, and it ranks views as the test pairs do (\cref{fig:reward_vs_test}). We keep the best candidate (\cref{fig:search} lists every candidate), rewrite the whole corpus with it, and train a JEPA on these views, the \emph{Platonic JEPA}, though nothing certifies that it reaches the ceiling.

\paragraph{Reading the source modality.} One linear map, which we call \emph{the map}, is fitted on the $n$ available pairs to align the anchor with the Platonic JEPA. A rotation is the right map only when the anchor carries no nuisance (\cref{thm:search}(ii)). Otherwise, the least-squares map aligns better. The least-squares map is a rotation followed by a shrinkage of each direction. With fewer pairs than dimensions, we regularize it with a ridge penalty, which gives the \emph{ridge map}. With few pairs, the map can also be fitted on the unpaired samples of both modalities, anchored on the pairs, which gives the \emph{anchored alignment} \citep{schnaus2026rosetta}. We choose the map for each $n$ on validation pairs. An input of the source modality, encoded by the anchor and sent through the map, is a \emph{mapped input}, which is read in one of three ways. \emph{Retrieval} returns the corpus texts nearest to a mapped input, the first of which serves as its caption, or the mapped inputs nearest to a text query. The \emph{decoder} is a language model fine-tuned on text alone to write the text of a Platonic JEPA embedding, or to answer a question about it, and it reads the mapped input at test time. The anchor, the map and the decoder form our \emph{captioner}. The \emph{in-context reader} is a generalist language model that is never trained. Its prompt holds the five corpus texts nearest to the mapped input, which we call its \emph{hypotheses}, followed by a question. The same steps, prompts and recipe serve every pair of modalities, and no network is ever trained on a pair.

\subsection{Natural Images: Finding Better Views than Human Captions}\label{sec:flickr}

We first validate PVS where good views are known. In Flickr30k \citep{young2014flickr}, each photo has five human captions. The anchor is DINOv2 ViT-B, and the corpus is the 145k training captions, each treated as a text on its own. Only the \emph{oracle} uses the other captions of the same photo as positives, which are the good views of \citet{balestriero2026prh}. We report image-to-caption R@1 and R@5 on the 1{,}000 test pairs, through one rotation fitted on the validation pairs, and mutual 10-NN, which fits no map \citep{huh2024platonic}.

\begin{figure}[t]
\centering
\includegraphics{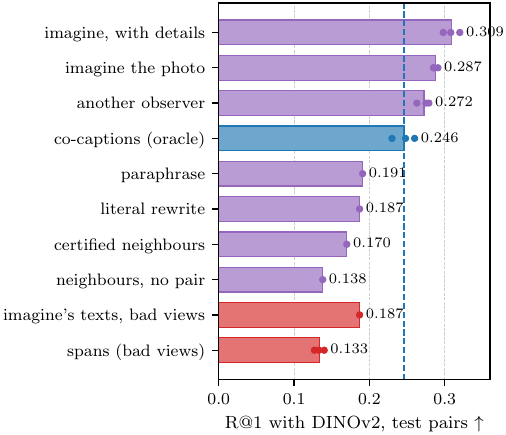}
\caption{\textbf{The reward ranks text views, and the best are text-only.} BERT-S text JEPAs trained on the Flickr30k training captions with one LeJEPA recipe. We report R@1 with DINOv2 on the test pairs. Dots: seeds. Dashed: the oracle, whose positives are the other human captions of the same photo. We observe that \emph{a language model imagining the photo gives better positives than the other human captions of the same photo}.}
\label{fig:flickr}
\end{figure}

\paragraph{Imagined views beat human captions.} The reward ranks the prompts as the test does, except for one near tie (\cref{fig:flickr,tab:flickr_views}). Imagining the photo beats the oracle. A paraphrase or a literal rewrite does much worse than imagining and stays well below the oracle, as \cref{prop:imagine} predicts. The best prompt asks for the photo's visual details. This prompt reaches an R@1 of .309 against .246 for the oracle, in every seed. Each caption has as many imagined views as co-captions, but the views add new texts to the corpus. Trained with bad views, this enlarged corpus explains only a third of the gain of imagined views over bad views, and using the views as positives explains the rest (\cref{fig:flickr}). As \cref{thm:ceiling}(ii) predicts, imagined views leave less nuisance in the JEPA. The fraction of each JEPA explained by the scene, $\rho_f=\mathrm{tr}\,\mC_f^{1/2}/k$, rises from bad views to imagined views with visual details, and it orders the five main encoders as alignment does (\cref{tab:flickr_views}). In short, \emph{imagined views leave less caption-specific nuisance than real co-captions}.

\paragraph{Imagined views do not recover the photo.} The language model never sees the photo, so the details that a caption leaves open come from its prior. The photo of \cref{fig:teaser} shows a woman in front of a wall with graffiti, which the caption does not mention but a co-caption does. The views written for this caption describe a typical tennis court instead, with a yellow shirt in one and an orange outfit in another (\cref{tab:gallery}). The rewrites neither copy Flickr30k captions nor drift towards the other captions of their photo (\cref{app:flickr_details}). As \cref{sec:exist} explains, the views keep what the caption states and resample the rest, which the caption cannot predict, whereas a co-caption adds details that the caption does not state, e.g., the graffiti. Resampling unstated colors suits DINOv2, which is trained to be invariant to color jitter and grayscale, but not a task that needs a detail that the anchor encodes and the views resample (\cref{sec:limits}). Unlike co-captions, imagined views also exist for texts without any photo (\cref{sec:scale}).

\begin{table}[t]
\centering
\small
\setlength{\tabcolsep}{3.5pt}
\caption{\textbf{The ranking is not tuned to DINOv2, nor to one learning rate.} Top: the stable recipe of \citet{balestriero2026prh} (learning rate $10^{-4}$, gradient clipping) over three seeds; no run collapses. Bottom: other evaluations of the seed-0 encoders. We report the test R@1 with six image encoders that played no role in the search (DINOv2 \citep{oquab2024dinov2}, DINOv3 \citep{simeoni2025dinov3}, I-JEPA \citep{assran2023ijepa}, MAE \citep{he2022masked}), each through a rotation fitted on 30k pairs (\texttt{scripts/slurm/families.sh}). We also report the STS-B Spearman correlation (text only) and the R@1 on COCO-5k, which no model was trained on.}
\label{tab:stable}
\begin{tabular}{lcccc}
\toprule
(all $\uparrow$) & bad & oracle & imagine & + details \\
\midrule
R@1, stable recipe & .138 & .245 & .272 & \textbf{.279} \\
R@5, stable recipe & .350 & .564 & .579 & \textbf{.596} \\
\midrule
DINOv2-L & .216 & .352 & \textbf{.411} & .394 \\
DINOv2-g & .246 & .357 & .398 & \textbf{.414} \\
DINOv3-L & .206 & .357 & \textbf{.400} & .398 \\
DINOv3-H & .174 & .300 & \textbf{.359} & .354 \\
I-JEPA-H & .096 & .170 & .174 & \textbf{.189} \\
MAE-H & .049 & .091 & \textbf{.099} & .090 \\
STS-B & .528 & .584 & \textbf{.640} & .637 \\
COCO-5k R@1 & .066 & .095 & .114 & \textbf{.124} \\
\bottomrule
\end{tabular}
\end{table}

\begin{figure}[t]
\centering
\includegraphics{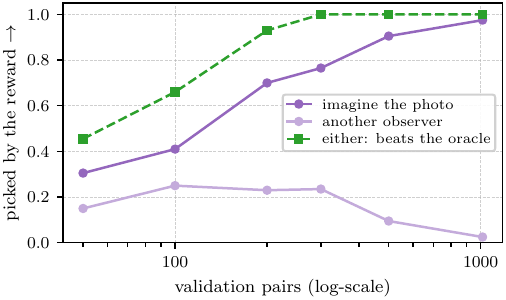}
\caption{\textbf{A few hundred pairs suffice to choose.} Among the first eight candidates, how often the reward computed from $n$ validation pairs picks each one, over 200 draws of the pairs. We observe that \emph{from 300 pairs on, the reward always picks a generator that beats the human captions}.}
\label{fig:pairs}
\end{figure}

\paragraph{Robustness.} The ranking holds with a stable learning rate, with six image encoders, on a text-only task and on another dataset (\cref{tab:stable}). With 300 validation pairs or more, the reward always picks a generator that beats the oracle (\cref{fig:pairs}). The ranking holds even with a much weaker anchor: a ViT-S trained on the 29k Flickr photos still ranks the imagined generators above the human captions (\cref{app:flickr_details}).

\subsection{Platonic JEPAs of Language at Scale}\label{sec:scale}

Our goal is now to pretrain a text JEPA on plain text, which, unlike captions, was not written for images, and to show that Platonic views make it align with images better than any language model we tested. Existing text JEPAs and sentence encoders use dropout noise, nearby spans, masks or meaning-preserving rewrites as positives \citep{gao2021simcse,giorgi2021declutr,zhang2023syncse,baevski2022data2vec,huang2025llmjepa,huang2026stp}, all of which keep a text's wording and are thus bad views. We do not use contrastive embedders as our text side, because their supervised pairs include caption-derived pairs (all-mpnet-base-v2) or may include them (Qwen3-Embedding \citep{zhang2025qwen3embedding}); we report two of them only as references (\cref{app:scale_details}).

\paragraph{Setup.} We split English Wikipedia into sentences. We remove every sentence that matches a caption of WIT-1024 \citep{srinivasan2021wit}, the benchmark of the PRH \citep{huh2024platonic}. Hence, no sentence was written for an image, selected by an image--text model or grouped with another sentence. The text-only language model of \cref{sec:impl} writes the positives. The bad-view baseline trains on the sentences alone with the same recipe. The reward and the test still use DINOv2 and the Flickr30k pairs. With BERT-S on 1M sentences, the reward ranks the generators as on Flickr30k, and the test pairs confirm this ranking. A photograph imagined for each sentence gives 2.5 times the R@1 of bad views, while a paraphrase of the sentence barely improves on them (\cref{tab:wiki1m}).

\begin{table}[t]
\centering\small
\caption{\textbf{PVS on plain Wikipedia.} BERT-S text JEPAs trained on 1M Wikipedia sentences, each with one rewrite (10M samples, one seed), and evaluated against frozen DINOv2 on Flickr30k. We report the reward (validation halves) and the test R@1, R@5 and mutual 10-NN (mNN). We observe that \emph{the reward picks the imagined photograph, which multiplies the bad-view R@1 by 2.5 without using a single caption}.}
\label{tab:wiki1m}
\setlength{\tabcolsep}{4pt}
\begin{tabular}{lcccc}
\toprule
Positives of a sentence (all $\uparrow$) & reward & R@1 & R@5 & mNN\\
\midrule
itself (bad views) & .081 & .066 & .211 & .070\\
LLM: paraphrase & .092 & .088 & .233 & .078\\
LLM: caption a photograph & .148 & .131 & .318 & .123\\
LLM: imagine a photograph & \textbf{.174} & \textbf{.164} & \textbf{.392} & \textbf{.154}\\
\bottomrule
\end{tabular}
\end{table}

\begin{figure}[t]
\centering
\includegraphics[width=\linewidth]{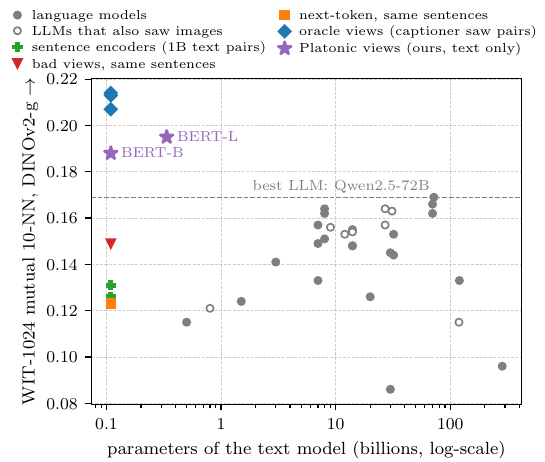}
\caption{\textbf{A 109M text JEPA trained on Wikipedia alone aligns with DINOv2 better than every language model we tested.} WIT-1024 mutual 10-NN with DINOv2-g \citep{huh2024platonic} against the size of the text model. We show 30 language models from 0.5B to 284B parameters (open circles: multimodal models), two contrastive sentence encoders trained on about a billion human-made text pairs, and our JEPAs (BERT-B, 109M; BERT-L, 335M, from Google's masked-language-model weights). Oracle views: the LLaVA recaptions of 940M web images, written by an image captioner (three seeds). Platonic views: 10M Wikipedia sentences, each with two descriptions of a photograph that a text-only language model imagined for it. Bad views and next-token prediction use the same sentences. We observe that \emph{Platonic views, with no image anywhere in training, beat language models up to 2{,}600 times larger, multimodal ones included}.}
\label{fig:models}
\end{figure}

\paragraph{Platonic views beat every language model.} We pretrain BERT-B on 10M sentences, each with two descriptions of a photograph imagined for it. We use the LeJEPA recipe \citep{balestriero2025lejepa} of the 940M-image JEPA of \citet{balestriero2026prh}. On WIT-1024, we measure the mutual 10-NN with DINOv2-g, which fits no map, and this JEPA reaches .188. All 30 language models we tested, from 0.5B to 284B parameters, stay below it (\cref{tab:wit_models}). The best is Qwen2.5-72B (.169), and the latest open models, e.g., Gemma-4-31B, Qwen3.8-27B, gpt-oss-120b and DeepSeek-V4-Flash, are lower. Our JEPA closes nearly two thirds of the gap between bad views and an oracle trained on the recaptions of 940M images (\cref{fig:models,tab:wiki_scale}). Contrastive sentence encoders, trained on about a billion human-made text pairs \citep{li2023gte,reimers2019sentencebert}, stay well below our JEPA, but they catch up in Flickr30k retrieval once the rotation is fitted on enough pairs (\cref{tab:flickr_pairs}). More views or more sentences do not help BERT-B, but capacity does: BERT-L reaches .195. These conclusions hold with the 7B DINOv3 anchor and over three seeds (\cref{tab:wiki_scale}).

\begin{table*}[t]
\centering\small
\caption{\textbf{Building a captioner whose networks never see a pair, with every image-side and decoding choice made on validation pairs.} COCO Karpathy CIDEr ($\times$100) on the 300 validation pairs used for every choice, and once on the other 4{,}700 images of the Karpathy test split (published rows: all 5{,}000). Our decoders read the Platonic JEPA trained on Wikipedia (\cref{sec:scale}). The only parameters fitted on pairs are those of one rotation (29k Flickr30k or 83k COCO training pairs). References, as published: text-only captioners built on CLIP inherit its 400M training pairs, and ClipCap and BLIP-2 train on COCO's pairs. Open VLMs are zero-shot and scored by our code on the same 4{,}700 images (best of a plain and a style-hinted prompt, \cref{sec:domains}; mean caption length in words). CIDEr rewards COCO's short captions. We observe that \emph{reading two self-supervised image encoders together and decoding by consensus each add more than the larger decoder}.}
\label{tab:captioner}
\setlength{\tabcolsep}{3pt}
\begin{tabular}{lccc}
\toprule
Captioner & image--text pairs & val $\uparrow$ & test $\uparrow$\\
\midrule
Qwen2.5-3B decoder, DINOv2-g & 29k, rotation only & 52.3 & 54.4\\
Qwen2.5-7B decoder, DINOv2-g & 29k, rotation only & 54.4 & 55.6\\
\quad + DINOv3-L read with DINOv2-g & 29k, rotation only & 64.2 & 63.6\\
\quad + consensus, 3 beams and 32 samples & 29k, rotation only & 67.0 & 66.4\\
\quad rotation from COCO's training pairs instead & 83k, rotation only & 91.1 & 85.7\\
\quad\quad + consensus & 83k, rotation only & 93.6 & 89.6\\
\quad\quad + DINOv3-7B read with the other two & 83k, rotation only & 97.4 & 88.0\\
\quad\quad\quad + consensus & 83k, rotation only & 99.0 & 92.0\\
\quad\quad\quad\quad 128 samples instead of 32 & 83k, rotation only & \textbf{100.9} & \textbf{93.0}\\
\quad\quad\quad decoder through the Platonic BERT-L instead of BERT-B & 83k, rotation only & 90.6 & 85.9\\
\midrule
\multicolumn{4}{l}{\emph{Other captioners: published test scores, or zero-shot on the test images}}\\
\citet{feng2019unsupervised}, unpaired & none (labeled detector) & \multicolumn{2}{c}{54.9}\\
DeCap \citep{li2023decap}, text-only & 400M (CLIP ViT-B/32) & \multicolumn{2}{c}{91.2}\\
CapDec \citep{nukrai2022capdec}, text-only & 400M (CLIP RN50x4) & \multicolumn{2}{c}{91.8}\\
ClipCap \citep{mokady2021clipcap} & 400M + COCO & \multicolumn{2}{c}{113.1}\\
BLIP-2 OPT-2.7B, fine-tuned \citep{li2023blip2,zhang2022opt} & 129M + COCO & \multicolumn{2}{c}{145.8}\\
\midrule
LLaVA-OneVision-7B \citep{li2025llavaonevision}, zero-shot (10.6 words) & web scale & \multicolumn{2}{c}{136.1}\\
InternVL3-8B \citep{zhu2025internvl3}, zero-shot (18.2 words) & web scale & \multicolumn{2}{c}{48.7}\\
Qwen2.5-VL-7B \citep{bai2025qwen25vl}, zero-shot (21.7 words) & web scale & \multicolumn{2}{c}{19.6}\\
Qwen2.5-VL-32B, zero-shot (29.4 words) & web scale & \multicolumn{2}{c}{7.1}\\
\bottomrule
\end{tabular}
\end{table*}

\begin{figure}[t]
\centering
\includegraphics{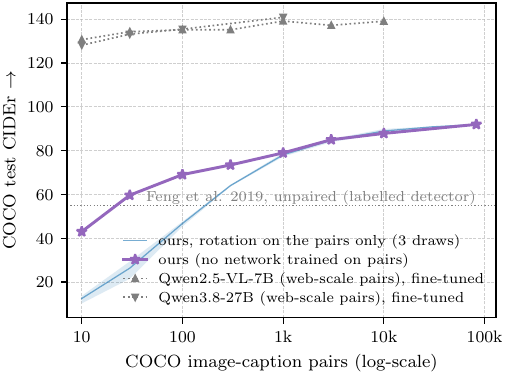}
\caption{\textbf{A handful of pairs anchor the alignment.} COCO test CIDEr against the number of image--caption pairs, with the same frozen networks throughout (DINOv2-g + DINOv3-L + DINOv3-7B, the text-only Qwen2.5-7B decoder, consensus of 3 beams and 32 samples). Purple: ours, with the map chosen on validation for each $n$ (\cref{sec:impl}). Blue: a rotation fitted on the $n$ pairs only (three draws, mean and range). Grey: VLMs fine-tuned on the same pairs (Qwen2.5-VL-7B; Qwen3.8-27B) (\texttt{paper/figures.py}, \texttt{data/domain\_captions.csv}, \texttt{data/domain\_vlm.csv}). We observe that \emph{the score never falls as pairs are added, and 30 pairs already beat the best unpaired captioner}.}
\label{fig:captioner_pairs}
\end{figure}

\paragraph{A captioner that never saw a pair.} We now build the captioner of \cref{sec:impl} on the Platonic JEPA of Wikipedia. The decoder (Qwen2.5-7B) is trained on Flickr30k and COCO captions as text, through this frozen JEPA \citep{mokady2021clipcap}. The anchor is a set of frozen image encoders (DINOv2-g and DINOv3), and the map is one rotation into the JEPA's space, the only parameter that ever sees a pair. We make every choice on 300 COCO validation pairs (\cref{tab:captioner}). With the rotation fitted on the Flickr30k training pairs, the captioner reaches a CIDEr of 66.4 on the COCO Karpathy test, above the unpaired captioner of \citet{feng2019unsupervised} (54.9), with no image--text model anywhere. With COCO's own pairs, the captioner reaches 93.0, on par with text-only captioners built on CLIP (\cref{tab:coco}). With few pairs, we fit the anchored alignment of \cref{sec:impl} \citep{schnaus2026rosetta} on them, and the captioner stays above \citet{feng2019unsupervised} from 30 pairs on. VLMs fine-tuned on the same pairs read photographs natively and remain far ahead (\cref{fig:captioner_pairs,tab:coco}; details in \cref{app:scale_details}).

\section{Platonic Reasoning Across Fourteen Domains}\label{sec:exp}

We now run the same protocol in 14 domains. Unlike in \cref{sec:pvs}, the anchors read modalities such as EEG, sounds and physical simulations. We first state what each component sees (\cref{sec:domains}). We then compare our captions and retrieval with fine-tuned models as the number of pairs grows (\cref{sec:vlm}), and finally answer short and open questions with 10 pairs (\cref{sec:reasoning}).

\subsection{Domains and Protocol}\label{sec:domains}

\paragraph{Who sees what.} Each domain has three ingredients: an anchor, i.e., a frozen self-supervised encoder trained without text; a corpus of unpaired texts about the modality; and a set of pairs. The anchor never sees a text. The Platonic JEPA is trained on the corpus and its views, and the decoder on the corpus and on questions that a language model writes from it, with their answers. The in-context reader is not trained. Only the map is fitted on pairs: \cref{sec:vlm} varies their number $n$, and \cref{sec:reasoning} uses 10 pairs unless stated. A few hundred separate validation pairs rank the views and choose every setting, but they train no network (\cref{app:domains}). At test time, retrieval, the decoder or the in-context reader reads each test input through the anchor and the map. The VLMs that we compare with read the test input itself, out of the box or fine-tuned on the same pairs.

\begin{table*}[t]
\centering\small
\caption{\textbf{What PVS picks, and how stably.} Reward of the five generic prompts and of bad views in every domain. Each candidate trains a BERT-S JEPA with one recipe. The reward is the R@1 ($\times$100) over the distinct texts through one rotation, on validation pairs only, averaged over five random halvings (\texttt{platonic/domain.py reward}). Bold: the pick (ties broken by mutual 10-NN). Gain: the pick's reward over that of bad views, in percent. Right: how often bad views, and the pick, have the highest reward on random subsets of 25, 100 and all validation pairs (50 draws, one halving each; \texttt{platonic/domain.py stability}). Last row: means over the domains. On photographs, the photo wording of the Wikipedia runs of \cref{sec:scale} scores 20.0, within noise of the generic imagine prompt. We observe that \emph{the pick beats bad views in every domain, while the generic prompts are often within noise of one another}.}
\label{tab:pvs}
\setlength{\tabcolsep}{4pt}
\begin{tabular}{lccccccc|cc}
\toprule
 & \multicolumn{6}{c}{reward of each candidate $\uparrow$} & gain & bad views win & pick wins\\
Domain & bad views & imagine & paraphrase & literal & report & caption & (\%) & 25 / 100 / all pairs & all pairs\\
\midrule
Photographs & 13.1 & \textbf{21.6} & 15.5 & 19.2 & 17.2 & 18.0 & 65 & .08 / .00 / .00 & .72 \\
Satellite images & 23.1 & \textbf{24.5} & 23.8 & 24.0 & 24.0 & 23.2 & 6 & .08 / .22 / .02 & .28 \\
Sounds (AudioCaps) & 5.2 & 7.3 & 7.2 & 7.1 & 6.9 & \textbf{7.7} & 49 & .08 / .02 / .02 & .40 \\
Chest X-rays & 3.0 & 4.3 & 2.7 & 2.3 & 4.0 & \textbf{4.5} & 50 & .16 / .10 / .04 & .38 \\
Molecules & 17.2 & 19.3 & 18.5 & 19.2 & \textbf{19.6} & 18.1 & 14 & .08 / .04 / .04 & .12 \\
Proteins & 5.2 & 6.4 & 5.5 & 5.9 & 4.9 & \textbf{7.5} & 44 & .16 / .12 / .12 & .30 \\
Videos & 7.7 & 8.5 & \textbf{10.0} & 7.9 & 9.1 & 9.6 & 29 & .04 / .04 / .06 & .20 \\
Physical simulations & 23.3 & 25.3 & 26.3 & 23.3 & 25.7 & \textbf{30.5} & 31 & .04 / .06 / .02 & .50 \\
Galaxies & 0.8 & \textbf{1.1} & 0.7 & 0.4 & 0.7 & 0.9 & 34 & .16 / .16 / .06 & .38 \\
EEG & 28.5 & 33.1 & 31.0 & 30.1 & 30.9 & \textbf{33.3} & 17 & .02 / .10 / .00 & .24 \\
Wrist motion & 36.8 & 37.5 & 36.1 & 36.3 & \textbf{38.1} & 36.3 & 4 & .18 / .20 / .20 & .34 \\
Lung sounds & 42.9 & \textbf{45.2} & 43.4 & 42.4 & 41.9 & 45.0 & 5 & .02 / .04 / .00 & .50 \\
Bird calls & 5.4 & \textbf{6.7} & 6.0 & 4.8 & 5.4 & 6.6 & 25 & .20 / .22 / .24 & .36 \\
Insect and frog calls & 10.4 & 9.4 & 9.8 & \textbf{10.9} & 10.1 & 10.6 & 5 & .16 / .20 / .04 & .32 \\
\midrule
mean & & & & & & & & .10 / .11 / .06 & .36 \\
\bottomrule
\end{tabular}
\end{table*}

\begin{table}[t]
\centering\footnotesize
\caption{\textbf{Platonic views, as selected by PVS.} One text and two of its views in two domains. The views are written by the generic prompt that PVS picked (\cref{tab:pvs}). \Cref{app:gallery} shows every domain (\texttt{paper/analysis.py gallery}).}
\label{tab:views_main}
\begin{tabular}{@{}lp{6.4cm}@{}}
\toprule
\multicolumn{2}{l}{\emph{Satellite images} (pick: \texttt{pvs-imagine})}\\
Caption & There is a piece of cropland .\\
View 1 & There is a rectangular plot of cropland with rows of crops running north to south.\\
View 2 & There is a rectangular plot of cropland, with rows of crops aligned in a grid pattern.\\
\midrule
\multicolumn{2}{l}{\emph{Molecules} (pick: \texttt{pvs-report})}\\
Description & The molecule is a member of the class of biphenyls that is biphenyl in which one of the phenyl groups is substituted at positions 2and 4 by a~\ldots\\
View 1 & Molecule Report - Class: Biphenyl - Structure: - Phenyl 1: Me, 2-Me, 5-(N-(cyclopropyl)aminocarboxyl) - Phenyl 2: Me, 2-Me, 5-5-Me-1,3,4-oxadiazol-2-\\
View 2 & Molecule Analysis Report - Class: Biphenyl - Structure: - Phenyl 1: 2-Me, 4-(5-Me-1,3,4-oxadiazol-2-yl) - Phenyl 2: 2-Me, 5-(N-cyclopropylaminocarbonyl) - Role:\\
\bottomrule
\end{tabular}
\end{table}

\paragraph{The search in every domain.} In all 14 domains, the selected views have a higher reward than bad views of the same texts (\cref{tab:pvs}; \cref{tab:views_main} shows the selected views). The gain is largest on photographs, chest X-rays, everyday sounds and proteins (44 to 65\%), and smallest on satellite images, wrist motion, lung sounds, and insect and frog calls (4 to 6\%). On random halves of the validation pairs, bad views have the highest reward in only 6\% of draws on average. The winning prompt is less stable, as the prompts are often within a few hits of one another. In 9 of the 14 domains, the generic imagine prompt is the search's pick, or is within a fifth of the gap between the pick and bad views. Elsewhere, the imagine prompt can fall well short of the pick, e.g., on simulations, or even below bad views, on insect and frog calls (\cref{tab:pvs}).

\subsection{Retrieval and Captioning Against Fine-Tuned VLMs}\label{sec:vlm}

\begin{figure*}[t]
\centering
\includegraphics[width=\textwidth]{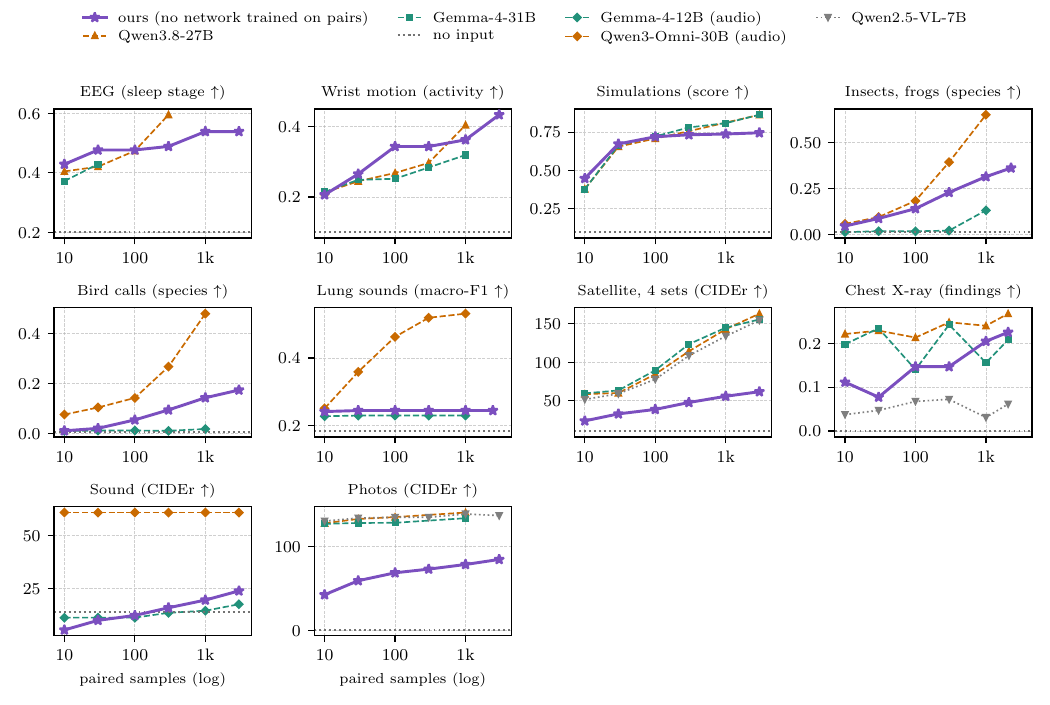}
\caption{\textbf{Ours against the latest open VLMs, from ten pairs to thousands.} Test score of each domain against the number of pairs. Dotted: the best caption that reads no input. Ours (purple): one frozen self-supervised encoder, one Platonic text JEPA, and one map per number of pairs, rotation or anchored alignment, chosen on validation. This figure uses one draw of the pairs, and the three-draw comparisons of the text, up to 300 pairs, also allow the ridge map. VLMs: LoRA-fine-tuned on the same pairs, with configuration and epoch chosen on validation. We observe that \textbf{with few pairs, a frozen encoder and one map beat the fine-tuned VLMs on EEG and simulations, and on wrist motion from 30 pairs. Ours beats Gemma-4-12B on every sound from 100 pairs, although on everyday sounds both stay below the caption that reads no input up to 100 pairs. Ours loses on photographs, satellite images and chest X-rays, and on sounds to Qwen3-Omni, except on lung sounds with 10 pairs}.}
\label{fig:vlm_curves}
\end{figure*}

Here, one map is fitted for each number $n$ of pairs, from 10 to thousands, and our captions are compared with those of VLMs fine-tuned on the same pairs. We use the latest open models, Qwen3.8-27B \citep{qwen2026qwen38} and Gemma-4-31B \citep{gemma2026gemma4}, and for sounds Qwen3-Omni-30B \citep{qwen2025qwen3omni} and Gemma-4-12B. Each VLM reads the modality in a format it was built for. Sounds are given as waveforms, and signals and simulations as images of their traces or fields. Each VLM is LoRA-fine-tuned on the same $n$ pairs as ours, and we choose its configuration and epoch on validation (\texttt{scripts/vlm\_finetune.py}). Our caption of an input is the corpus text nearest to the mapped input, except on photographs and satellite images, where our captioner writes it. We score what a text states about its input, e.g., the sleep stage of an EEG epoch, the activity of a wrist recording, or the system and parameters of a simulation (\cref{app:domains}). Unlike the VLMs, which read the input and are fine-tuned on our pairs, each of our networks sees one modality and none is trained on a pair.

\begin{figure*}[t]
\centering
\includegraphics[width=\textwidth]{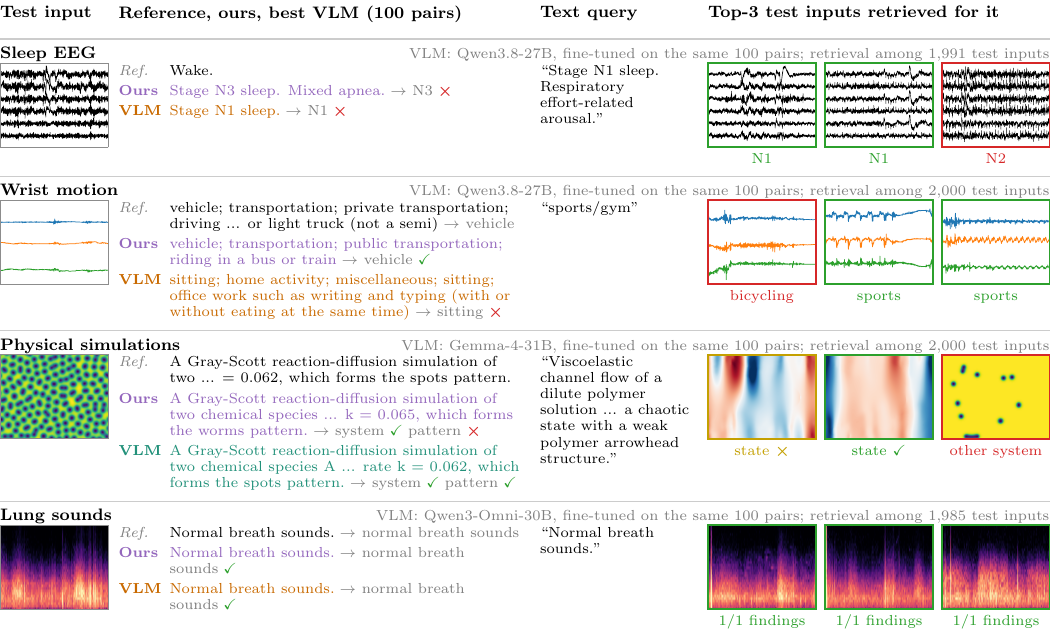}
\caption{\textbf{Captions and retrieval on held-out inputs, with 100 pairs.} We show the four domains where our captions lead the fine-tuned VLMs with few pairs (\cref{sec:vlm}); with 100 pairs, the VLM leads on lung sounds and is on par on simulations. Each row is one test input drawn at random (seed 0), never chosen by whether an answer is right. We show its reference, our caption through the map chosen on validation, and the caption of the best VLM fine-tuned on the same 100 pairs. For EEG and lung sounds, validation chose the map fitted on 30 of the 100 pairs. Marks follow each domain's metric. Right: the three test inputs that our map ranks first for a text query, the reference of another random test input. Frames: green, the input has all of the query's label; amber, part of it; red, none (\texttt{paper/gallery.py}). \Cref{fig:gallery} shows three domains where VLMs lead. We observe that \emph{our captions and the VLM's are each fully right on 2 of the 4 inputs: both miss the EEG stage, ours names the wrist activity where the VLM says sitting, and the VLM names the simulation's pattern where ours names another one. 8 of the 12 retrieved inputs have all of the query's label.}}
\label{fig:gallery_main}
\end{figure*}

\paragraph{Ahead on signals and scientific images with ten pairs.} With 10 pairs, our captions are ahead of the best fine-tuned VLM on sleep EEG, read by CBraMod \citep{wang2025cbramod}, and on simulations from The Well \citep{ohana2024well}, read by Walrus \citep{mccabe2025walrus}. Our captions are also ahead of Qwen3-Omni-30B on lung sounds, and on wrist motion from 30 pairs. Each comparison averages the test score over three random draws of the $n$ pairs (one for a few VLM runs, e.g., Qwen3.8-27B on EEG; \cref{tab:comparison}). We call ours \emph{ahead} when the whole 95\% confidence interval of the difference between the two mean scores is above zero. This interval holds the central 95\% of the differences over 2{,}000 resamples of the test inputs alone, not of the draws, with both methods scored on the same inputs. With 10 pairs selected with the anchor \citep{hacohen2022typiclust}, the VLMs improve too, but ours stays ahead on EEG, simulations and lung sounds, and moves ahead on wrist motion (\cref{app:selection}). With 100 and 300 pairs, our captions name the right bird, insect or frog species several times as often as Gemma-4-12B, which is barely above chance. Qwen3-Omni-30B names some species out of the box and stays ahead. On photographs, satellite images, chest X-rays and everyday sounds, which VLMs already read, the best fine-tuned VLM is far ahead from 10 pairs on (\cref{fig:vlm_curves,tab:curves}). There, our captions of everyday sounds only beat a caption that reads no input from about 300 pairs. On bird calls, our answers to the label questions of \cref{sec:reasoning} and our retrieval also beat this floor only with more than 10 pairs (dotted in \cref{fig:vlm_curves}; \cref{tab:labelqa,tab:retrieval}). On templated corpora, the anchor does the reading rather than the text side, and off-the-shelf sentence encoders, through the same rotation, are on par with our Platonic JEPAs or ahead of them. With more pairs, a classifier on the anchor's features takes over when the metric only reads labels (\cref{tab:comparison}).

\paragraph{Retrieval and cost.}\label{sec:robust} The same maps also retrieve inputs from a text query (\cref{fig:gallery_main} shows random examples). With 10 pairs, ours retrieves EEG as well as embedding models fine-tuned on the same pairs \citep{tschannen2025siglip2,qwen2026qwen3vlemb,meta2025peav}. With more pairs, ours retrieves chest X-rays better. Those models are far ahead on photographs, satellite images and everyday sounds (\cref{tab:retrieval}). Captioning by retrieval needs only the anchor, the map and a lookup in the corpus. Captioning by retrieval is much faster and needs far less memory than the fine-tuned VLMs, and for most domains it runs fast on a CPU. Adding pairs only requires fitting the map again, in closed form (\cref{tab:cost}; \cref{app:vlm_details} gives every comparison).

\begin{table}[t]
\centering\footnotesize
\caption{\textbf{Reasoning with no image--text model.} Test VQAv2 accuracy, in percent. Ours: the decoder reads the input through the map chosen on validation, either the rotation or the anchored alignment, fitted on the number of pairs in parentheses. The decoder is trained on text alone and never reads an image. The last row feeds it the embedding of the photo's own caption instead of the mapped input (\texttt{scripts/vqa.py}).}
\label{tab:reasoning}
\setlength{\tabcolsep}{2.5pt}
\begin{tabular}{lcccc}
\toprule
\multicolumn{5}{l}{\emph{VQAv2 (photographs), VQA accuracy}}\\
 & all & yes/no & number & other\\
\midrule
Qwen3.8 text, question alone & 42.8 & 67.8 & 29.7 & 27.1\\
gpt-oss, question alone & 41.1 & 67.4 & 25.7 & 24.9\\
gpt-oss, nearest caption (10) & 36.6 & 66.2 & 15.1 & 19.4\\
gpt-oss, nearest caption (1{,}000) & 45.7 & 70.7 & 25.1 & 31.9\\
\textbf{ours} + gpt-oss, our caption (10) & 43.0 & 69.6 & 22.0 & 28.0\\
\textbf{ours} + gpt-oss, our caption (1{,}000) & 48.9 & 72.8 & 27.5 & 36.1\\
Qwen3.8-27B out of the box & 81.5 & 94.6 & 73.8 & 73.4\\
Gemma-4-31B out of the box & 76.1 & 94.2 & 71.3 & 63.5\\
same decoder, no input & 37.8 & 65.5 & 26.4 & 19.5\\
same decoder, nearest pair (100) & 43.2 & 67.8 & 30.7 & 27.5\\
\textbf{ours} (10) & 44.1 & 68.4 & 31.0 & 28.9\\
\textbf{ours} (100) & 46.8 & 69.8 & 31.4 & 33.2\\
\textbf{ours} (1{,}000) & 46.8 & 69.5 & 32.6 & 33.1\\
\textbf{ours} (10{,}000) & 47.5 & 70.0 & 32.2 & 34.2\\
\textbf{ours} (82{,}783) & 47.5 & 70.0 & 32.7 & 34.1\\
same decoder, photo's own caption & 47.1 & 70.0 & 34.8 & 32.7\\
\bottomrule
\end{tabular}
\end{table}

\subsection{Short and Open Questions}\label{sec:reasoning}

\begin{figure*}[p]
\centering\footnotesize
\setlength{\tabcolsep}{4pt}
\begin{tabular}{>{\raggedright\arraybackslash}p{3.3cm}p{\dimexpr\textwidth-3.3cm-4\tabcolsep\relax}}
\toprule
\multicolumn{2}{@{}l}{\emph{Questions about the simulated system}}\\[1pt]
\raisebox{-\height}{\includegraphics[width=3.2cm,height=2.2cm]{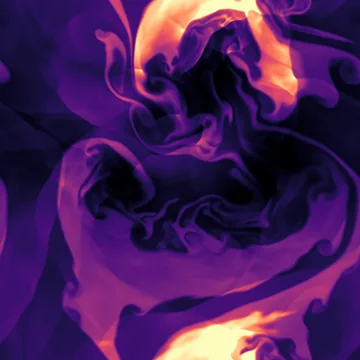}}\newline{\footnotesize Euler multi-quadrant gas dynamics} & \textbf{Q:} What physical system and governing equations are depicted in this simulation snapshot?\newline\textcolor{black!60}{\textbf{Reference:} It shows a compressible, inviscid gas evolving according to the two-dimensional Euler equations, initialized as a four-quadrant Riemann problem.}\newline\textcolor{black!60}{\textbf{Map's top hypothesis:} Compressible Euler equations for an inviscid gas: a two-dimensional four-quadrant Riemann problem with adiabatic index $\gamma$ = 1.4 (dry air at 20\dots{}}\newline\textcolor{found}{\textbf{Ours (10.0/10):}} The snapshot shows a two-dimensional, four-quadrant Riemann problem for an inviscid, compressible gas governed by the Euler equations of fluid dynamics. The simulation uses the compressible Euler system (conservation of mass, momentum\dots{}\newline\textcolor{bad}{\textbf{VLM (10.0/10):}} It shows a two-dimensional inviscid $\gamma$ = 1.4 air model governed by the Euler equations, where a symmetric arrangement of shock waves and contact discontinuities has emerged from the initial four-quadrant discontinuities.\\[4pt]
\midrule
\raisebox{-\height}{\includegraphics[width=3.2cm,height=2.2cm]{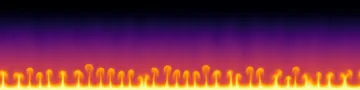}}\newline{\footnotesize Rayleigh-Bénard convection} & \textbf{Q:} How would the pattern change if the Rayleigh number were reduced by an order of magnitude?\newline\textcolor{black!60}{\textbf{Reference:} Reducing the Rayleigh number to 10$^{7}$ would weaken the buoyancy forcing, leading to larger, more coherent convection cells and a less turbulent, more steady roll-like pattern.}\newline\textcolor{black!60}{\textbf{Map's top hypothesis:} Two-dimensional Rayleigh-Bénard convection: a horizontally periodic fluid layer heated from below and cooled from above, at Rayleigh number 100,000,000 and Prandtl number\dots{}}\newline\textcolor{found}{\textbf{Ours (10.0/10):}} Reducing the Rayleigh number by a factor of ten weakens the buoyancy drive, so the convective rolls become larger and more regular and the flow speed drops. The pattern typically shifts from\dots{}\newline\textcolor{bad}{\textbf{VLM (3.0/10):}} Lowering the Rayleigh number to 10$^{10}$ would suppress the vigorous plumes, leading to smoother buoyancy contours and more subdued velocity fields, as the buoyancy force would be weaker relative to viscous damping.\\[4pt]
\midrule
\raisebox{-\height}{\includegraphics[width=3.2cm,height=2.2cm]{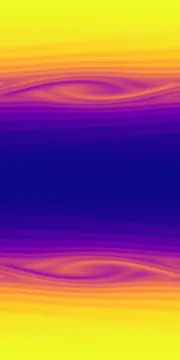}}\newline{\footnotesize Shear flow} & \textbf{Q:} How does the interaction of the two shear layers influence the development of structures in the flow over time?\newline\textcolor{black!60}{\textbf{Reference:} The shear layers roll up into vortices that interact and merge, producing a turbulent cascade that continuously stretches and folds the tracer field, enhancing mixing.}\newline\textcolor{black!60}{\textbf{Map's top hypothesis:} A two-dimensional periodic shear flow of an incompressible fluid carrying a passive tracer, at Reynolds number 10$^{5}$ and Schmidt number 0.1, started\dots{}}\newline\textcolor{found}{\textbf{Ours (10.0/10):}} The neighboring shear layers roll up into counter-rotating vortices that quickly interact, stretching and tilting each other's vorticity; this mutual straining amplifies the initial perturbations and produces a cascade of increasingly fine\dots{}\newline\textcolor{bad}{\textbf{VLM (9.0/10):}} The coexistence of the two shear layers leads to a symmetric pattern of vortex rolls that interact and merge, creating complex filamentary structures of the tracer that evolve as the flow becomes\dots{}\\[4pt]
\midrule
\raisebox{-\height}{\includegraphics[width=3.2cm,height=2.2cm]{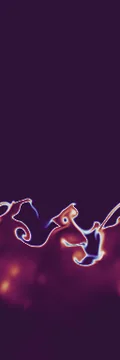}}\newline{\footnotesize Turbulent radiative layer} & \textbf{Q:} What practical implication does this snapshot suggest about radiative cooling in astrophysical shear layers?\newline\textcolor{black!60}{\textbf{Reference:} It indicates that turbulence can quickly generate intermediate-temperature gas that cools efficiently, implying that radiative cooling can be a dominant energy-loss mechanism in shear-driven mixing layers in astrophysical\dots{}}\newline\textcolor{black!60}{\textbf{Map's top hypothesis:} Hot and cold gas in relative motion become turbulent and mix, and the mixed gas radiates its energy away; this turbulent radiative\dots{}}\newline\textcolor{found}{\textbf{Ours (9.5/10):}} The snapshot shows that a turbulent shear layer formed by hot and cold gas mixes and then radiates away its thermal energy on a timescale comparable to the dynamical time (t\_cool $\approx$\dots{}\newline\textcolor{bad}{\textbf{VLM (0.0/10):}} It indicates that radiative cooling can suppress the growth of shear-driven instabilities, implying that in cooling astrophysical plasmas, shear layers may remain more coherent and less turbulent than in adiabatic models.\\[4pt]
\midrule\midrule
\multicolumn{2}{@{}l}{\emph{Questions about the dynamics of one window}}\\[1pt]
\raisebox{-\height}{\includegraphics[width=3.2cm,height=2.2cm]{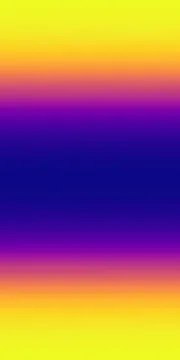}}\newline{\footnotesize Shear flow} & \textbf{Q:} What changes do you observe in the vortical activity and kinetic energy over the four snapshots?\newline\textcolor{black!60}{\textbf{Reference:} Over the window the overall vortical intensity remains steady and the kinetic energy is nearly constant, although the detailed pattern reshapes too quickly to be described as a\dots{}}\newline\textcolor{black!60}{\textbf{Map's top hypothesis:} \dots{} The vortical activity is strongest in the top-left and bottom-left of the domain, and weakest in the right. It is organised\dots{}}\newline\textcolor{found}{\textbf{Ours (6.0/10):}} The vortical activity does not change appreciably from one snapshot to the next -- the same coherent structures and banded pattern persist and the intensity in each region remains essentially steady. Likewise\dots{}\newline\textcolor{bad}{\textbf{VLM (2.5/10):}} The overall vortical activity remains steady, the kinetic energy is nearly constant, and the pattern does not drift or evolve noticeably during the window.\\[4pt]
\midrule
\raisebox{-\height}{\includegraphics[width=3.2cm,height=2.2cm]{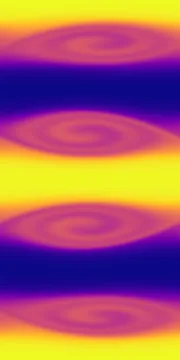}}\newline{\footnotesize Shear flow} & \textbf{Q:} Based on the current state, what would you predict for the next few snapshots---where might activity intensify or diminish?\newline\textcolor{black!60}{\textbf{Reference:} The simulation is likely to maintain the same pattern, with the central and left regions staying strongly vortical and the weak corners remaining calm; no significant intensification or\dots{}}\newline\textcolor{black!60}{\textbf{Map's top hypothesis:} \dots{} The vortical activity is strongest in the center and bottom of the domain, and weakest in the top-right. It is patchy\dots{}}\newline\textcolor{found}{\textbf{Ours (5.0/10):}} The vortical activity will probably stay concentrated in the central and lower parts of the domain, with the intense patches persisting or slowly expanding into the surrounding weak-vortex regions. The calm top-right\dots{}\newline\textcolor{bad}{\textbf{VLM (1.5/10):}} The top and bottom regions are expected to gain vortical strength, while the already strong bottom-left and top-left zones should remain active; the right side is likely to stay weak.\\[4pt]
\bottomrule
\end{tabular}
\caption{\textbf{Open questions about simulations, with 10 pairs.} All inputs are test windows of The Well, no model has seen them described. Top: questions about the simulated system. We draw 4 of the seven systems at random (seed 0) and one window per system. Bottom: questions about the dynamics of one window. We draw 2 other windows at random. Each window has five questions, and we draw one, never by grade. We show the reference answer and the first corpus description that the map retrieves, written for another window (at the bottom, from its second sentence on). Ours is gpt-oss-120b reading five such descriptions. The VLM is fine-tuned on the questions of the same 10 pairs and reads images of the window. Both use the first draw of the pairs. Each grade is the mean of two judges, out of 10. We observe that \emph{ours gets a higher grade than the VLM on 5 of these 6 questions, and never a lower one}.}
\label{fig:reasoning_examples}
\end{figure*}

\begin{table*}[t]
\centering\footnotesize
\caption{\textbf{Open questions with 10 pairs.} Test score on blind questions about each input (\texttt{scripts/openqa.py}, \texttt{scripts/dynqa.py}). Two judges, gpt-oss-120b and Qwen3.8-27B, grade each answer from 0 to 10, and the score is their mean grade times ten. VLMs read the input (best of Qwen3.8-27B and Gemma-4-31B), out of the box or LoRA-fine-tuned on the questions of the same 10 pairs as ours (first draw and, in parentheses, averaged over our three draws). No input: gpt-oss-120b answers from the question alone, our decoder answers from its prior, and gpt-oss-120b reads five corpus descriptions drawn at random in the prompt of our reader. Ours: the readers see the input through one map fitted on 10 pairs, on the first of three draws and, in parentheses, averaged over the three draws. Last column: the same in-context reader with gte-base instead of our Platonic JEPA as the text side. Bold: best on the first draw. Selected pairs: chosen with the anchor (\cref{app:selection}). --: not run.}
\label{tab:openqa}
\setlength{\tabcolsep}{3pt}
\begin{tabular}{lcccccccc}
\toprule
 & \multicolumn{2}{c}{VLM reading the input} & \multicolumn{3}{c}{no input} & \multicolumn{2}{c}{\textbf{ours}, through the map} & control\\
\cmidrule(lr){2-3}\cmidrule(lr){4-6}\cmidrule(lr){7-8}\cmidrule(lr){9-9}
 & out of the box & fine-tuned & gpt-oss & our decoder & gpt-oss, 5 random & our decoder & gpt-oss, top 5 & gte-base\\
\midrule
Simulations & 36.6 & 44.6 (51.3) & 37.1 & 51.9 & 36.8 (36.4) & 59.1 (61.0) & 59.7 (66.2) & \textbf{61.1} (69.0)\\
\quad selected pairs & & (54.5) & & & & (63.5) & (75.9) & --\\
One run's dynamics & 31.2 & 29.7 (32.6) & 25.2 & 34.5 & 33.8 (34.4) & \textbf{42.0} (43.3) & 41.1 (43.7) & --\\
\quad selected pairs & & (36.4) & & & & (44.9) & (46.7) & --\\
\bottomrule
\end{tabular}
\end{table*}

We now turn from captions to questions, with maps fitted on 10 pairs unless stated. Two simple baselines use the pairs directly: copying the caption of the nearest pair, and a classifier trained on the labels of the pairs. Both can only answer what some pair already states. A language model that reads the input's embedding can instead answer any question about what this embedding holds. We give two recipes for such a reader, and neither trains a network on a pair (\cref{fig:incontext}).

\paragraph{Two readers.} The first reader is the decoder of \cref{sec:impl}. During training, a base language model (Qwen3-8B-Base or Qwen2.5-7B, chosen on validation) reads the Platonic JEPA embedding of a corpus text, with Gaussian noise, through ten soft-prompt tokens, then reads a question and writes its answer. A text-only language model writes these questions and answers from the corpus (\texttt{scripts/vqa.py}). At test time, the mapped input replaces the text embedding. The second reader is the in-context reader. Its fixed prompt holds the map's five hypotheses, ranked, followed by the question. Any language model can read this prompt, and we use gpt-oss-120b \citep{openai2025gptoss}, a text-only model.

\paragraph{Baselines and benchmarks.} We compare with the VLMs of \cref{sec:vlm}, both out of the box and LoRA-fine-tuned on the questions of the same pairs. We also compare with strong text-only models that read the question alone or the text of the nearest pair. The benchmarks are VQAv2 \citep{goyal2017vqav2}, label questions in six domains, which ask for an attribute that the input's annotation states, e.g., the sleep stage, and open questions about the simulated system and about the dynamics of a single run. The open questions are blind, i.e., an expert who has not seen the input would ask them, as in CaReAQA \citep{wang2025careaqa}. Two judges from different model families grade the open answers from 0 to 10 (\cref{app:reasoning_details} gives every detail).

\paragraph{Short answers.} On VQAv2, the anchor, the Platonic JEPA and the map are those of the captioner of \cref{sec:scale}, and the decoder is trained on questions written from COCO captions. With no image--text model anywhere, the decoder reaches 44.1 with 10 pairs. The same decoder with no input reaches 37.8, and Qwen3.8-27B reaches 42.8 from the question alone, because most VQAv2 answers follow from the question. With 10 pairs, our decoder also beats the decoders of a bad-view JEPA and of gte-base whose maps use all 82{,}783 pairs (\cref{tab:vqa_detail}). VLMs that read photographs remain far ahead (81.5; \cref{tab:reasoning,fig:reasoning}). On label questions, the decoder beats the VLMs fine-tuned on the same pairs on simulations, and narrowly on sleep EEG, where the same decoder reading the nearest pair's text matches it (64.0 against 63.9). The decoder also beats its own score with no input. Models that already read a modality stay ahead on it, and Gemma-4-31B also leads on wrist motion (\cref{tab:labelqa}). Pairs selected with the anchor raise our scores in four of the six domains. By attribute (\cref{tab:attributes}), the map reads the system of a simulation and, with more pairs, its pattern and viscoelastic state. The map barely reads the transport coefficients or the initial conditions. These attributes vary within one system, and a VLM reads them from the rendering when they are visible.

\paragraph{Open questions.} With 10 pairs, the in-context reader scores 66.2 on blind open questions about simulations, averaged over three draws of the pairs. Qwen3.8-27B and Gemma-4-31B, fine-tuned on the questions of the same pairs, reach at most 51.3 over the same three draws. With 10 pairs selected with the anchor, which label all seven systems, the reader scores 75.9, against \selvlm{} for the VLMs fine-tuned on them. The newest open VLM, Qwen3.8-Flash-Next, a 177B preview of the Qwen4 architecture \citep{qwen2026qwen38}, also stays well behind, out of the box and fine-tuned on the same pairs (\cref{tab:openqa_detail}). This lead reflects the map and the reader more than the views. The simulation descriptions are templated: the corpus holds 21{,}000 descriptions but only 1{,}266 distinct texts, which describe 147 configurations. This corpus is thus close to the closed vocabulary of \cref{prop:closed}, where every encoder whose values are affinely independent is Platonic and only the map is left to learn. Any good sentence encoder should thus serve as the text side, and indeed gte-base through the same ridge map scores slightly higher than our Platonic JEPA (69.0 against 66.2 over three draws; \cref{tab:openqa}). The Platonic JEPA matters on free text instead, as the short answers on VQAv2 show (\cref{tab:vqa_detail}).

\paragraph{The gain comes mostly from recognizing the system.} Two controls show that the gain comes from reading the input. The decoder with no input scores 51.9. The same in-context reader given five random corpus descriptions instead of the map's hypotheses scores 36.4. Checked against each simulation's own description, with no judge, the first hypothesis names the right system for most test simulations, far more often than chance (\cref{tab:factcheck}). On the open questions about the dynamics of a single run, our in-context reader also scores 43.7, against 32.6 for the fine-tuned VLMs, 34.5 with no input and 34.4 with random descriptions (\cref{tab:openqa}). A third control gives the reader five random descriptions of the input's own system, which tells it the system for free. On the inputs whose system the map recognizes, this control scores as well as the map's hypotheses, on simulations and on the dynamics of one run (\cref{tab:system_control}). The gain on open questions thus comes mostly from recognizing the system from the input. Recognition improves quickly with pairs: on simulations, the reader with 30 or 100 pairs scores as high as when it is told the system, and the VLMs fine-tuned on the same pairs stay well behind (\cref{tab:openqa_detail}). Beyond the system, the map's hypotheses help predict how one run evolves, where they beat this control. The random-description control also tells when not to use the map. On mass spectra, an extra domain that the anchor barely aligns with text, the map's hypotheses name the wrong molecules, and the reader scores below the question alone (\cref{tab:openqa_detail}).

\paragraph{Other language models can read the hypotheses.} The in-context reader learns nothing from the pairs. Hence, any language model can take its place without fine-tuning, and we test three open-weight models (\cref{fig:incontext,fig:reasoning_examples}). For example, on simulations, Qwen3.8-27B and Gemma-4-31B score higher when they read the same hypotheses as text than when they are fine-tuned on the pairs (\cref{tab:icl_readers}). With gpt-oss-120b, fifteen times larger than our decoder, the in-context reader scores about as well as the decoder on the first draw, and a few points higher over three draws.

\FloatBarrier
\section{Conclusion}\label{sec:conclusion}

We proved that a frozen anchor caps how well any encoder of the target modality can align with it. Every anchor has Platonic views whose JEPA reaches that cap, and the alignment of a view's JEPA is maximal exactly at these views. PVS turns this theory into a recipe. With PVS, we found text views that beat human captions and pretrained language JEPAs on plain Wikipedia. Platonic Reasoning then reads modalities with little paired text, fitting only the map on 10 pairs and training no network on a pair. Platonic Reasoning uses the decoder, trained on text alone, or the in-context reader, i.e., a language model given the map's hypotheses. On simulations, and narrowly on label questions about sleep EEG, these readers answer better than the latest open-weight VLMs fine-tuned on the same pairs, and better than a model that never sees the input (\cref{tab:openqa,tab:labelqa,tab:factcheck}). In short, a VLM that already reads the modality should be fine-tuned on the pairs. Otherwise, open questions go to the in-context reader, short answers with about ten pairs to the decoder, and label sets with a few pairs per label to a classifier on the anchor. With about ten pairs and a few distinct classes, the pairs are best selected with the anchor. \Cref{tab:guide} gives this guide with its evidence, and \cref{sec:limits} the limitations. We hope that Platonic views will make JEPAs as easy to pretrain on text, sensors and science as they are on images, and will make language the interface to what they learn.

\section*{Acknowledgements}
We thank Matteo Gamba, Marcin Przewi\k{e}\'{z}likowski, Fabian Gr\"oger and Quentin Garrido for their feedback on the paper.

\bibliographystyle{plainnat}
\bibliography{references}

\begin{thebibliography}{121}
\providecommand{\natexlab}[1]{#1}
\providecommand{\url}[1]{\texttt{#1}}
\expandafter\ifx\csname urlstyle\endcsname\relax
  \providecommand{\doi}[1]{doi: #1}\else
  \providecommand{\doi}{doi: \begingroup \urlstyle{rm}\Url}\fi

\bibitem[Abdine et~al.(2024)Abdine, Chatzianastasis, Bouyioukos, and
  Vazirgiannis]{abdine2024prot2text}
Hadi Abdine, Michail Chatzianastasis, Costas Bouyioukos, and Michalis
  Vazirgiannis.
\newblock {Prot2Text}: Multimodal protein's function generation with {GNNs} and
  transformers.
\newblock In \emph{AAAI Conference on Artificial Intelligence}, pages
  10757--10765, 2024.

\bibitem[Assran et~al.(2023)Assran, Duval, Misra, Bojanowski, Vincent, Rabbat,
  LeCun, and Ballas]{assran2023ijepa}
Mahmoud Assran, Quentin Duval, Ishan Misra, Piotr Bojanowski, Pascal Vincent,
  Michael Rabbat, Yann LeCun, and Nicolas Ballas.
\newblock Self-supervised learning from images with a joint-embedding
  predictive architecture.
\newblock In \emph{IEEE/CVF Conference on Computer Vision and Pattern
  Recognition}, 2023.

\bibitem[Assran et~al.(2025)Assran, Bardes, Fan, Garrido, Howes,
  et~al.]{assran2025vjepa2}
Mahmoud Assran, Adrien Bardes, David Fan, Quentin Garrido, Russell Howes,
  et~al.
\newblock {V-JEPA 2}: Self-supervised video models enable understanding,
  prediction and planning.
\newblock \emph{arXiv preprint arXiv:2506.09985}, 2025.

\bibitem[Baevski et~al.(2022)Baevski, Hsu, Xu, Babu, Gu, and
  Auli]{baevski2022data2vec}
Alexei Baevski, Wei-Ning Hsu, Qiantong Xu, Arun Babu, Jiatao Gu, and Michael
  Auli.
\newblock data2vec: A general framework for self-supervised learning in speech,
  vision and language.
\newblock In \emph{International Conference on Machine Learning}, 2022.

\bibitem[Bai et~al.(2025)Bai, Chen, Liu, Wang, Ge, et~al.]{bai2025qwen25vl}
Shuai Bai, Keqin Chen, Xuejing Liu, Jialin Wang, Wenbin Ge, et~al.
\newblock {Qwen2.5-VL} technical report.
\newblock \emph{arXiv preprint arXiv:2502.13923}, 2025.

\bibitem[Balestriero and LeCun(2022)]{balestriero2022contrastive}
Randall Balestriero and Yann LeCun.
\newblock Contrastive and non-contrastive self-supervised learning recover
  global and local spectral embedding methods.
\newblock In \emph{Advances in Neural Information Processing Systems}, 2022.

\bibitem[Balestriero and LeCun(2025)]{balestriero2025lejepa}
Randall Balestriero and Yann LeCun.
\newblock {LeJEPA}: Provable and scalable self-supervised learning without the
  heuristics.
\newblock \emph{arXiv preprint arXiv:2511.08544}, 2025.

\bibitem[Balestriero and LeCun(2026)]{balestriero2026prh}
Randall Balestriero and Yann LeCun.
\newblock Proving and improving the {Platonic} representation hypothesis with
  {JEPAs}.
\newblock \emph{arXiv preprint arXiv:2610.39302}, 2026.

\bibitem[Banerjee and Lavie(2005)]{banerjee2005meteor}
Satanjeev Banerjee and Alon Lavie.
\newblock {METEOR}: An automatic metric for {MT} evaluation with improved
  correlation with human judgments.
\newblock In \emph{ACL Workshop on Intrinsic and Extrinsic Evaluation Measures
  for Machine Translation and/or Summarization}, pages 65--72, 2005.

\bibitem[Blum and Mitchell(1998)]{blum1998cotraining}
Avrim Blum and Tom Mitchell.
\newblock Combining labeled and unlabeled data with co-training.
\newblock In \emph{Proceedings of the Eleventh Annual Conference on
  Computational Learning Theory}, pages 92--100, 1998.

\bibitem[Bowman et~al.(2015)Bowman, Angeli, Potts, and Manning]{bowman2015snli}
Samuel~R. Bowman, Gabor Angeli, Christopher Potts, and Christopher~D. Manning.
\newblock A large annotated corpus for learning natural language inference.
\newblock In \emph{Proceedings of the Conference on Empirical Methods in
  Natural Language Processing (EMNLP)}, 2015.

\bibitem[Chan et~al.(2024)Chan, others, and Doherty]{chan2024capture24}
Shing Chan, others, and Aiden Doherty.
\newblock {CAPTURE-24}: A large dataset of wrist-worn activity tracker data
  collected in the wild for human activity recognition.
\newblock \emph{Scientific Data}, 11:\penalty0 1135, 2024.

\bibitem[Chasmai et~al.(2024)Chasmai, Shepard, Maji, and
  Van~Horn]{chasmai2024inatsounds}
Mustafa Chasmai, Alexander Shepard, Subhransu Maji, and Grant Van~Horn.
\newblock The {iNaturalist} sounds dataset.
\newblock In \emph{NeurIPS Datasets and Benchmarks Track}, 2024.

\bibitem[Chen et~al.(2024)Chen, Liang, Ma, Zheng, and Chen]{chen2024eat}
Wenxi Chen, Yuzhe Liang, Ziyang Ma, Zhisheng Zheng, and Xie Chen.
\newblock {EAT}: Self-supervised pre-training with efficient audio transformer.
\newblock In \emph{International Joint Conference on Artificial Intelligence},
  pages 3807--3815, 2024.

\bibitem[Chen et~al.(2020)Chen, Song, Chang, and Wan]{chen2020r2gen}
Zhihong Chen, Yan Song, Tsung-Hui Chang, and Xiang Wan.
\newblock Generating radiology reports via memory-driven transformer.
\newblock In \emph{Conference on Empirical Methods in Natural Language
  Processing}, pages 1439--1449, 2020.

\bibitem[Cheng et~al.(2022)Cheng, Huang, Xu, Zhou, Li, and Wang]{cheng2022nwpu}
Qimin Cheng, Haiyan Huang, Yuan Xu, Yuzhuo Zhou, Huanying Li, and Zhongyuan
  Wang.
\newblock {NWPU-Captions} dataset and {MLCA-Net} for remote sensing image
  captioning.
\newblock \emph{IEEE Transactions on Geoscience and Remote Sensing},
  60:\penalty0 1--19, 2022.

\bibitem[Chithrananda et~al.(2020)Chithrananda, Grand, and
  Ramsundar]{chithrananda2020chemberta}
Seyone Chithrananda, Gabriel Grand, and Bharath Ramsundar.
\newblock {ChemBERTa}: Large-scale self-supervised pretraining for molecular
  property prediction.
\newblock \emph{arXiv preprint arXiv:2010.09885}, 2020.

\bibitem[Chu et~al.(2024)Chu, Xu, Yang, Wei, Wei, et~al.]{chu2024qwen2audio}
Yunfei Chu, Jin Xu, Qian Yang, Haojie Wei, Xipin Wei, et~al.
\newblock {Qwen2-Audio} technical report.
\newblock \emph{arXiv preprint arXiv:2407.10759}, 2024.

\bibitem[Conneau et~al.(2018)Conneau, Lample, Ranzato, Denoyer, and
  J{\'e}gou]{conneau2018word}
Alexis Conneau, Guillaume Lample, Marc'Aurelio Ranzato, Ludovic Denoyer, and
  Herv{\'e} J{\'e}gou.
\newblock Word translation without parallel data.
\newblock In \emph{International Conference on Learning Representations}, 2018.

\bibitem[Cubuk et~al.(2019)Cubuk, Zoph, Mane, Vasudevan, and
  Le]{cubuk2019autoaugment}
Ekin~D Cubuk, Barret Zoph, Dandelion Mane, Vijay Vasudevan, and Quoc~V Le.
\newblock {AutoAugment}: Learning augmentation strategies from data.
\newblock In \emph{IEEE/CVF Conference on Computer Vision and Pattern
  Recognition}, 2019.

\bibitem[Dasgupta et~al.(2001)Dasgupta, Littman, and
  McAllester]{dasgupta2001pac}
Sanjoy Dasgupta, Michael~L Littman, and David McAllester.
\newblock {PAC} generalization bounds for co-training.
\newblock In \emph{Advances in Neural Information Processing Systems}, 2001.

\bibitem[Demner-Fushman et~al.(2016)Demner-Fushman, Kohli, Rosenman, Shooshan,
  Rodriguez, Antani, Thoma, and McDonald]{demnerfushman2016preparing}
Dina Demner-Fushman, Marc~D. Kohli, Marc~B. Rosenman, Sonya~E. Shooshan,
  Laritza Rodriguez, Sameer Antani, George~R. Thoma, and Clement~J. McDonald.
\newblock Preparing a collection of radiology examinations for distribution and
  retrieval.
\newblock \emph{Journal of the American Medical Informatics Association},
  23\penalty0 (2):\penalty0 304--310, 2016.

\bibitem[Devlin et~al.(2015)Devlin, Gupta, Girshick, Mitchell, and
  Zitnick]{devlin2015exploring}
Jacob Devlin, Saurabh Gupta, Ross Girshick, Margaret Mitchell, and C.~Lawrence
  Zitnick.
\newblock Exploring nearest neighbor approaches for image captioning.
\newblock \emph{arXiv preprint arXiv:1505.04467}, 2015.

\bibitem[Devlin et~al.(2019)Devlin, Chang, Lee, and Toutanova]{devlin2019bert}
Jacob Devlin, Ming-Wei Chang, Kenton Lee, and Kristina Toutanova.
\newblock {BERT}: Pre-training of deep bidirectional transformers for language
  understanding.
\newblock In \emph{Conference of the North American Chapter of the Association
  for Computational Linguistics}, 2019.

\bibitem[Dey et~al.(2019)]{dey2019legacy}
Arjun Dey et~al.
\newblock Overview of the {DESI} legacy imaging surveys.
\newblock \emph{The Astronomical Journal}, 157:\penalty0 168, 2019.

\bibitem[Dinh and Vo(2026)]{dinh2026paradox}
Anh Trac~Duc Dinh and Khang Nhat~Hoang Vo.
\newblock The {JEPA} paradox in language: The geometry of linguistic
  alternatives.
\newblock \emph{arXiv preprint arXiv:2607.23531}, 2026.

\bibitem[Doherty et~al.(2017)]{doherty2017ukb}
Aiden Doherty et~al.
\newblock Large scale population assessment of physical activity using wrist
  worn accelerometers: the {UK Biobank} study.
\newblock \emph{PLoS ONE}, 12\penalty0 (2):\penalty0 e0169649, 2017.

\bibitem[Edwards et~al.(2021)Edwards, Zhai, and Ji]{edwards2021text2mol}
Carl Edwards, ChengXiang Zhai, and Heng Ji.
\newblock {Text2Mol}: Cross-modal molecule retrieval with natural language
  queries.
\newblock In \emph{Conference on Empirical Methods in Natural Language
  Processing}, pages 595--607, 2021.

\bibitem[Edwards et~al.(2022)Edwards, Lai, Ros, Honke, Cho, and
  Ji]{edwards2022molt5}
Carl Edwards, Tuan Lai, Kevin Ros, Garrett Honke, Kyunghyun Cho, and Heng Ji.
\newblock Translation between molecules and natural language.
\newblock In \emph{Conference on Empirical Methods in Natural Language
  Processing}, pages 375--413, 2022.

\bibitem[Fai{\ss} et~al.(2025)]{faiss2025insectset459}
Marius Fai{\ss} et~al.
\newblock {InsectSet459}: an open dataset of insect sounds for bioacoustic
  machine learning.
\newblock Zenodo, record 18554693, v1.1, 2025.

\bibitem[Feng et~al.(2019)Feng, Ma, Liu, and Luo]{feng2019unsupervised}
Yang Feng, Lin Ma, Wei Liu, and Jiebo Luo.
\newblock Unsupervised image captioning.
\newblock In \emph{Conference on Computer Vision and Pattern Recognition},
  2019.

\bibitem[Gadre et~al.(2023)Gadre, Ilharco, Fang, Hayase, Smyrnis, Nguyen,
  et~al.]{gadre2023datacomp}
Samir~Yitzhak Gadre, Gabriel Ilharco, Alex Fang, Jonathan Hayase, Georgios
  Smyrnis, Thao Nguyen, et~al.
\newblock {DataComp}: In search of the next generation of multimodal datasets.
\newblock In \emph{Advances in Neural Information Processing Systems Datasets
  and Benchmarks Track}, 2023.

\bibitem[Gao et~al.(2021)Gao, Yao, and Chen]{gao2021simcse}
Tianyu Gao, Xingcheng Yao, and Danqi Chen.
\newblock {SimCSE}: Simple contrastive learning of sentence embeddings.
\newblock In \emph{Empirical Methods in Natural Language Processing}, 2021.

\bibitem[{Gemma Team}(2026)]{gemma2026gemma4}
{Gemma Team}.
\newblock {Gemma 4} technical report.
\newblock \emph{arXiv preprint arXiv:2607.02770}, 2026.

\bibitem[Gemmeke et~al.(2017)Gemmeke, Ellis, Freedman, Jansen, Lawrence, Moore,
  Plakal, and Ritter]{gemmeke2017audioset}
Jort~F. Gemmeke, Daniel P.~W. Ellis, Dylan Freedman, Aren Jansen, Wade
  Lawrence, R.~Channing Moore, Manoj Plakal, and Marvin Ritter.
\newblock Audio set: An ontology and human-labeled dataset for audio events.
\newblock In \emph{IEEE International Conference on Acoustics, Speech and
  Signal Processing}, pages 776--780, 2017.

\bibitem[Ghassemi et~al.(2018)]{ghassemi2018snooze}
Mohammad~M. Ghassemi et~al.
\newblock You snooze, you win: the {PhysioNet/Computing in Cardiology}
  challenge 2018.
\newblock In \emph{Computing in Cardiology (CinC)}, 2018.

\bibitem[Giorgi et~al.(2021)Giorgi, Nitski, Wang, and Bader]{giorgi2021declutr}
John Giorgi, Osvald Nitski, Bo~Wang, and Gary Bader.
\newblock {DeCLUTR}: Deep contrastive learning for unsupervised textual
  representations.
\newblock In \emph{Association for Computational Linguistics}, 2021.

\bibitem[Girdhar et~al.(2023)Girdhar, El-Nouby, Liu, Singh, Alwala, Joulin, and
  Misra]{girdhar2023imagebind}
Rohit Girdhar, Alaaeldin El-Nouby, Zhuang Liu, Mannat Singh, Kalyan~Vasudev
  Alwala, Armand Joulin, and Ishan Misra.
\newblock {ImageBind}: One embedding space to bind them all.
\newblock In \emph{Conference on Computer Vision and Pattern Recognition},
  2023.

\bibitem[Goyal et~al.(2017)Goyal, Khot, Summers-Stay, Batra, and
  Parikh]{goyal2017vqav2}
Yash Goyal, Tejas Khot, Douglas Summers-Stay, Dhruv Batra, and Devi Parikh.
\newblock Making the {V} in {VQA} matter: Elevating the role of image
  understanding in visual question answering.
\newblock In \emph{Conference on Computer Vision and Pattern Recognition
  (CVPR)}, 2017.

\bibitem[Gupta et~al.(2016)Gupta, Hoffman, and Malik]{gupta2016cross}
Saurabh Gupta, Judy Hoffman, and Jitendra Malik.
\newblock Cross modal distillation for supervision transfer.
\newblock In \emph{Conference on Computer Vision and Pattern Recognition},
  2016.

\bibitem[Hacohen et~al.(2022)Hacohen, Dekel, and
  Weinshall]{hacohen2022typiclust}
Guy Hacohen, Avihu Dekel, and Daphna Weinshall.
\newblock Active learning on a budget: Opposite strategies suit high and low
  budgets.
\newblock In \emph{International Conference on Machine Learning}, 2022.

\bibitem[Hagiwara(2023)]{hagiwara2023aves}
Masato Hagiwara.
\newblock {AVES}: Animal vocalization encoder based on self-supervision.
\newblock In \emph{IEEE International Conference on Acoustics, Speech and
  Signal Processing (ICASSP)}, 2023.

\bibitem[HaoChen et~al.(2021)HaoChen, Wei, Gaidon, and Ma]{haochen2021provable}
Jeff~Z. HaoChen, Colin Wei, Adrien Gaidon, and Tengyu Ma.
\newblock Provable guarantees for self-supervised deep learning with spectral
  contrastive loss.
\newblock In \emph{Advances in Neural Information Processing Systems}, 2021.

\bibitem[He et~al.(2022)He, Chen, Xie, Li, Doll{\'a}r, and
  Girshick]{he2022masked}
Kaiming He, Xinlei Chen, Saining Xie, Yanghao Li, Piotr Doll{\'a}r, and Ross
  Girshick.
\newblock Masked autoencoders are scalable vision learners.
\newblock In \emph{IEEE/CVF Conference on Computer Vision and Pattern
  Recognition}, 2022.

\bibitem[Hotelling(1936)]{hotelling1936relations}
Harold Hotelling.
\newblock Relations between two sets of variates.
\newblock \emph{Biometrika}, 28\penalty0 (3/4):\penalty0 321--377, 1936.

\bibitem[Hsu et~al.(2021)Hsu, Bolte, Tsai, Lakhotia, Salakhutdinov, and
  Mohamed]{hsu2021hubert}
Wei-Ning Hsu, Benjamin Bolte, Yao-Hung~Hubert Tsai, Kushal Lakhotia, Ruslan
  Salakhutdinov, and Abdelrahman Mohamed.
\newblock {HuBERT}: Self-supervised speech representation learning by masked
  prediction of hidden units.
\newblock \emph{IEEE/ACM Transactions on Audio, Speech, and Language
  Processing}, 29:\penalty0 3451--3460, 2021.

\bibitem[Huang et~al.(2025)Huang, LeCun, and Balestriero]{huang2025llmjepa}
Hai Huang, Yann LeCun, and Randall Balestriero.
\newblock {LLM-JEPA}: Large language models meet joint embedding predictive
  architectures.
\newblock \emph{arXiv preprint arXiv:2509.14252}, 2025.

\bibitem[Huang et~al.(2026)Huang, LeCun, and Balestriero]{huang2026stp}
Hai Huang, Yann LeCun, and Randall Balestriero.
\newblock Semantic tube prediction: Beating {LLM} data efficiency with {JEPA}.
\newblock \emph{arXiv preprint arXiv:2602.22617}, 2026.

\bibitem[Huh et~al.(2024)Huh, Cheung, Wang, and Isola]{huh2024platonic}
Minyoung Huh, Brian Cheung, Tongzhou Wang, and Phillip Isola.
\newblock Position: The platonic representation hypothesis.
\newblock In \emph{International Conference on Machine Learning}, 2024.

\bibitem[Jha et~al.(2025)Jha, Zhang, Shmatikov, and Morris]{jha2025vec2vec}
Rishi Jha, Collin Zhang, Vitaly Shmatikov, and John~X. Morris.
\newblock Harnessing the universal geometry of embeddings.
\newblock \emph{arXiv preprint arXiv:2505.12540}, 2025.

\bibitem[Karpathy and Fei-Fei(2015)]{karpathy2015deep}
Andrej Karpathy and Li~Fei-Fei.
\newblock Deep visual-semantic alignments for generating image descriptions.
\newblock In \emph{IEEE Conference on Computer Vision and Pattern Recognition},
  2015.

\bibitem[Kim et~al.(2019)Kim, Kim, Lee, and Kim]{kim2019audiocaps}
Chris~Dongjoo Kim, Byeongchang Kim, Hyunmin Lee, and Gunhee Kim.
\newblock {AudioCaps}: Generating captions for audios in the wild.
\newblock In \emph{Conference of the North American Chapter of the Association
  for Computational Linguistics: Human Language Technologies}, pages 119--132,
  2019.

\bibitem[Kumar and Byrne(2004)]{kumar2004mbr}
Shankar Kumar and William Byrne.
\newblock Minimum {Bayes}-risk decoding for statistical machine translation.
\newblock In \emph{Proceedings of the Human Language Technology Conference of
  the North American Chapter of the Association for Computational Linguistics},
  2004.

\bibitem[Kwon et~al.(2023)Kwon, Li, Zhuang, Sheng, Zheng, Yu, Gonzalez, Zhang,
  and Stoica]{kwon2023vllm}
Woosuk Kwon, Zhuohan Li, Siyuan Zhuang, Ying Sheng, Lianmin Zheng, Cody~Hao Yu,
  Joseph~E. Gonzalez, Hao Zhang, and Ion Stoica.
\newblock Efficient memory management for large language model serving with
  {PagedAttention}.
\newblock In \emph{Symposium on Operating Systems Principles}, 2023.

\bibitem[LeCun(2022)]{lecun2022path}
Yann LeCun.
\newblock A path towards autonomous machine intelligence.
\newblock OpenReview, 2022.

\bibitem[Li et~al.(2025)Li, Zhang, Guo, Zhang, Li, Zhang, Zhang, Zhang, Li,
  Liu, and Li]{li2025llavaonevision}
Bo~Li, Yuanhan Zhang, Dong Guo, Renrui Zhang, Feng Li, Hao Zhang, Kaichen
  Zhang, Peiyuan Zhang, Yanwei Li, Ziwei Liu, and Chunyuan Li.
\newblock {LLaVA-OneVision}: Easy visual task transfer.
\newblock \emph{Transactions on Machine Learning Research}, 2025.

\bibitem[Li et~al.(2023{\natexlab{a}})Li, Li, Savarese, and Hoi]{li2023blip2}
Junnan Li, Dongxu Li, Silvio Savarese, and Steven Hoi.
\newblock {BLIP-2}: Bootstrapping language-image pre-training with frozen image
  encoders and large language models.
\newblock In \emph{International Conference on Machine Learning (ICML)},
  2023{\natexlab{a}}.

\bibitem[Li(1995)]{li1995polar}
Ren-Cang Li.
\newblock New perturbation bounds for the unitary polar factor.
\newblock \emph{SIAM Journal on Matrix Analysis and Applications}, 16\penalty0
  (1):\penalty0 327--332, 1995.

\bibitem[Li et~al.(2023{\natexlab{b}})Li, Zhu, Wen, and Yang]{li2023decap}
Wei Li, Linchao Zhu, Longyin Wen, and Yi~Yang.
\newblock {DeCap}: Decoding {CLIP} latents for zero-shot captioning via
  text-only training.
\newblock In \emph{International Conference on Learning Representations},
  2023{\natexlab{b}}.

\bibitem[Li et~al.(2024)Li, Tu, Hui, Wang, Zhao, Xiao, Ren, Mei, Liu, Zheng,
  Zhou, and Xie]{li2024recap}
Xianhang Li, Haoqin Tu, Mude Hui, Zeyu Wang, Bingchen Zhao, Junfei Xiao,
  Sucheng Ren, Jieru Mei, Qing Liu, Huangjie Zheng, Yuyin Zhou, and Cihang Xie.
\newblock What if we recaption billions of web images with {LLaMA-3}?
\newblock \emph{arXiv preprint arXiv:2406.08478}, 2024.

\bibitem[Li et~al.(2023{\natexlab{c}})Li, Zhang, Zhang, Long, Xie, and
  Zhang]{li2023gte}
Zehan Li, Xin Zhang, Yanzhao Zhang, Dingkun Long, Pengjun Xie, and Meishan
  Zhang.
\newblock Towards general text embeddings with multi-stage contrastive
  learning.
\newblock \emph{arXiv preprint arXiv:2308.03281}, 2023{\natexlab{c}}.

\bibitem[Lin(2004)]{lin2004rouge}
Chin-Yew Lin.
\newblock {ROUGE}: A package for automatic evaluation of summaries.
\newblock In \emph{Text Summarization Branches Out}, pages 74--81, 2004.

\bibitem[Lin et~al.(2014)Lin, Maire, Belongie, Hays, Perona, Ramanan,
  Doll{\'a}r, and Zitnick]{lin2014coco}
Tsung-Yi Lin, Michael Maire, Serge Belongie, James Hays, Pietro Perona, Deva
  Ramanan, Piotr Doll{\'a}r, and C.~Lawrence Zitnick.
\newblock Microsoft {COCO}: Common objects in context.
\newblock In \emph{European Conference on Computer Vision}, pages 740--755,
  2014.

\bibitem[Lin et~al.(2023)Lin, Akin, Rao, Hie, Zhu, Lu, Smetanin, Verkuil,
  Kabeli, Shmueli, et~al.]{lin2023esm2}
Zeming Lin, Halil Akin, Roshan Rao, Brian Hie, Zhongkai Zhu, Wenting Lu, Nikita
  Smetanin, Robert Verkuil, Ori Kabeli, Yaniv Shmueli, et~al.
\newblock Evolutionary-scale prediction of atomic-level protein structure with
  a language model.
\newblock \emph{Science}, 379\penalty0 (6637):\penalty0 1123--1130, 2023.

\bibitem[Lu et~al.(2018)Lu, Wang, Zheng, and Li]{lu2018rsicd}
Xiaoqiang Lu, Binqiang Wang, Xiangtao Zheng, and Xuelong Li.
\newblock Exploring models and data for remote sensing image caption
  generation.
\newblock \emph{IEEE Transactions on Geoscience and Remote Sensing},
  56\penalty0 (4):\penalty0 2183--2195, 2018.

\bibitem[Maniparambil et~al.(2024)Maniparambil, Akshulakov, Djilali, Seddik,
  Narayan, Mangalam, and O'Connor]{maniparambil2024vision}
Mayug Maniparambil, Raiymbek Akshulakov, Yasser Abdelaziz~Dahou Djilali,
  Mohamed El~Amine Seddik, Sanath Narayan, Karttikeya Mangalam, and Noel~E.
  O'Connor.
\newblock Do vision and language encoders represent the world similarly?
\newblock In \emph{IEEE/CVF Conference on Computer Vision and Pattern
  Recognition}, 2024.

\bibitem[McCabe et~al.(2025)McCabe, Mukhopadhyay, Marwah,
  et~al.]{mccabe2025walrus}
Michael McCabe, Payel Mukhopadhyay, Tanya Marwah, et~al.
\newblock Walrus: A cross-domain foundation model for continuum dynamics.
\newblock \emph{arXiv preprint arXiv:2511.15684}, 2025.

\bibitem[Merullo et~al.(2023)Merullo, Castricato, Eickhoff, and
  Pavlick]{merullo2023linearly}
Jack Merullo, Louis Castricato, Carsten Eickhoff, and Ellie Pavlick.
\newblock Linearly mapping from image to text space.
\newblock In \emph{International Conference on Learning Representations}, 2023.

\bibitem[{Meta AI}(2025)]{meta2025peav}
{Meta AI}.
\newblock Perception encoder audio-visual.
\newblock \emph{arXiv preprint arXiv:2512.19687}, 2025.

\bibitem[Mokady et~al.(2021)Mokady, Hertz, and Bermano]{mokady2021clipcap}
Ron Mokady, Amir Hertz, and Amit~H. Bermano.
\newblock {ClipCap}: {CLIP} prefix for image captioning.
\newblock \emph{arXiv preprint arXiv:2111.09734}, 2021.

\bibitem[Moschella et~al.(2023)Moschella, Maiorca, Fumero, Norelli, Locatello,
  and Rodol{\`a}]{moschella2023relative}
Luca Moschella, Valentino Maiorca, Marco Fumero, Antonio Norelli, Francesco
  Locatello, and Emanuele Rodol{\`a}.
\newblock Relative representations enable zero-shot latent space communication.
\newblock In \emph{International Conference on Learning Representations}, 2023.

\bibitem[Norelli et~al.(2023)Norelli, Fumero, Maiorca, Moschella, Rodol{\`a},
  and Locatello]{norelli2023asif}
Antonio Norelli, Marco Fumero, Valentino Maiorca, Luca Moschella, Emanuele
  Rodol{\`a}, and Francesco Locatello.
\newblock {ASIF}: Coupled data turns unimodal models to multimodal without
  training.
\newblock In \emph{Advances in Neural Information Processing Systems}, 2023.

\bibitem[Nukrai et~al.(2022)Nukrai, Mokady, and Globerson]{nukrai2022capdec}
David Nukrai, Ron Mokady, and Amir Globerson.
\newblock Text-only training for image captioning using noise-injected {CLIP}.
\newblock In \emph{Findings of the Association for Computational Linguistics:
  EMNLP}, 2022.

\bibitem[Obeid and Picone(2016)]{obeid2016tuh}
Iyad Obeid and Joseph Picone.
\newblock The {Temple University Hospital} {EEG} data corpus.
\newblock \emph{Frontiers in Neuroscience}, 10:\penalty0 196, 2016.

\bibitem[Oh et~al.(2023)Oh, Lee, Bae, Kwon, and Choi]{oh2023ecgqa}
Jungwoo Oh, Gyubok Lee, Seongsu Bae, Joon-myoung Kwon, and Edward Choi.
\newblock {ECG-QA}: A comprehensive question answering dataset combined with
  electrocardiogram.
\newblock In \emph{Advances in Neural Information Processing Systems (NeurIPS),
  Datasets and Benchmarks Track}, 2023.

\bibitem[Ohana et~al.(2024)Ohana, McCabe, Meyer, Morel, et~al.]{ohana2024well}
Ruben Ohana, Michael McCabe, Lucas Meyer, Rudy Morel, et~al.
\newblock The {Well}: a large-scale collection of diverse physics simulations
  for machine learning.
\newblock In \emph{Advances in Neural Information Processing Systems Datasets
  and Benchmarks Track}, 2024.

\bibitem[{OpenAI}(2025)]{openai2025gptoss}
{OpenAI}.
\newblock gpt-oss-120b \& gpt-oss-20b model card.
\newblock \emph{arXiv preprint arXiv:2508.10925}, 2025.

\bibitem[Oquab et~al.(2024)Oquab, Darcet, Moutakanni, Vo, Szafraniec, Khalidov,
  Fernandez, Haziza, Massa, El-Nouby, et~al.]{oquab2024dinov2}
Maxime Oquab, Timoth{\'e}e Darcet, Th{\'e}o Moutakanni, Huy Vo, Marc
  Szafraniec, Vasil Khalidov, Pierre Fernandez, Daniel Haziza, Francisco Massa,
  Alaaeldin El-Nouby, et~al.
\newblock {DINOv2}: Learning robust visual features without supervision.
\newblock \emph{Transactions on Machine Learning Research}, 2024.

\bibitem[Papineni et~al.(2002)Papineni, Roukos, Ward, and
  Zhu]{papineni2002bleu}
Kishore Papineni, Salim Roukos, Todd Ward, and Wei-Jing Zhu.
\newblock {BLEU}: a method for automatic evaluation of machine translation.
\newblock In \emph{Annual Meeting of the Association for Computational
  Linguistics}, 2002.

\bibitem[P{\'e}rez-Garc{\'i}a et~al.(2025)P{\'e}rez-Garc{\'i}a, Sharma,
  Bond-Taylor, Bouzid, Salvatelli, Ilse, Bannur, Castro, Schwaighofer, Lungren,
  et~al.]{perezgarcia2025raddino}
Fernando P{\'e}rez-Garc{\'i}a, Harshita Sharma, Sam Bond-Taylor, Kenza Bouzid,
  Valentina Salvatelli, Maximilian Ilse, Shruthi Bannur, Daniel~C. Castro,
  Anton Schwaighofer, Matthew~P. Lungren, et~al.
\newblock Exploring scalable medical image encoders beyond text supervision.
\newblock \emph{Nature Machine Intelligence}, 7\penalty0 (1):\penalty0
  119--130, 2025.

\bibitem[Qu et~al.(2016)Qu, Li, Tao, and Lu]{qu2016deep}
Bo~Qu, Xuelong Li, Dacheng Tao, and Xiaoqiang Lu.
\newblock Deep semantic understanding of high resolution remote sensing image.
\newblock In \emph{International Conference on Computer, Information and
  Telecommunication Systems}, pages 1--5, 2016.

\bibitem[{Qwen Team}(2024)]{qwen2024qwen25}
{Qwen Team}.
\newblock Qwen2.5 technical report.
\newblock \emph{arXiv preprint arXiv:2412.15115}, 2024.

\bibitem[{Qwen Team}(2025)]{qwen2025qwen3omni}
{Qwen Team}.
\newblock Qwen3-{O}mni technical report.
\newblock \emph{arXiv preprint arXiv:2509.17765}, 2025.

\bibitem[{Qwen Team}(2026{\natexlab{a}})]{qwen2026qwen38}
{Qwen Team}.
\newblock {Qwen3.8-Max}: A new bar for coding and cowork.
\newblock \url{https://qwen.ai/blog?id=qwen3.8}, 2026{\natexlab{a}}.

\bibitem[{Qwen Team}(2026{\natexlab{b}})]{qwen2026qwen3vlemb}
{Qwen Team}.
\newblock {Qwen3-VL-Embedding} technical report.
\newblock \emph{arXiv preprint arXiv:2601.04720}, 2026{\natexlab{b}}.

\bibitem[Radford et~al.(2021)Radford, Kim, Hallacy, Ramesh, Goh, Agarwal,
  Sastry, Askell, Mishkin, Clark, Krueger, and Sutskever]{radford2021clip}
Alec Radford, Jong~Wook Kim, Chris Hallacy, Aditya Ramesh, Gabriel Goh,
  Sandhini Agarwal, Girish Sastry, Amanda Askell, Pamela Mishkin, Jack Clark,
  Gretchen Krueger, and Ilya Sutskever.
\newblock Learning transferable visual models from natural language
  supervision.
\newblock In \emph{International Conference on Machine Learning}, 2021.

\bibitem[Reed et~al.(2021)Reed, Metzger, Srinivas, Darrell, and
  Keutzer]{reed2021selfaugment}
Colorado~J Reed, Sean Metzger, Aravind Srinivas, Trevor Darrell, and Kurt
  Keutzer.
\newblock {SelfAugment}: Automatic augmentation policies for self-supervised
  learning.
\newblock In \emph{IEEE/CVF Conference on Computer Vision and Pattern
  Recognition}, 2021.

\bibitem[Reimers and Gurevych(2019)]{reimers2019sentencebert}
Nils Reimers and Iryna Gurevych.
\newblock Sentence-{BERT}: Sentence embeddings using {Siamese} {BERT}-networks.
\newblock In \emph{Conference on Empirical Methods in Natural Language
  Processing}, 2019.

\bibitem[Schnaus et~al.(2025)Schnaus, Araslanov, and Cremers]{schnaus2025blind}
Dominik Schnaus, Nikita Araslanov, and Daniel Cremers.
\newblock It's a (blind) match! {T}owards vision-language correspondence
  without parallel data.
\newblock In \emph{Conference on Computer Vision and Pattern Recognition},
  2025.

\bibitem[Schnaus et~al.(2026)Schnaus, Dag{\`e}s, Cremers, Wang, and
  Isola]{schnaus2026rosetta}
Dominik Schnaus, Thomas Dag{\`e}s, Daniel Cremers, Xi~Wang, and Phillip Isola.
\newblock Shared geometry as a rosetta stone: Cross-modal alignment without
  paired data.
\newblock \emph{arXiv preprint arXiv:2610.09411}, 2026.

\bibitem[Sch{\"o}nemann(1966)]{schonemann1966generalized}
Peter~H. Sch{\"o}nemann.
\newblock A generalized solution of the orthogonal {Procrustes} problem.
\newblock \emph{Psychometrika}, 31\penalty0 (1):\penalty0 1--10, 1966.

\bibitem[Shao et~al.(2023)Shao, Yu, Wang, and Yu]{shao2023prompting}
Zhenwei Shao, Zhou Yu, Meng Wang, and Jun Yu.
\newblock Prompting large language models with answer heuristics for
  knowledge-based visual question answering.
\newblock In \emph{Proceedings of the IEEE/CVF Conference on Computer Vision
  and Pattern Recognition (CVPR)}, 2023.

\bibitem[Sim{\'e}oni et~al.(2025)Sim{\'e}oni, Vo, Seitzer, Baldassarre, Oquab,
  Jose, Khalidov, Szafraniec, Yi, Ramamonjisoa, et~al.]{simeoni2025dinov3}
Oriane Sim{\'e}oni, Huy~V. Vo, Maximilian Seitzer, Federico Baldassarre, Maxime
  Oquab, Cijo Jose, Vasil Khalidov, Marc Szafraniec, Seungeun Yi, Micha{\"e}l
  Ramamonjisoa, et~al.
\newblock {DINOv3}.
\newblock \emph{arXiv preprint arXiv:2508.10104}, 2025.

\bibitem[Smit et~al.(2020)Smit, Jain, Rajpurkar, Pareek, Ng, and
  Lungren]{smit2020chexbert}
Akshay Smit, Saahil Jain, Pranav Rajpurkar, Anuj Pareek, Andrew~Y. Ng, and
  Matthew~P. Lungren.
\newblock Combining automatic labelers and expert annotations for accurate
  radiology report labeling using {BERT}.
\newblock In \emph{Proceedings of the 2020 Conference on Empirical Methods in
  Natural Language Processing (EMNLP)}, 2020.

\bibitem[Smith et~al.(2024)Smith, Roberts, Angeloudi, and
  Huertas-Company]{smith2024astropt}
Michael~J. Smith, Ryan~J. Roberts, Eirini Angeloudi, and Marc Huertas-Company.
\newblock {AstroPT}: Scaling large observation models for astronomy.
\newblock \emph{arXiv preprint arXiv:2405.14930}, 2024.

\bibitem[Song et~al.(2020)Song, Tan, Qin, Lu, and Liu]{song2020mpnet}
Kaitao Song, Xu~Tan, Tao Qin, Jianfeng Lu, and Tie-Yan Liu.
\newblock {MPNet}: Masked and permuted pre-training for language understanding.
\newblock In \emph{Advances in Neural Information Processing Systems}, 2020.

\bibitem[Srinivasan et~al.(2021)Srinivasan, Raman, Chen, Bendersky, and
  Najork]{srinivasan2021wit}
Krishna Srinivasan, Karthik Raman, Jiecao Chen, Michael Bendersky, and Marc
  Najork.
\newblock {WIT}: Wikipedia-based image text dataset for multimodal multilingual
  machine learning.
\newblock In \emph{International ACM SIGIR Conference on Research and
  Development in Information Retrieval}, 2021.

\bibitem[Suzek et~al.(2015)Suzek, Wang, Huang, McGarvey, Wu, and {the UniProt
  Consortium}]{suzek2015uniref}
Baris~E. Suzek, Yuqi Wang, Hongzhan Huang, Peter~B. McGarvey, Cathy~H. Wu, and
  {the UniProt Consortium}.
\newblock {UniRef} clusters: a comprehensive and scalable alternative for
  improving sequence similarity searches.
\newblock \emph{Bioinformatics}, 31\penalty0 (6):\penalty0 926--932, 2015.

\bibitem[Tamkin et~al.(2021)Tamkin, Wu, and Goodman]{tamkin2021viewmaker}
Alex Tamkin, Mike Wu, and Noah Goodman.
\newblock Viewmaker networks: Learning views for unsupervised representation
  learning.
\newblock In \emph{International Conference on Learning Representations}, 2021.

\bibitem[{The UniProt Consortium}(2023)]{uniprot2023uniprot}
{The UniProt Consortium}.
\newblock {UniProt}: the universal protein knowledgebase in 2023.
\newblock \emph{Nucleic Acids Research}, 51\penalty0 (D1):\penalty0 D523--D531,
  2023.

\bibitem[Tian et~al.(2020)Tian, Sun, Poole, Krishnan, Schmid, and
  Isola]{tian2020infomin}
Yonglong Tian, Chen Sun, Ben Poole, Dilip Krishnan, Cordelia Schmid, and
  Phillip Isola.
\newblock What makes for good views for contrastive learning?
\newblock In \emph{Advances in Neural Information Processing Systems}, 2020.

\bibitem[Tosh et~al.(2021)Tosh, Krishnamurthy, and Hsu]{tosh2021contrastive}
Christopher Tosh, Akshay Krishnamurthy, and Daniel Hsu.
\newblock Contrastive learning, multi-view redundancy, and linear models.
\newblock In \emph{International Conference on Algorithmic Learning Theory},
  2021.

\bibitem[Tschannen et~al.(2025)]{tschannen2025siglip2}
Michael Tschannen et~al.
\newblock {SigLIP} 2: Multilingual vision-language encoders with improved
  semantic understanding, localization, and dense features.
\newblock \emph{arXiv preprint arXiv:2502.14786}, 2025.

\bibitem[Tsimpoukelli et~al.(2021)Tsimpoukelli, Menick, Cabi, Eslami, Vinyals,
  and Hill]{tsimpoukelli2021frozen}
Maria Tsimpoukelli, Jacob Menick, Serkan Cabi, S.~M.~Ali Eslami, Oriol Vinyals,
  and Felix Hill.
\newblock Multimodal few-shot learning with frozen language models.
\newblock In \emph{Advances in Neural Information Processing Systems}, 2021.

\bibitem[Vedantam et~al.(2015)Vedantam, Zitnick, and Parikh]{vedantam2015cider}
Ramakrishna Vedantam, C.~Lawrence Zitnick, and Devi Parikh.
\newblock {CIDEr}: Consensus-based image description evaluation.
\newblock In \emph{IEEE Conference on Computer Vision and Pattern Recognition},
  2015.

\bibitem[Walmsley et~al.(2023)]{walmsley2023gzdesi}
Mike Walmsley et~al.
\newblock {Galaxy Zoo DESI}: Detailed morphology measurements for 8.7{M}
  galaxies in the {DESI} legacy imaging surveys.
\newblock \emph{Monthly Notices of the Royal Astronomical Society},
  526:\penalty0 4768--4786, 2023.

\bibitem[Wang et~al.(2025{\natexlab{a}})]{wang2025cbramod}
Jiquan Wang et~al.
\newblock {CBraMod}: A criss-cross brain foundation model for {EEG} decoding.
\newblock In \emph{International Conference on Learning Representations
  (ICLR)}, 2025{\natexlab{a}}.

\bibitem[Wang et~al.(2025{\natexlab{b}})Wang, Chen, Zeghidour, and
  Saeed]{wang2025careaqa}
Tsai-Ning Wang, Lin-Lin Chen, Neil Zeghidour, and Aaqib Saeed.
\newblock {CaReAQA}: A cardiac and respiratory audio question answering model
  for open-ended diagnostic reasoning.
\newblock In \emph{Proceedings of the Conference on Health, Inference, and
  Learning}, volume 287 of \emph{PMLR}, 2025{\natexlab{b}}.

\bibitem[Wang et~al.(2017)Wang, Peng, Lu, Lu, Bagheri, and
  Summers]{wang2017chestxray}
Xiaosong Wang, Yifan Peng, Le~Lu, Zhiyong Lu, Mohammadhadi Bagheri, and
  Ronald~M. Summers.
\newblock {ChestX-ray8}: Hospital-scale chest {X}-ray database and benchmarks
  on weakly-supervised classification and localization of common thorax
  diseases.
\newblock In \emph{IEEE Conference on Computer Vision and Pattern Recognition},
  pages 3462--3471, 2017.

\bibitem[Xu et~al.(2016)Xu, Mei, Yao, and Rui]{xu2016msrvtt}
Jun Xu, Tao Mei, Ting Yao, and Yong Rui.
\newblock {MSR-VTT}: A large video description dataset for bridging video and
  language.
\newblock In \emph{IEEE Conference on Computer Vision and Pattern Recognition},
  2016.

\bibitem[Yang et~al.(2024)Yang, Wang, Lu, Liu, Le, Zhou, and
  Chen]{yang2024opro}
Chengrun Yang, Xuezhi Wang, Yifeng Lu, Hanxiao Liu, Quoc~V Le, Denny Zhou, and
  Xinyun Chen.
\newblock Large language models as optimizers.
\newblock In \emph{International Conference on Learning Representations}, 2024.

\bibitem[Yang et~al.(2022)Yang, Gan, Wang, Hu, Lu, Liu, and
  Wang]{yang2022empirical}
Zhengyuan Yang, Zhe Gan, Jianfeng Wang, Xiaowei Hu, Yumao Lu, Zicheng Liu, and
  Lijuan Wang.
\newblock An empirical study of {GPT-3} for few-shot knowledge-based {VQA}.
\newblock In \emph{Proceedings of the AAAI Conference on Artificial
  Intelligence}, 2022.

\bibitem[Young et~al.(2014)Young, Lai, Hodosh, and
  Hockenmaier]{young2014flickr}
Peter Young, Alice Lai, Micah Hodosh, and Julia Hockenmaier.
\newblock From image descriptions to visual denotations: New similarity metrics
  for semantic inference over event descriptions.
\newblock \emph{Transactions of the Association for Computational Linguistics},
  2:\penalty0 67--78, 2014.

\bibitem[Yuan et~al.(2024)Yuan, Chan, others, and
  Doherty]{yuan2024sslwearables}
Hang Yuan, Shing Chan, others, and Aiden Doherty.
\newblock Self-supervised learning for human activity recognition using 700,000
  person-days of wearable data.
\newblock \emph{npj Digital Medicine}, 7:\penalty0 91, 2024.

\bibitem[Zhai et~al.(2022)Zhai, Wang, Mustafa, Steiner, Keysers, Kolesnikov,
  and Beyer]{zhai2022lit}
Xiaohua Zhai, Xiao Wang, Basil Mustafa, Andreas Steiner, Daniel Keysers,
  Alexander Kolesnikov, and Lucas Beyer.
\newblock {LiT}: Zero-shot transfer with locked-image text tuning.
\newblock In \emph{Conference on Computer Vision and Pattern Recognition},
  2022.

\bibitem[Zhang et~al.(2023)Zhang, Lan, and He]{zhang2023syncse}
Junlei Zhang, Zhenzhong Lan, and Junxian He.
\newblock Contrastive learning of sentence embeddings from scratch.
\newblock In \emph{Empirical Methods in Natural Language Processing}, 2023.

\bibitem[Zhang et~al.(2022{\natexlab{a}})]{zhang2022sprsound}
Qing Zhang et~al.
\newblock {SPRSound}: Open-source {SJTU} paediatric respiratory sound database.
\newblock \emph{IEEE Transactions on Biomedical Circuits and Systems},
  2022{\natexlab{a}}.

\bibitem[Zhang et~al.(2022{\natexlab{b}})Zhang, Roller, Goyal, Artetxe, Chen,
  et~al.]{zhang2022opt}
Susan Zhang, Stephen Roller, Naman Goyal, Mikel Artetxe, Moya Chen, et~al.
\newblock {OPT}: Open pre-trained transformer language models.
\newblock \emph{arXiv preprint arXiv:2205.01068}, 2022{\natexlab{b}}.

\bibitem[Zhang et~al.(2025{\natexlab{a}})Zhang, Acosta, Miller, Huang, and
  Rajpurkar]{zhang2025rexgradient}
Xiaoman Zhang, Juli{\'a}n~N. Acosta, Josh Miller, Ouwen Huang, and Pranav
  Rajpurkar.
\newblock {ReXGradient-160K}: A large-scale publicly available dataset of chest
  radiographs with free-text reports.
\newblock \emph{arXiv preprint arXiv:2505.00228}, 2025{\natexlab{a}}.

\bibitem[Zhang et~al.(2025{\natexlab{b}})Zhang, Li, Long, Zhang, Lin, Yang,
  Xie, Yang, Liu, Lin, Huang, and Zhou]{zhang2025qwen3embedding}
Yanzhao Zhang, Mingxin Li, Dingkun Long, Xin Zhang, Huan Lin, Baosong Yang,
  Pengjun Xie, An~Yang, Dayiheng Liu, Junyang Lin, Fei Huang, and Jingren Zhou.
\newblock {Qwen3 Embedding}: Advancing text embedding and reranking through
  foundation models.
\newblock \emph{arXiv preprint arXiv:2506.05176}, 2025{\natexlab{b}}.

\bibitem[Zhu et~al.(2025)Zhu, Wang, Chen, Liu, Ye, et~al.]{zhu2025internvl3}
Jinguo Zhu, Weiyun Wang, Zhe Chen, Zhaoyang Liu, Shenglong Ye, et~al.
\newblock {InternVL3}: Exploring advanced training and test-time recipes for
  open-source multimodal models.
\newblock \emph{arXiv preprint arXiv:2504.10479}, 2025.

\end{thebibliography}

\clearpage
\appendixparttrue
\appendix
\crefalias{section}{appendix}%
\crefalias{subsection}{subappendix}%
\onecolumn
\setlength{\textfloatsep}{20pt plus 2pt minus 4pt}\setlength{\dbltextfloatsep}{20pt plus 2pt minus 4pt}%
\titleformat{\section}[block]{\sffamily\bfseries\Large\raggedright}{\colorbox{rule!45}{\strut\hspace{.25em}\thesection\hspace{.25em}}}{0.7em}{}[\vspace{-2pt}{\color{rule}\titlerule[0.8pt]}]
\begin{center}
{\sffamily\bfseries\Huge Appendix}\par\vspace{6pt}
{\color{rule}\rule{0.6\linewidth}{1.2pt}}\par\vspace{8pt}
\begin{minipage}{0.86\linewidth}\small
In this appendix, we first state every result of \cref{sec:theory} in full and prove it. We then detail the view search and every experiment of the main text. Next, we describe when each method leads, and we discuss the limitations and further related work. Finally, we report every number of the main text, the domains and their Platonic views.
\end{minipage}
\end{center}
\vspace{10pt}
\startcontents[appendix]
\printcontents[appendix]{}{1}{\setcounter{tocdepth}{2}}
\clearpage

\section{Proofs}\label{app:proofs}

Each result of \cref{sec:theory} is restated here in full, with every condition and constant, before its proof. Throughout, (A1)--(A5) hold where a statement assumes them, $r=\min(k,s)$, $\bar k=\min(k,d)$, and functions of $x_1$ carry the inner product $\langle\psi,\psi'\rangle=\E[\psi(x_1)\psi'(x_1)]$, so that a centered, whitened encoder is an orthonormal family of centered functions.

\subsection{\texorpdfstring{Proof of \Cref{thm:ceiling}}{Proof of the ceiling theorem}}\label{prf:ceiling}

\begin{fullstatement}{\Cref{thm:ceiling}}
Assume (A5). (i) For every centered, whitened $f:\X_1\to\R^k$, however it is trained and on however many pairs, with $\mM_f=\E[f(x_1)g(x_2)^\top]$,
\[
\max_{\mQ\in\mathrm O(k,d)}\E\langle f(x_1),\mQ g(x_2)\rangle=\|\mM_f\|_*\le k\rho,
\]
i.e., the misalignment is at least $k+\min(k,d)-2k\rho$ ($=2k(1-\rho)$ if $k=d$), with equality \emph{if and only if} the span of $f$ contains $h_{1:r}$ (for some eigenbasis $\mU$, if eigenvalues tie). If $k\le s$, and $\Lambda_{kk}>\Lambda_{k+1,k+1}$ when $k<s$, the only such $f$ are $\mQ h_{1:k}$, $\mQ\in\Oset$.\par
(ii) In the world model, $b=a_g$, so $\mC=\mC_g$, and for every rotation $\mQ$
\begin{equation*}
\begin{split}
\E\|f(x_1)-\mQ g(x_2)\|^2={}&N(f)+N(\mQ g)\\
&+\E\|a_f(z)-\mQ a_g(z)\|^2,
\end{split}
\end{equation*}
with $N(\mQ g)=N(g)$ if $k\ge d$. The misalignment is zero if and only if $f(x_1)=\mQ g(x_2)=\mQ a_g(z)$: both encode the same function of the state, and nothing else.
\end{fullstatement}

\paragraph{Proof.} (i) For $\mQ\in\mathrm O(k,d)$, $\mQ^\top\mQ$ is a projection of rank $\bar k$ and $g$ is whitened, so $\E\|\mQ g\|^2=\bar k$ and the misalignment is $k+\bar k-2\,\mathrm{tr}(\mQ^\top\mM_f)$. By von Neumann's trace inequality, $\mathrm{tr}(\mQ^\top\mM_f)\le\sum_i\sigma_i(\mQ)\sigma_i(\mM_f)=\|\mM_f\|_*$, since $\mQ$ has $\bar k\ge\mathrm{rank}\,\mM_f$ unit singular values, with equality at $\mQ=\boldsymbol U_{\!f}\boldsymbol V_{\!f}^\top$ for a thin singular value decomposition $\mM_f=\boldsymbol U_{\!f}\boldsymbol\Sigma_f\boldsymbol V_{\!f}^\top$ with $\bar k$ columns \citep{schonemann1966generalized}. Conditioning on $x_1$, $\mM_f=\E[f b^\top]$; since $b=\mU\boldsymbol\Lambda^{1/2}h$ almost surely, $\mM_f=\boldsymbol H\boldsymbol\Lambda^{1/2}\mU^\top$ with $\boldsymbol H=\E[fh^\top]\in\R^{k\times s}$, and $\|\mM_f\|_*=\|\boldsymbol H\boldsymbol\Lambda^{1/2}\|_*$ as $\mU$ has orthonormal columns. The entries of $\boldsymbol H$ are the coefficients of each $h_i$ on the orthonormal coordinates of $f$, so $\boldsymbol G=\boldsymbol H^\top\boldsymbol H$ is the Gram matrix of the projections of $h_1,\dots,h_s$ on the span of $f$, and $\boldsymbol G\preceq\mI_s$ by Bessel's inequality. Let $\mK=\boldsymbol\Lambda^{1/2}\boldsymbol G\boldsymbol\Lambda^{1/2}\preceq\boldsymbol\Lambda$, of rank at most $r$, with eigenvalues $\kappa_1\ge\kappa_2\ge\dots$. By Weyl's monotonicity $\kappa_i\le\Lambda_{ii}$, so $\|\boldsymbol H\boldsymbol\Lambda^{1/2}\|_*=\sum_{i\le r}\kappa_i^{1/2}\le\sum_{i\le r}\Lambda_{ii}^{1/2}=k\rho$, with equality iff $\kappa_i=\Lambda_{ii}$ for every $i\le r$. If $h_1,\dots,h_r$ lie in the span of $f$, then $\boldsymbol G_{ij}=\langle h_i,h_j\rangle=\delta_{ij}$ for $i\le r$ (the projection is self-adjoint and fixes $h_i$), so $\boldsymbol G=\mathrm{diag}(\mI_r,\boldsymbol G')$ with $\boldsymbol G'\succeq0$, $\mK\succeq\mathrm{diag}(\Lambda_{11},\dots,\Lambda_{rr},0,\dots,0)$, and $\kappa_i\ge\Lambda_{ii}$ for $i\le r$: equality holds. Conversely, let $\kappa_i=\Lambda_{ii}$ for $i\le r$ and let $\mY\in\R^{s\times r}$ hold orthonormal eigenvectors of $\mK$ for $\kappa_1,\dots,\kappa_r$. By Ky Fan's maximum principle, $\sum_{i\le r}\Lambda_{ii}\ge\mathrm{tr}(\mY^\top\boldsymbol\Lambda\mY)=\sum_{i\le r}\kappa_i+\mathrm{tr}\big(\mY^\top(\boldsymbol\Lambda-\mK)\mY\big)$, so $(\boldsymbol\Lambda-\mK)\mY=0$ and $\boldsymbol\Lambda\mY=\mK\mY=\mY\,\mathrm{diag}(\Lambda_{11},\dots,\Lambda_{rr})$: the columns of $\mY$ span an invariant subspace of the diagonal $\boldsymbol\Lambda$ that carries its top $r$ eigenvalues. Choosing the eigenbasis $\mU$ within tied eigenvalues, which changes neither $\mC$, $\boldsymbol\Lambda$ nor $\rho$, makes that subspace the span of the first $r$ coordinate vectors, each column of $\mY$ staying an eigenvector of $\boldsymbol\Lambda$ for its $\Lambda_{ii}$; then $\mK e_i=\Lambda_{ii}e_i$, so $\boldsymbol G_{ii}=\mK_{ii}/\Lambda_{ii}=1$ for $i\le r$: the projection of each $h_i$, $i\le r$, has unit norm, so $h_i$ lies in the span of $f$. A maximizer exists: $h_{1:r}$ completed by $k-r$ centered functions orthonormal to $h$, which exist when $|\X_1|>k$. If $k\le s$ and $h_{1:k}$ is unique ($k=s$, or $\Lambda_{kk}>\Lambda_{k+1,k+1}$), the span of $f$ contains the $k$-dimensional span of $h_{1:k}$, so they coincide, and two orthonormal bases of one space differ by an orthogonal matrix: $f=\mQ h_{1:k}$, $\mQ\in\Oset$.\\
(ii) In the world model, $z$ is a function of $x_1$ (injective rendering) and $x_2$ is independent of $x_1$ given $z$, so $b(x_1)=\E[g(x_2)\mid z]=a_g(z)$. Write $f=a_f+\eta_f$ and $g=a_g+\eta_g$ with $\E[\eta_f\mid z]=\E[\eta_g\mid z]=0$. Expanding $\E\|(a_f-\mQ a_g)+(\eta_f-\mQ\eta_g)\|^2$, the cross terms vanish: $\E\langle a_f-\mQ a_g,\eta_f\rangle=\E\langle a_f-\mQ a_g,\E[\eta_f\mid z]\rangle=0$, likewise with $\eta_g$, and $\E\langle\eta_f,\mQ\eta_g\rangle=\E\langle\E[\eta_f\mid z],\mQ\E[\eta_g\mid z]\rangle=0$ by conditional independence. As $a_{\mQ g}=\mQ a_g$, $\E\|\mQ\eta_g\|^2=N(\mQ g)$, which equals $N(g)$ if $k\ge d$ ($\mQ^\top\mQ=\mI_d$); this gives \eqref{eq:split}. All three terms are non-negative, so the misalignment vanishes iff $\eta_f=0$, $\mQ\eta_g=0$ and $a_f=\mQ a_g$, i.e., $f(x_1)=\mQ a_g(z)=\mQ g(x_2)$.

\subsection{\texorpdfstring{Proof of \Cref{thm:every}}{Proof that every anchor has Platonic views}}\label{prf:every}

\begin{fullstatement}{\Cref{thm:every}}
Let $e$ be $k-r$ centered, whitened functions of $x_1$ orthogonal to $h$ (none if $k\le s$), and $\omega_g(x,x')=\lambda\,h_{1:r}(x)^\top h_{1:r}(x')+\lambda'e(x)^\top e(x')$, with $\lambda,\lambda'>0$ ($\lambda'=0$ if $k\le s$) and $\lambda\max\|h_{1:r}\|^2+\lambda'\max\|e\|^2\le1$. Assume (A1)--(A5) and $|\X_1|>k$. (i) The view law $W_g(x,x')=P(x)P(x')\big(1+\omega_g(x,x')\big)$ is Platonic: its JEPA spans $h_{1:r}$ and $e$.\par
(ii) If $k\le s$, and $\Lambda_{kk}>\Lambda_{k+1,k+1}$ when $k<s$, every Platonic JEPA is $\mQ h_{1:k}$ for a rotation $\mQ\in\Oset$ ($=\mQ'\mC^{-1/2}b$ if $k=s=d$): all Platonic views train one JEPA.\par
(iii) In the world model, with $e$ a function of the state (possible if $|\Z|>k$), $W_g$ is an imagination (\cref{prop:imagine}). If the anchor is a good-view JEPA, $g=\mQ\vPhi_\pi$ with $d$ outputs, good views of $\pi$ are Platonic views, and the Platonic JEPAs are $\mQ'\vPhi_\pi$ if $k=d$.
\end{fullstatement}

\paragraph{Proof.} Take $h_{1:r}$ and $e$ as in the statement, centered and orthonormal; $e$ exists when $k>s$, since the centered functions of $x_1$ have dimension $|\X_1|-1\ge k$. (i) By Cauchy--Schwarz, $|h_{1:r}(x)^\top h_{1:r}(x')|\le\max\|h_{1:r}\|^2$ and likewise for $e$, so $W_g\ge0$; it is symmetric, and $\sum_{x'}W_g(x,x')=P(x)\big(1+\lambda h_{1:r}(x)^\top\E h_{1:r}+\lambda'e(x)^\top\E e\big)=P(x)$ since $h$ and $e$ are centered: (A2) holds. Its Markov operator is $T\psi=\E\psi+\lambda h_{1:r}^\top\E[h_{1:r}\psi]+\lambda'e^\top\E[e\psi]$: $T1=1$, $Th_i=\lambda h_i$ for $i\le r$, $Te_j=\lambda'e_j$, and $T\psi=0$ for $\psi$ orthogonal to $1,h_{1:r},e$ (including $h_{r+1},\dots,h_s$ if $k<s$). The $k$-th largest eigenvalue after the constant, $\min(\lambda,\lambda')$ ($\lambda$ if $k\le s$), is positive and exceeds the next, $0$ (if any), so (A3) holds and the JEPA is $\mR\,(h_{1:r},e)$ for a rotation $\mR\in\Oset$, whose span contains $h_{1:r}$: it reaches the ceiling by \cref{thm:ceiling}(i). (ii) By the uniqueness part of \cref{thm:ceiling}(i), $f=\mQ h_{1:k}=\mQ\boldsymbol\Lambda_k^{-1/2}\mU_k^\top b$, with $\boldsymbol\Lambda_k$ and $\mU_k$ the top $k$ eigenvalues and eigenvectors of $\mC$; if $k=s=d$, $\mU$ is orthogonal and $\boldsymbol\Lambda^{-1/2}\mU^\top=\mU^\top\mC^{-1/2}$. (iii) In the world model $b=a_g$ depends on $x_1=\gamma_1(z,\nu)$ through $z$ alone, so $h$ is a function of the state, and $e$ can be taken so, as the centered functions of the state have dimension $|\mathcal Z|-1\ge k$. Then $P(x)=p(z)q_1(\nu)$ and $W_g(x,x')=\pi_g(z,z')\,q_1(\nu)q_1(\nu')$ with $\pi_g(z,z')=p(z)p(z')\big(1+\lambda h_{1:r}(z)^\top h_{1:r}(z')+\lambda'e(z)^\top e(z')\big)$, a symmetric kernel with marginal $p$: the views of the imagination $\kappa(z'\mid z)=\pi_g(z,z')/p(z)$ of \cref{prop:imagine}. If $g=\mQ\vPhi_\pi$, with $\vPhi_\pi$ the top $d$ eigenfunctions of $\pi$ (a gap after the $d$-th, as the anchor's own training requires), then $g$ is a function of the state, so $b=g$, $\mC=\mI_d$, $s=d$, and, all eigenvalues being tied, every orthonormal basis of the span of $\vPhi_\pi$ is a valid $h$. By \cref{thm:ceiling}(i), $f$ is Platonic iff its span contains $\bar k$ orthonormal functions of that span: it contains the span of $\vPhi_\pi$ if $k\ge d$, and lies in it if $k\le d$. The good views of $\pi$ have the eigenvalues of $\pi$ on functions of the state and $0$ on the rest (fresh nuisances), so under (A3) they train the top $k$ eigenfunctions of $\pi$ \citep[Prop.~2]{balestriero2026prh}, and the top-$k$ and top-$d$ eigenspaces of $\pi$, both defined by their gaps, are nested: good views of $\pi$ are Platonic. If $k=d$, the Platonic JEPAs are $\mQ'\vPhi_\pi$ by (ii).

\subsection{\texorpdfstring{Proof of \Cref{thm:search}}{Proof that Platonic views maximize the reward}}\label{prf:search}

\begin{fullstatement}{\Cref{thm:search}}
Assume (A1)--(A5). (i) $R(W)\le k\rho$ for every view law $W$, with equality \emph{if and only if} $f_W$ spans $h_{1:r}$, i.e., is a Platonic JEPA.\par
(ii) For a Platonic JEPA $f^\star$, the least-squares linear map from $f^\star$ to $g$, $\mM^\top=\E[gf^{\star\top}]$, gives $\mM^\top f^\star(x_1)=\mU_r\mU_r^\top\E[g(x_2)\mid x_1]$, which is $\E[g(x_2)\mid x_1]$ if $k\ge s$. Conversely, every centered, whitened $f$ has $\E\|g(x_2)-\mM_f^\top f(x_1)\|^2\ge d-\sum_{i\le r}\Lambda_{ii}$, which is $\mathrm{tr}(\mI-\mC)=\E\|g(x_2)-\E[g(x_2)\mid x_1]\|^2$ if $k\ge s$, with equality \emph{if and only if} the span of $f$ contains $h_{1:r}$. If $k=d$, $\min_{\mQ\in\mathrm O(d)}\E\|g(x_2)-\mQ^\top f(x_1)\|^2\ge\mathrm{tr}(\mI-\mC)+\mathrm{tr}\big((\mI-\mC^{1/2})^2\big)$, with equality iff the span of $f$ contains $h_{1:s}$, so a rotation predicts the anchor as well as any function of $x_1$ \emph{if and only if}, in addition, $\mC=\mI$, i.e., $g(x_2)$ is almost surely a function of $x_1$ ($N(g)=0$ in the world model). For a Platonic JEPA, $\mM^\top=(\mM^\top\mM)^{1/2}\mQ^{\star\top}$ for every polar factor $\mQ^\star$ of $\mM$; these rotations maximize the alignment and agree on the range of $\mM^\top\mM$. The reverse map $\mM g$ is only the best linear predictor of $f^\star$ from $g$.\par
(iii) If $k=d=s$, every polar factor $\hat\mQ$ of $\hat\mM$ has $\|\hat\mQ-\mQ^\star\|_F\le2\|\hat\mM-\mM\|_F/\Lambda_{kk}^{1/2}$, so $\E\|\hat\mQ-\mQ^\star\|_F\le2\sqrt{V/(n\Lambda_{kk})}$, informative only for $n\gg V/(k\Lambda_{kk})\approx k/\Lambda_{kk}$, as $\|\hat\mQ-\mQ^\star\|_F\le2\sqrt k$ and $V\approx k^2$.\par
(iv) $\E\big(\hat R_n(W)-R(W)\big)^2\le\bar kV_W/n$, $\bar k=\min(k,d)$. For any candidates $W_1,\dots,W_m$ and $\epsilon>0$, the maximizer $\hat W$ of $\hat R_n$ has $R(\hat W)>\max_iR(W_i)-\epsilon$ with probability at least $1-4\bar k\sum_iV_{W_i}/(n\epsilon^2)$; if one is Platonic and every non-Platonic one trails it by at least $\Delta>0$, $\epsilon=\Delta$ makes $\hat W$ Platonic.
\end{fullstatement}

\paragraph{Proof.} (i) By (A4), $f_W$ is centered, whitened and unique up to a rotation, which leaves $\|\mM_W\|_*$ unchanged; by \cref{thm:ceiling}(i), $R(W)=\|\mM_W\|_*\le k\rho$, with equality iff the span of $f_W$ contains $h_{1:r}$, which by definition makes $f_W$ a Platonic JEPA.\\
(ii) A Platonic $f^\star$ is whitened and its span contains $h_{1:r}$, so $f^\star=\mR\,(h_{1:r};e)$ with $\mR\in\Oset$ and $e$ ($k-r$ functions, none if $k\le s$) orthonormal and orthogonal to $h$ (if $k>s$, $r=s$). Then $\mM=\E[f^\star b^\top]=\mR\,\E[(h_{1:r};e)h^\top]\boldsymbol\Lambda^{1/2}\mU^\top=\mR\,(\boldsymbol\Lambda_r^{1/2}\mU_r^\top;\mathbf 0)$, so $\mM^\top\mM=\mU_r\boldsymbol\Lambda_r\mU_r^\top$ and $\mM^\top f^\star=\mU_r\boldsymbol\Lambda_r^{1/2}h_{1:r}=\mU_r\mU_r^\top b$, which is $b$ if $k\ge s$ ($r=s$, $\mU_r=\mU$). Since $\E[f^\star f^{\star\top}]=\mI_k$, $\mM^\top=\E[gf^{\star\top}]$ is the least-squares coefficient of $g$ on $f^\star$; since $\E[gg^\top]=\mI_d$, $\mM=\E[f^\star g^\top]$ is that of $f^\star$ on $g$, a linear function of $g$, whereas $\E[f^\star(x_1)\mid x_2]$ need not be. For the rotation, a thin singular value decomposition $\mM=\mU_M\boldsymbol\Sigma_M\boldsymbol V_{\!M}^\top$ with $\bar k$ columns (completed by orthonormal columns if $\mathrm{rank}\,\mM<\bar k$) gives $\mQ^\star=\mU_M\boldsymbol V_{\!M}^\top\in\mathrm O(k,d)$ with $\mM=\mQ^\star(\mM^\top\mM)^{1/2}$, hence $\mM^\top=(\mM^\top\mM)^{1/2}\mQ^{\star\top}$, and $\mathrm{tr}(\mQ^{\star\top}\mM)=\mathrm{tr}(\mM^\top\mM)^{1/2}=\|\mM\|_*$, the maximum of (i). Let $\mQ$ be any maximizer and $(\mM^\top\mM)^{1/2}=\sum_{i\le r}c_iv_iv_i^\top$ with $c_i>0$. Then $\|\mM\|_*=\mathrm{tr}(\mQ^\top\mQ^\star(\mM^\top\mM)^{1/2})=\sum_ic_i\,v_i^\top\mZ v_i$ with $\mZ=\mQ^\top\mQ^\star$ of operator norm at most 1, so $v_i^\top\mZ v_i\le1$, and equality forces $\mZ v_i=v_i$ for every $i\le r$. With $w_i=\mQ^\star v_i$, $1=\|\mQ^\top w_i\|\le\|w_i\|\le1$. If $k\le d$, $\mQ\mQ^\top=\mI_k$ and $\mQ v_i=\mQ\mQ^\top w_i=w_i$; if $k>d$, $\mQ\mQ^\top$ projects on the range of $\mQ$, and $\|\mQ^\top w_i\|=\|w_i\|$ puts $w_i$ in it, so $w_i=\mQ\mQ^\top w_i=\mQ v_i$. Either way $\mQ v_i=\mQ^\star v_i$ for $i\le r$: all maximizers agree on the range of $\mM^\top\mM$, and the maximizer is unique if $k=d=s$. For the converse, let $f$ be centered and whitened and $\boldsymbol A\in\R^{d\times k}$. As $g-b$ is orthogonal to every function of $x_1$, $\E\|g-\boldsymbol A f\|^2=\mathrm{tr}(\mI-\mC)+\E\|b-\boldsymbol A f\|^2$, and $\mM_f^\top=\E[gf^\top]=\E[bf^\top]$, the least-squares coefficient of $g$ on $f$, is also that of $b$, so $\E\|b-\mM_f^\top f\|^2=\mathrm{tr}\,\mC-\|\mM_f\|_F^2$. With $\boldsymbol H$ and $\mK$ as in the proof of \cref{thm:ceiling}(i), $\|\mM_f\|_F^2=\|\boldsymbol H\boldsymbol\Lambda^{1/2}\|_F^2=\mathrm{tr}\,\mK\le\sum_{i\le r}\Lambda_{ii}$, as $\kappa_i\le\Lambda_{ii}$ and $\mathrm{rank}\,\mK\le r$, with equality iff $\kappa_i=\Lambda_{ii}$ for every $i\le r$, i.e., iff the span of $f$ contains $h_{1:r}$; and $\sum_{i\le r}\Lambda_{ii}=\mathrm{tr}\,\mC$ iff $r=s$, i.e., $k\ge s$. If $k=d$, then $s\le k$, $k\rho=\mathrm{tr}\,\mC^{1/2}$ and $\E\|\mQ^\top f\|^2=d$, so $\E\|g-\mQ^\top f\|^2=2d-2\,\mathrm{tr}(\mQ^\top\mM_f)\ge2d-2\|\mM_f\|_*\ge2d-2\,\mathrm{tr}\,\mC^{1/2}=\mathrm{tr}(\mI-\mC)+\mathrm{tr}\big((\mI-\mC^{1/2})^2\big)$, with equality at $\mQ=\mQ^\star$ in the first step and, by \cref{thm:ceiling}(i), iff the span of $f$ contains $h_{1:s}$ in the second. Since $\E[gg^\top]=\mC+\E[(g-b)(g-b)^\top]$, $\mC\preceq\mI$, so $\mathrm{tr}\big((\mI-\mC^{1/2})^2\big)=\sum_{i\le d}(1-c_i^{1/2})^2$ over the eigenvalues $c_i\in[0,1]$ of $\mC$ vanishes iff $\mC=\mI$, i.e., iff $\E\|g-b\|^2=\mathrm{tr}(\mI-\mC)=0$. In the world model, if $g$ is a function of the state and $k=d$, then $\mC=\mI$, $b=a_g=g$, and $g=\mM^\top f^\star=\mQ^{\star\top}f^\star$, i.e., $f^\star=\mQ^\star g$.\\
(iii) By (A5), $\hat\mM-\mM$ is the mean of $n$ independent centered copies of $f^\star(x_1)g(x_2)^\top-\mM$, so $\E\|\hat\mM-\mM\|_F^2=\big(\E\|f^\star\|^2\|g\|^2-\|\mM\|_F^2\big)/n=V/n$, with $\|\mM\|_F^2=\mathrm{tr}\,\mC$ here. For the orthogonal factor \citep{li1995polar}, let $k=d=s$, $\mE=\hat\mM-\mM$, $\hat\mM=\hat\mQ\hat{\boldsymbol S}$ with $\hat{\boldsymbol S}=(\hat\mM^\top\hat\mM)^{1/2}\succeq0$ (any orthogonal factor if $\hat\mM$ is singular), and $\mM=\mQ^\star\boldsymbol S$ with $\boldsymbol S=\mC^{1/2}\succ0$. With $\mX=\hat\mQ^\top\mQ^\star$, $\hat\mQ^\top\mE=\hat{\boldsymbol S}-\mX\boldsymbol S$ and $\mE^\top\mQ^\star=\hat{\boldsymbol S}\mX-\boldsymbol S$, hence $\hat{\boldsymbol S}(\mI-\mX)+(\mI-\mX)\boldsymbol S=\hat\mQ^\top\mE-\mE^\top\mQ^\star$. In eigenbases of $\hat{\boldsymbol S}$ and $\boldsymbol S$, each entry of $\mI-\mX$ is the matching entry of the right side divided by a sum of two eigenvalues, at least $\sigma_{\min}(\mM)=\Lambda_{kk}^{1/2}>0$, so $\|\hat\mQ-\mQ^\star\|_F=\|\mI-\mX\|_F\le2\|\mE\|_F/\Lambda_{kk}^{1/2}$; Jensen's inequality gives the bound in expectation, and $\|\hat\mQ-\mQ^\star\|_F\le\|\hat\mQ\|_F+\|\mQ^\star\|_F=2\sqrt k$ always.\\
(iv) Since $\hat\mM_W-\mM_W$ has rank at most $\bar k$, $|\|\hat\mM_W\|_*-\|\mM_W\|_*|\le\|\hat\mM_W-\mM_W\|_*\le\sqrt{\bar k}\,\|\hat\mM_W-\mM_W\|_F$, so $\E(\hat R_n(W)-R(W))^2\le\bar k\,\E\|\hat\mM_W-\mM_W\|_F^2=\bar kV_W/n$, the variance being computed as in (iii) with $f_W$ in place of $f^\star$. By Chebyshev's inequality, $|\hat R_n(W_i)-R(W_i)|\ge\epsilon/2$ with probability at most $4\bar kV_{W_i}/(n\epsilon^2)$, and by a union bound all $m$ deviations are below $\epsilon/2$ with probability at least $1-4\bar k\sum_iV_{W_i}/(n\epsilon^2)$. On that event, for a best candidate $W_b$, $R(\hat W)>\hat R_n(\hat W)-\epsilon/2\ge\hat R_n(W_b)-\epsilon/2>R(W_b)-\epsilon$. If $W_b$ is Platonic and every other candidate has $R\le k\rho-\Delta$, then $\epsilon=\Delta$ gives $R(\hat W)>k\rho-\Delta$, so $\hat W$ is Platonic. Both bounds are worst-case (Chebyshev, and $\sqrt{\bar k}$ from the Frobenius norm), and what they certify is the rate. The checks of \texttt{tests/test\_theory.py} verify \cref{thm:ceiling,thm:every,thm:search} on generic laws of pairs (no world model), on lossy worlds where the target modality sees a coarse function of the state, and on world models with five kinds of anchors and adversarial cases (a pure-nuisance anchor, $|\mathcal Z|=k+1$, non-uniform $p$); the bound of (iii) on every sample, including $n<k$ where $\hat\mM$ is singular, and Li's sharper bound on 2{,}000 random matrices.

\subsection{\texorpdfstring{Certified directions}{Certified directions}}\label{prf:certified}

\paragraph{Certified directions.} For a unit-variance direction $u=w^\top f(x_1)$ of an encoder of the target modality and a unit-variance direction $v$ of the anchor, with world parts $a_u=\E[u\mid z]$ and $a_v=\E[v\mid z]$, conditional independence gives $c=\E[uv]=\E[a_ua_v]$, so $|c|\le(\E a_u^2)^{1/2}(\E a_v^2)^{1/2}\le(1-N(u))^{1/2}$ by Cauchy--Schwarz, i.e., $N(u)\le1-c^2$. This holds in population, while canonical correlations fitted on $n$ pairs are biased upward, so the certificate needs $c$ measured on pairs not used to fit the directions.

\subsection{\texorpdfstring{Proof of \Cref{prop:imagine}}{Proof that imagined observations are good views}}\label{prf:imagine}

\begin{fullstatement}{\Cref{prop:imagine}}
Assume (A1)--(A5) in the world model. Let a view sampler draw, from an observation $x$ of state $z$, a new observation $x'\sim\sum_{z'}\kappa(z'\mid z)\,p(x'\mid z')$: it imagines a state $z'$, which may depend on the state of $x$ but not on its nuisance, then renders a fresh observation of it with the modality's own rendering and nuisance law. If $p(z)\kappa(z'\mid z)$ is symmetric, the pairs $(x,x')$ are good views of the world kernel $\pi_\kappa(z,z')=p(z)\kappa(z'\mid z)$, and their JEPA learns $\vPhi_{\pi_\kappa}$, a function of the state alone. A sampler that instead corrupts $x$ independently in state and nuisance (\cref{prop:nuisance}), e.g., a crop, which keeps the viewpoint of a photograph, gives \emph{bad views}, a JEPA that spends a direction on nuisance, as soon as it keeps the nuisance more than the state's $k$-th direction ($\beta_1>\alpha_k$).
\end{fullstatement}

\paragraph{Proof.} With $x=\gamma(z,\nu)$ and $x'=\gamma(z',\nu')$, injectivity gives $P(x)=p(z)q(\nu)$ and $p(x'\mid z')=q(\nu')$, so the pair has the law $p(z)\kappa(z'\mid z)\,q(\nu)q(\nu')=\pi_\kappa(z,z')q(\nu)q(\nu')$, the good-view law of \citet[Prop.~2(i)]{balestriero2026prh}: $\pi_\kappa$ has marginal $p$ and is symmetric by assumption, so (A2) holds, and when $\pi_\kappa$ satisfies (A3) the JEPA learns $\vPhi_{\pi_\kappa}$. For a product corruption $\boldsymbol A\otimes\boldsymbol B$ with the eigenstructure of \cref{prop:nuisance}, the pure-nuisance eigenfunction $\chi_1(\nu)$ has eigenvalue $\beta_1=\alpha_0\beta_1$; if $\beta_1>\alpha_k$, fewer than $k$ eigenvalues exceed it (at most $k-1$ of the $\alpha_i$, $i\ge1$, and none of the $\alpha_i\beta_j\le\beta_1$, $j\ge1$), so by (A3) $\chi_1$ is in the JEPA's top $k$: the JEPA spends a direction on the nuisance. One that keeps the nuisance exactly ($\boldsymbol B=\mI$) gives every function of the nuisance the eigenvalue 1, above every direction of the state that is not kept exactly. The exact posterior $\kappa(z'\mid z)=\mathbf 1[z'=z]$ gives $\pi_\kappa=\mathrm{diag}(p)$, whose Markov operator is the identity on functions of the state: (A3) fails and every whitened function of the state is a minimizer.

\subsection{\texorpdfstring{What nuisance costs: \Cref{prop:nuisance} and its proof}{What nuisance costs}}\label{prf:nuisance}

\begin{proposition}[What Nuisance Costs]\label{prop:nuisance}
Assume (A1)--(A5) in the world model, and let $W$ corrupt one observation independently in its state and its nuisance (a product corruption $\boldsymbol A\otimes\boldsymbol B$), keeping the state's $i$-th direction with correlation $\alpha_i$ and the nuisance's $j$-th with $\beta_j$, both sorted ($1=\alpha_0\ge\alpha_1\ge\dots$, $1=\beta_0\ge\beta_1\ge\dots$, all in $[0,1]$), and let the anchor's world part span the top $s\le k$ state directions of the corruption ($\alpha_s>\alpha_{s+1}$). Then $R(W)<k\rho$ \emph{if and only if} at least $k-s+1$ of the products $\alpha_i\beta_j$ with $j\ge1$ are at least $\alpha_s$: always if $\beta_{k-s+1}\ge\alpha_s$, never if $\beta_1<\alpha_s$, and, for an anchor of full rank ($s=k$), exactly when $\beta_1>\alpha_k$.
\end{proposition}

\paragraph{Proof.} For corruptions $\boldsymbol A\otimes\boldsymbol B$ of state and nuisance, the Markov operator has eigenfunctions $\psi_i(z)\chi_j(\nu)$ with eigenvalues $\alpha_i\beta_j$, and by (A3)--(A4) $f_W$ spans the $k$ largest after $(0,0)$, all those at least the $k$-th \citep[Prop.~2]{balestriero2026prh}. As $s\le k$, $r=s$ and the anchor's $h$ spans $\psi_1,\dots,\psi_s$, so by \cref{thm:search}(i) $R(W)=k\rho$ iff all of them are among the top $k$, iff $\psi_s$ is, iff at most $k$ eigenvalues are at least $\alpha_s$. Those with $j=0$ are $\alpha_1,\dots,\alpha_s$, since $\alpha_s>\alpha_{s+1}$, so the condition is that at most $k-s$ products $\alpha_i\beta_j$ with $j\ge1$ are at least $\alpha_s$. These include $\beta_j=\alpha_0\beta_j$ for every $j$ with $\beta_j\ge\alpha_s$, which gives the sufficient condition, and none exists if $\beta_1<\alpha_s$, since $\alpha_i\beta_j\le\beta_1$. For $s=k$ the condition fails iff $\beta_1\ge\alpha_k$, and $\beta_1=\alpha_k$ would tie the $k$-th and $(k+1)$-th eigenvalues, which (A3) excludes. The numerical checks of \texttt{tests/test\_theory.py} verify the boundary, including a case with $s<k$ in which the reward stays maximal past $\beta_1=\alpha_s$.

\subsection{\texorpdfstring{Proof of \Cref{prop:closed}}{Proof for a closed vocabulary}}\label{prf:closed}

\begin{fullstatement}{\Cref{prop:closed}}
Assume (A1), (A5), $|\X_1|\le k+1$, and let $\vphi$ be a whitened basis of the centered functions of $x_1$. (i) If $|\X_1|=k+1$, every centered, whitened $f:\X_1\to\R^k$ is $\mQ\vphi$, $\mQ\in\Oset$, and attains the ceiling $k\rho=\mathrm{tr}\,\mC^{1/2}$: $R(W)$ is the same for every view law. (ii) If $|\X_1|\le k$, no $k$-dimensional encoder is whitened and (A4) is void; an encoder whose values are affinely independent, re-whitened on their span, is $\mQ\vphi$ with $|\X_1|-1$ outputs and attains the ceiling $\mathrm{tr}\,\mC^{1/2}$ at that width. Separating the values is not enough: three collinear values span one dimension.
\end{fullstatement}

\paragraph{Proof.} Let $\boldsymbol D=\mathrm{diag}(P)$ hold the law of the $|\X_1|$ values; the centered functions of $x_1$ form a subspace of dimension $|\X_1|-1$, which contains the centered $h$, so $s\le|\X_1|-1$. (i) If $|\X_1|=k+1$, the $k$ coordinates of a centered, whitened $f$ are an orthonormal family in that $k$-dimensional subspace, hence a basis of it, and two orthonormal bases of one space differ by an orthogonal matrix: $f=\mQ\vphi$. Its span contains $h$, so \cref{thm:ceiling}(i) holds with equality, and $k\rho=\mathrm{tr}\,\mC^{1/2}$ as $s\le k$. Every view law has exactly $k$ non-constant eigenfunctions, so (A3) holds vacuously ($\mu_{k+1}$ does not exist) and every JEPA is such an $f$: $R(W)=\mathrm{tr}\,\mC^{1/2}$ for all $W$. (ii) If $|\X_1|\le k$, at most $k-1$ centered functions are orthonormal, so no $k$-dimensional encoder is whitened. Let $\mF\in\R^{|\X_1|\times k}$ hold the centered values of $f$ ($\mathbf 1^\top\boldsymbol D\mF=0$). The values are affinely independent iff $\mF$ has rank $|\X_1|-1$, i.e., iff the coordinates of $f$ span the centered functions; re-whitening on that span, $\mF\boldsymbol T$ with $\boldsymbol T\in\R^{k\times(|\X_1|-1)}$ and $\boldsymbol T^\top\mF^\top\boldsymbol D\mF\boldsymbol T=\mI$, is an orthonormal basis of them, and (i)'s argument applies at width $|\X_1|-1\ge s$, where \cref{thm:ceiling}(i) gives the ceiling $\mathrm{tr}\,\mC^{1/2}$. Distinct but affinely dependent values give $\mathrm{rank}\,\mF<|\X_1|-1$, a span that misses some centered function, and by \cref{thm:ceiling}(i) the encoder falls short of $\mathrm{tr}\,\mC^{1/2}$ unless that span still contains $h$. \texttt{tests/test\_theory.py} checks (i) on random laws of pairs, for random encoders and the JEPAs of random view laws, and that with more values than $k+1$ random encoders fall short of the bound.

\subsection{Synthetic Checks of the Theory}\label{app:synthetic}

\begin{figure*}[t]
\centering
\includegraphics{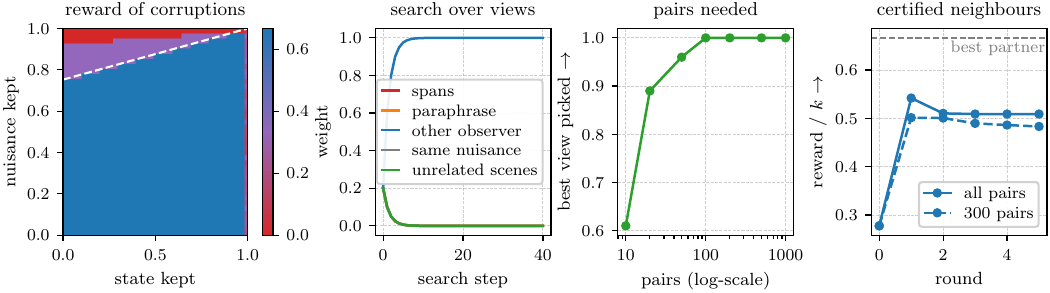}
\caption{\textbf{PVS on a world where everything is known.} Scenes on a circle, observed by an image modality through bad views (crops that keep a viewpoint) and by a text modality; $k=6$. \textbf{(a)} The reward of every corruption of one observation, by how much of the state and of the nuisance it keeps. Dashed: the boundary of \cref{prop:nuisance} (\cref{prf:nuisance}), here $\beta_1=\alpha_s$, since the corruption keeps every nuisance direction alike. \textbf{(b)} A search over five text view laws: spans of one caption, a paraphrase, another observer, positives that share a nuisance, and unrelated scenes. With its reward estimated from 300 pairs, the search puts all its weight on the other observer. \textbf{(c)} How often the reward picks the best of the five from $n$ pairs. \textbf{(d)} Neighbors in certified directions (\cref{app:flickr_details}), iterated from a bad-view JEPA. We observe that \emph{the reward collapses exactly where the theory says, and a few hundred pairs find the best views}.}
\label{fig:synthetic}
\end{figure*}

\begin{figure*}[t]
\centering
\includegraphics{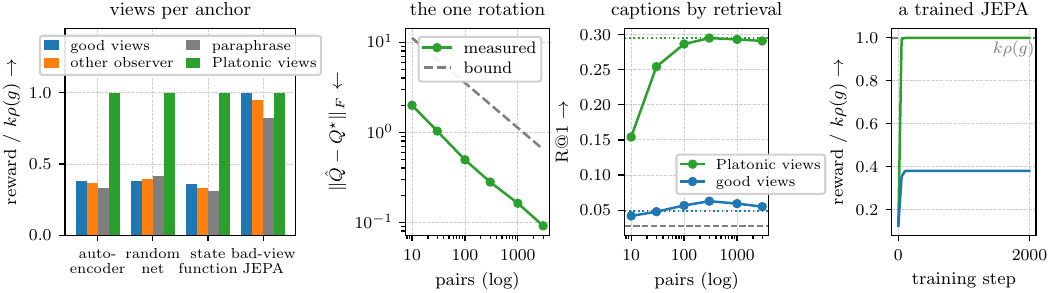}
\caption{\textbf{Every anchor has Platonic views (\cref{thm:every}).} The world of \cref{fig:synthetic}, $k=6$, with four anchors: a linear autoencoder of raw renderings, a random network, a function of the state alone, and the bad-view JEPA of \cref{fig:synthetic}. \textbf{(a)} Reward of four view laws relative to the ceiling $k\rho$. Good views of the world reach it only for the JEPA anchor. An anchor's Platonic views reach it for every anchor. \textbf{(b)} Autoencoder anchor: the rotation fitted on $n$ pairs converges to the unique $\mQ^\star$ at the rate $n^{-1/2}$, below the bound of \cref{thm:search}(iii). \textbf{(c)} Retrieval of the right scene from the source to the target modality through that rotation (R@1; dashed: chance, $1/36$). The Platonic JEPA reaches its ceiling (dotted), set by the anchor's nuisance, from about 100 pairs. \textbf{(d)} A two-layer MLP JEPA trained by Adam on the views, with invariance plus a covariance penalty. On Platonic views, its reward climbs to the bound. On good views of the world, it stalls at their spectral value. We observe that \emph{the views to search for are those of the anchor, whatever trained it, and a few pairs then fit the one rotation}.}
\label{fig:anchors}
\end{figure*}

We check \cref{thm:every,thm:search,prop:nuisance} on a synthetic world where every quantity is known. \Cref{fig:synthetic} runs PVS on this world. \Cref{fig:anchors} builds the Platonic views of four anchors and fits the rotation from $n$ pairs.

\section{The View Search, Step by Step}\label{app:search}

\begin{figure*}[t]
\centering
\includegraphics{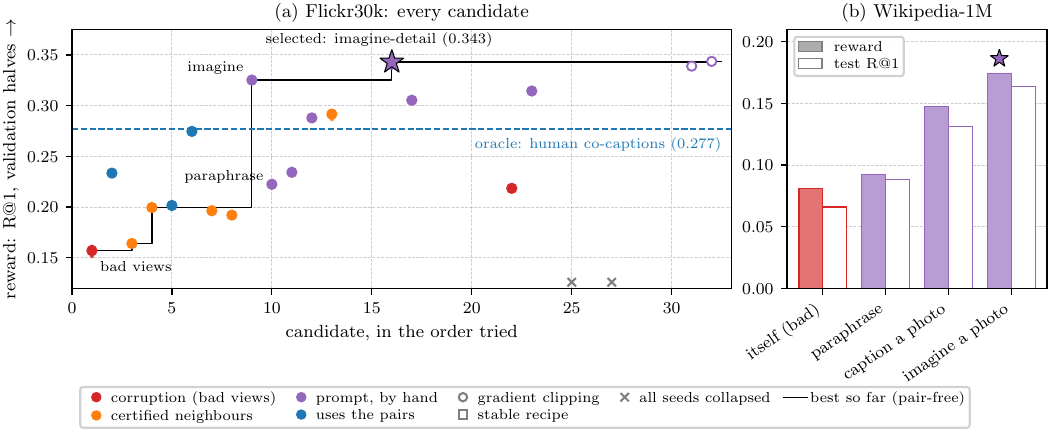}
\caption{\textbf{Every candidate of the search.} Reward of each view generator on Flickr30k (validation halves; dots: seeds), in the order in which the search proposed them. The plot also shows the best reward so far and the human co-captions (oracle, dashed). Some arms use the reward's pairs: the neighbors certified on these pairs, and their descendants. We mark these arms and leave them out of the best so far (\texttt{paper/analysis.py search}, \texttt{data/search.csv}).}
\label{fig:search}
\end{figure*}

\Cref{fig:search} lists every candidate of the Flickr30k search with its reward.

\clearpage
\section{The Algorithm and Natural Images: Details}\label{app:flickr_details}

This section gives the details of \cref{sec:impl,sec:flickr}.

\subsection{The Algorithm}

\paragraph{The anchor.} The anchor is a frozen encoder of the source modality, JEPA or not (\cref{thm:every}). When we pretrain on text, the anchor is DINOv2 ViT-B \citep{oquab2024dinov2} for images. For every other source modality, the anchor is a self-supervised encoder trained without text (\cref{app:domains}). The anchor is never trained and scores the candidates (\cref{eq:reward}) on a few hundred paired observations. These observations are the validation pairs of a dataset, held out from every test. Later, the anchor reads its modality (\cref{sec:exp}).

\paragraph{The candidate samplers.} A candidate sampler can be any distribution of positive pairs over the unpaired data of the target modality. When the target modality is text, the imagining samplers of \cref{prop:imagine} are a language model (Qwen2.5-7B-Instruct \citep{qwen2024qwen25}) that rewrites each text on its own. This language model never sees an item of the source modality or another text of the same item. Sentence encoders take their positives from dropout \citep{gao2021simcse} or have a language model write them \citep{zhang2023syncse}. PVS also has a language model write candidate positives, but PVS selects them by their agreement with a frozen encoder of the source modality (below). We use five prompts of one generic family, the same in every domain: imagine (another observer of the same item describes it), paraphrase, literal (keep only what the item itself would show), report (what an automatic analyzer would print) and caption (what a person would write). We fill each prompt with the name of the modality and nothing else (e.g., ``30-second sleep EEG recording''; \texttt{platonic/generate.py}, \texttt{data/domains/<domain>/modality.txt}). Two other families serve as references. The first family is \emph{corruptions}, i.e., spans of one text or crops, the bad views of \citet{balestriero2026prh}. The second family is \emph{certified neighbors}. By the conditional independence behind \cref{thm:ceiling}(ii), a direction of the target encoder whose population correlation with a direction of the anchor is $c$ carries at most $1-c^2$ of nuisance (\cref{prf:certified}). Sample correlations are inflated, so such a certificate needs held-out pairs. We thus fit $r$ canonical directions on 1{,}000 other pairs, and we pair every text with its nearest texts in these directions.

\paragraph{The search.} Each candidate rewrites the corpus once. Each set of rewrites trains a JEPA with one fixed recipe, the BERT-S LeJEPA (12 layers of width 384) of \citet{balestriero2026prh}. We train every JEPA with stable-pretraining on the same number of samples, so that only the positives change. The reward is the retrieval R@1 from the source modality to the distinct texts of the target modality. Identical texts count once, so that templated reports do not tie with their true pair. We fit a rotation on half of the validation pairs and score R@1 on the other half. Mutual nearest neighbors break ties. This reward is a rank-based, held-out proxy for the agreement that $R$ measures (\cref{eq:reward}). \Cref{thm:search}(iv) bounds only the estimate of $R$ itself, not this proxy. Moreover, the theory holds for spectral JEPAs under (A1)--(A5), while we apply it to LeJEPA. In practice, this proxy ranks views as the test pairs do (\cref{fig:reward_vs_test}). We keep the best candidate, rewrite the whole corpus with it and train a JEPA on these rewrites. For short, we call this JEPA the Platonic JEPA of the anchor. This JEPA is the best candidate found, not one proven to reach the ceiling $k\rho$, which few pairs cannot estimate. Mixtures and iterations of candidates are candidates too (\cref{fig:search} lists every candidate). Prior work tuned views against a downstream label \citep{tian2020infomin}, against a supervised or self-supervised score \citep{cubuk2019autoaugment,reed2021selfaugment}, or adversarially \citep{tamkin2021viewmaker}. Our reward needs no label, and the theory ties its population version to the best reachable alignment.

\paragraph{Reading the source modality.} We align the Platonic JEPA with the anchor in one of three ways. The first way is one rotation fitted on the $n$ available pairs. The second way aligns the unpaired samples of both modalities around those pairs, where each pair weighs as much as $w$ pseudo-pairs \citep{schnaus2026rosetta}. The third way is the least-squares map. With few pairs, the rotation weighs every direction alike, however poorly the pairs predict that direction. The least-squares map instead scales the rotation by how well the pairs predict each direction. From the JEPA to the anchor, the least-squares map is the conditional expectation (\cref{thm:search}(ii)). We use the reverse map, i.e., the best linear predictor of the JEPA's embedding from the anchor. We estimate this predictor in closed form by ridge regression from the anchor's features onto the JEPA's embeddings of the pairs' texts. We standardize the anchor's features on its unlabeled samples. Retrieval may also discount texts that are close to many inputs at once. For this, we use cross-domain similarity local scaling (CSLS, \citealp{conneau2018word}), measured on unlabeled inputs. Validation first chooses one $w$ per domain. For each $n$, validation then chooses the map: the rotation, the anchored alignment or the ridge map, each with or without CSLS. If the map of a smaller $n$ is better on validation, we keep that map, so the validation score never decreases as pairs are added. We then read an item of the source modality in one of three ways: by retrieving the nearest text of the corpus, by a text-only decoder conditioned on the Platonic JEPA, or by a language model that reads the nearest texts of the corpus in context. Throughout this paper, the same steps, prompts and recipe serve every pair of modalities, and no network is ever trained on a pair.

\subsection{Natural Images}

\paragraph{Setup.} Every text JEPA follows the recipe of \cref{sec:impl}. On Flickr30k \citep{young2014flickr}, each of the 145k training captions is a text on its own. Only the oracle uses the grouping of captions by photo. The positives of the oracle are the other human captions of the same photo, the good views of \citet{balestriero2026prh}. We report image-to-caption R@1 and R@5 on the 1{,}000 test pairs, through one rotation fitted on the 1{,}014 validation pairs. We also report mutual 10-NN (mNN), which fits nothing \citep{huh2024platonic}. The test pairs never choose anything. About one run in seven collapses at the learning rate $5\cdot10^{-4}$ of \citet{balestriero2026prh}, whatever the views. As in \citet{balestriero2026prh}, we average over the seeds that do not collapse. A table at the stable learning rate $10^{-4}$ confirms every conclusion (\cref{tab:stable}).

\Cref{fig:flickr} reports the main candidates, \cref{fig:reward_vs_test} every run of the search, and \cref{tab:flickr_views} every measure of each run. The reward ranks four prompts as the test does. Imagining the photo and writing another observer's caption beat the oracle. A paraphrase and a literal rewrite do little better than bad views, as \cref{prop:imagine} predicts. The best prompt asks for the photo's visual details (people, objects, colors, clothing and setting). This prompt reaches an R@1 of .309 against .246 for the oracle, and it beats the oracle in R@5 and in every seed. These rewrites do not copy Flickr30k. For each text, we find its nearest human training caption by TF-IDF similarity. For a rewrite, this nearest caption is another caption of the same photo half as often as for the original caption. Almost no rewrite repeats a human caption (\cref{tab:flickr_views}; \texttt{scripts/recall.py}, \texttt{data/recall.csv}).

\paragraph{Separating views from data.} The same 725k texts as the imagine arm, each a positive of nothing but its own spans, reach .187 (\cref{fig:flickr}); hence, the extra text explains a third of the gain over bad views, while turning it into positives explains the rest.

\paragraph{Views with less nuisance align better.} By \cref{thm:ceiling}(ii), views that leave less nuisance should align better. We test this prediction on the five human captions of each validation and test photo. The scene-explained fraction of a text JEPA is $\rho_f=\mathrm{tr}\,\mC_f^{1/2}/k$, the encoder's analogue of $\rho$ ($k=32$). This fraction rises from .649 for bad views, through the paraphrase, the oracle and imagined views, to .831 with visual details. The fraction orders these five encoders as alignment does. Over all eight encoders scored, two adjacent pairs swap (\cref{tab:flickr_views}, \texttt{data/nuisance.csv}). We also compute the theory's own reward, $R$ of \cref{eq:reward}, on the whitened top 32 directions of these pairs. This reward ranks the eight encoders much as the search's R@1 reward does, and as test alignment does in R@1 and in mutual 10-NN (\texttt{paper/analysis.py theory\_reward}, \texttt{data/theory\_reward.csv}).

\paragraph{Neighbors need a space that already encodes the world.} Neighbors in certified directions of a bad-view JEPA reach .170. In the same directions of the imagine JEPA, neighbors alone reach .265, above the oracle in every seed (\cref{tab:flickr_views}). Neighbors thus propagate the world that a space already encodes, but they cannot create it.

\paragraph{Other learning rates, anchors and datasets.} \Cref{tab:stable,fig:pairs} hold the robustness checks of \cref{sec:flickr}. The weak anchor is a ViT-S image JEPA trained on the 29k photos of Flickr30k. This anchor ranks the three imagined generators above the human captions, and every generator above bad views (\texttt{data/weak\_anchor.csv}).

\clearpage
\section{Platonic JEPAs of Language: Details}\label{app:scale_details}

This section gives the details of \cref{sec:scale}.

\paragraph{Text JEPAs and their views.} Sentence encoders trained by agreement start from a pretrained language model. These encoders then corrupt or pair the inputs of this model, e.g., with dropout noise \citep{gao2021simcse}, nearby spans of one document \citep{giorgi2021declutr}, or positives written by a large language model \citep{zhang2023syncse}. data2vec predicts the latent targets of masked text from scratch \citep{baevski2022data2vec}. LLM-JEPA and semantic tube prediction add a JEPA term to next-token training \citep{huang2025llmjepa,huang2026stp}.

\paragraph{An image-free Wikipedia corpus.} We split the articles of English Wikipedia (dump of November 2023) into sentences, and we keep those of 8 to 60 words. We remove every sentence that matches a caption of WIT-1024 \citep{srinivasan2021wit} by its normalized text or its first eight words. WIT-1024 is the benchmark of the PRH \citep{huh2024platonic}. No sentence was written for an image, chosen by an image--text model, or grouped with another sentence. Web alt-texts differ, since DataComp \citep{gadre2023datacomp} filtered them with CLIP \citep{radford2021clip,li2024recap}. Nothing in our corpus has seen a pixel. The text-only language model of \cref{sec:impl}, served by vLLM \citep{kwon2023vllm}, writes the positives of each sentence, one rewrite per prompt. The bad-view baseline trains on the sentences alone, with the same corruptions, samples and recipe.

\paragraph{PVS on Wikipedia.} The BERT-S JEPAs of \cref{sec:scale} train on 1M sentences for 10M samples. Writing the caption of a photograph that could illustrate the sentence gives less than imagining that photograph (\cref{tab:wiki1m}), although none of those encoders has read a caption.

\paragraph{Scaling Platonic views.} We pretrain BERT-B on 10M Wikipedia sentences. Each sentence comes with two photographs that the language model of \cref{sec:impl} imagined for it (two samples of one prompt). We use the recipe of the 940M-image JEPA of \citet{balestriero2026prh} for 300M samples: batch 2{,}048, learning rate $2\cdot10^{-4}$, and weight .01 for SIGReg, the anti-collapse regularizer of LeJEPA \citep{balestriero2025lejepa}. \Cref{fig:models} compares this JEPA with 30 language models on WIT-1024 \citep{huh2024platonic}. With DINOv2-g, our JEPA reaches .188, above every language model we tested, from 0.5B to 284B parameters. The best language model is Qwen2.5-72B (.169), followed by Llama-3-70B, then by Qwen3.5-27B and Gemma-4-31B, which also read images (\cref{tab:wiki_scale}). The oracle is the same BERT-B trained on the LLaVA recaptions of 940M web images, written by an image captioner. The oracle stays above our JEPA in each of three seeds. Next-token prediction and bad views on the same sentences stay below our JEPA. Platonic views thus close nearly two thirds of the gap between bad views and the oracle, without a single image (\cref{tab:wiki_scale}). Following the PRH protocol \citep{huh2024platonic}, we read each model at the token that its objective trains: [CLS] for the JEPAs and the token mean for the language models. This token is also the best readout of each model. Over the 25 language models whose weights we hold, the last token loses .052 on average with DINOv2-g (.001 to .073). The next-token BERT, whose first token sees only itself, also reads best at the token mean (\texttt{scripts/slurm/readouts.sh}). Read at the token mean like the language models, the Platonic JEPA (.166) is on par with the best of them, Qwen2.5-72B (.169) and Llama-3-70B (.166), with 660 times fewer parameters.

\paragraph{Sentence encoders are JEPAs at a hundred times the scale.} Contrastive sentence encoders are text JEPAs whose positives are human-made pairs. For example, gte-base \citep{li2023gte} and all-mpnet-base-v2 \citep{reimers2019sentencebert,song2020mpnet} start from language models pretrained on billions of words. These encoders then train on about a billion pairs of texts. The pairs of all-mpnet-base-v2 include the other captions of the same Flickr30k and COCO photos, i.e., the oracle views of \cref{sec:flickr}. Qwen3-Embedding \citep{zhang2025qwen3embedding}, an 8B embedder started from the Qwen3 language model, first trains on about 150M synthetic weakly supervised pairs drawn from the pretraining corpus of Qwen3. Qwen3-Embedding then trains on about 7M labeled and about 12M synthetic supervised pairs, whose listed sources include MS MARCO, NQ, HotpotQA and natural language inference (NLI) data, among others. The report does not say which NLI sets it uses; the standard one, SNLI, takes its premises from Flickr30k captions \citep{bowman2015snli}. Such pairs may act like the oracle views. We thus use none of these encoders as our text side, we do not evaluate Qwen3-Embedding, and we report gte-base and all-mpnet-base-v2 only as references. gte-base and all-mpnet-base-v2 have the size of our BERT-B but a hundred times its text. Yet they reach at most .131 on WIT-1024, against .188 for our JEPA (\cref{fig:models}). Views written by a text-only model thus beat a billion human pairs, without using a pair. However, through a rotation fitted on Flickr30k pairs, the sentence encoders are on par with our JEPA up to 100 pairs, and ahead from 300 pairs (\cref{tab:flickr_pairs}; \texttt{scripts/flickr\_sweep.py}, \texttt{data/flickr\_pairs.csv}).

\paragraph{Views, data and capacity.} Two views per sentence are enough, and more sentences do not help BERT-B. The following variants align alike with DINOv2-g: one imagined photograph per sentence, two, six texts, or nine times more sentences with one photograph each (90M, 520M samples). Flickr30k test R@1 follows the same pattern (\cref{tab:wiki_scale}, \texttt{data/wiki\_scale.csv}, DINOv2-g at [CLS]). BERT-B saturates whatever the data, also when started from Google's weights. Capacity does help: BERT-L, started from Google's masked-language-model weights \citep{devlin2019bert}, reaches .195 after the same 30 epochs. This score is three quarters of the way from BERT-B's bad views to its oracle.

\paragraph{Other anchor, other seeds.} Our conclusions do not depend on DINOv2 (\cref{tab:wiki_scale}, \texttt{data/wit\_anchors.csv}). Against the largest DINOv3 (7B parameters), the four best language models gain a little, yet Platonic views still beat them. The Platonic BERT-L closes 77\% of the gap between bad views and the oracle. Over three seeds each, the Platonic BERT-L and BERT-B score alike (\texttt{data/wiki\_scale.csv}).

\paragraph{A captioner that never saw a pair.} The frozen Platonic JEPA conditions a text-only decoder \citep{mokady2021clipcap}. This decoder is a language model fine-tuned to write the caption of a text embedding. We train the decoder on the Flickr30k and COCO \citep{lin2014coco} captions as text, never with their images. One rotation, fitted on the Flickr30k training pairs, maps a frozen image encoder into the JEPA's space. Neither network sees a pair, and the rotation is the only parameter that does. The decoder is Qwen2.5-7B after its first epoch, and we show the 3B decoder for reference. We make every other choice on 300 COCO validation pairs, disjoint from the 4{,}700 test images (\cref{tab:captioner}). Validation makes three choices. First, the image side reads DINOv2-g and DINOv3-L \citep{simeoni2025dinov3} together instead of DINOv2-g alone. We concatenate the [CLS] and mean patch tokens of each encoder, each at unit scale. Second, DINOv3-7B later joins as a third image encoder. Third, we decode by consensus \citep{kumar2004mbr}: among three beams and 32 samples, we keep the caption most similar to the others, with a CIDEr-like similarity that uses no reference caption. The final captioner uses 128 samples (\cref{tab:coco}). Every decoder writes at most 64 tokens, the length it was trained on. On the COCO Karpathy test \citep{karpathy2015deep}, the captioner reaches a CIDEr \citep{vedantam2015cider} of 66.4, with no image--text model anywhere. This score is above the unpaired captioner of \citet{feng2019unsupervised} (54.9). \Cref{tab:coco} adds the captioner's BLEU-4 \citep{papineni2002bleu} and METEOR \citep{banerjee2005meteor}.

\paragraph{From many pairs to a few.} Flickr30k's pairs mostly depict people. We thus also fit the rotation on COCO's own training pairs (82{,}783 images, disjoint from the test images). With these pairs, the same frozen networks reach 92.0 with DINOv3-7B as a third image encoder, and 93.0 with 128 consensus samples. These scores are on par with the text-only captioners built on CLIP, which inherit its 400M pairs (\cref{tab:captioner,tab:coco}). We also fit the rotation on $n$ of these pairs, with three random draws per $n$. The rotation alone degrades as the pairs become few (\cref{fig:captioner_pairs,tab:coco}). However, the $n$ pairs can also anchor an alignment of the unlabeled images with the captions of other COCO images. For this alignment, we use the Wasserstein Procrustes of \citet{schnaus2026rosetta}, after averaging each caption with four Platonic views written for it (the generic \texttt{imagine} prompt of \cref{sec:impl}). Each pair then counts as $w$ matched pseudo-pairs, with $w$ chosen on validation ($w=10$ here, 100 in the other domains). In contrast, \citet{schnaus2026rosetta} give all known pairs together the weight of all pseudo-pairs. From 30 pairs on, the captioner is above the unpaired captioner of \citet{feng2019unsupervised}, which needs a labeled object detector. From 1{,}000 pairs on, the rotation fitted on the pairs alone is better on validation. The selection rule of \cref{sec:impl} keeps the curve from decreasing (\texttt{demo/bundle.py}, \cref{fig:captioner_pairs,tab:coco}).

\paragraph{Against fine-tuned VLMs, and beyond COCO.} Qwen2.5-VL-7B \citep{bai2025qwen25vl}, fine-tuned with LoRA on the same pairs, is far ahead from 10 pairs on. So is the newest open VLM, Qwen3.8-27B \citep{qwen2026qwen38} (\cref{fig:captioner_pairs,tab:coco}). Better alignment on WIT does not make a better decoder input. The Platonic BERT-L aligns better on WIT, yet the same decoder trained through BERT-L scores lower on validation and on test (\cref{tab:captioner}).

\clearpage
\section{Domains, Retrieval and Captioning: Details}\label{app:vlm_details}

\begin{table*}[t]
\centering\small
\setlength{\tabcolsep}{5pt}
\caption{\textbf{Reading a modality costs milliseconds.} Latency for one input (ms) and GPU memory (GB). Captioning by retrieval uses the frozen anchor, one matrix and the nearest corpus text, and we time it on one GPU and on 16 CPU cores. Each fine-tuned VLM runs on the same GPU and generates as many tokens as its answers have on average (\texttt{demo/bench.py}). The last rows are our captioners, which end with a text decoder (\texttt{data/results/bench/bench.jsonl}). We observe that \emph{captioning by retrieval is up to two orders of magnitude faster and smaller than the fine-tuned VLMs, and runs on a CPU}.}
\label{tab:cost}
\begin{tabular}{lccl rrrr}
\toprule
 & \multicolumn{2}{c}{ours} & \multicolumn{3}{c}{fine-tuned VLM} & \multicolumn{2}{c}{ratio} \\
Domain & ms $\downarrow$ & GB $\downarrow$ & model & ms $\downarrow$ & GB $\downarrow$ & speed $\uparrow$ & size $\uparrow$ \\
\midrule
EEG & 26 / 32 & 0.2 & Qwen3.8-27B & 598 & 51 & 23$\times$ & 292$\times$ \\
 &  &  & Gemma-4-31B & 341 & 58 & 13$\times$ & 334$\times$ \\
Wrist motion & 6 / 12 & 0.2 & Qwen3.8-27B & 1{,}415 & 51 & 235$\times$ & 261$\times$ \\
 &  &  & Gemma-4-31B & 960 & 58 & 159$\times$ & 299$\times$ \\
Lung sounds & 23 / 393 & 1.3 & Qwen3-Omni-30B & 311 & 59 & 13$\times$ & 45$\times$ \\
 &  &  & Gemma-4-12B & 190 & 22 & 8$\times$ & 17$\times$ \\
Bird calls & 47 / 617 & 1.3 & Qwen3-Omni-30B & 951 & 59 & 20$\times$ & 44$\times$ \\
 &  &  & Gemma-4-12B & 1{,}061 & 22 & 23$\times$ & 17$\times$ \\
Insect and frog calls & 49 / 730 & 1.3 & Qwen3-Omni-30B & 1{,}036 & 59 & 21$\times$ & 44$\times$ \\
 &  &  & Gemma-4-12B & 1{,}120 & 22 & 23$\times$ & 17$\times$ \\
Physical simulations & 194 / 13{,}288 & 5.0 & Qwen3.8-27B & 2{,}961 & 51 & 15$\times$ & 10$\times$ \\
 &  &  & Gemma-4-31B & 2{,}125 & 58 & 11$\times$ & 12$\times$ \\
\midrule
Satellite images, 3B decoder & 829 / 14{,}962 & 13.2 & & & & & \\
Photographs, 7B decoder & 1{,}935 / -- & 60.7 & & & & & \\
\bottomrule
\end{tabular}
\end{table*}

This section gives the details of \cref{sec:domains,sec:vlm}.

\subsection{Domains and Protocol}

\paragraph{Domains.} Each domain brings three ingredients. The first is the anchor, a frozen self-supervised encoder trained without text. The second is a corpus of texts about the modality. The third is a few hundred paired items, the validation pairs, which rank the views and train no network. The reasoning results fit their maps on only 10 pairs from a disjoint set (\cref{app:domains} lists the data, anchors and metrics). Everything follows \cref{sec:impl}, and we make every choice on validation pairs. Earlier work maps between frozen encoders that are given. A linear map from a frozen image encoder into a frozen language model captions images \citep{merullo2023linearly}. Relative representations \citep{moschella2023relative} and ASIF \citep{norelli2023asif} compare frozen encoders through their similarities to a set of paired anchors, with no training. Other work aligns frozen vision and language encoders by a map fitted on their pairs \citep{maniparambil2024vision}. With few or no pairs, other work matches the geometry of the unpaired samples of the two encoders \citep{schnaus2025blind,jha2025vec2vec,schnaus2026rosetta}. We use the pair-anchored Wasserstein Procrustes of \citet{schnaus2026rosetta} (\cref{sec:impl}). In contrast, PVS chooses how the encoder of the target modality is trained. PVS picks its positive views by their agreement with the anchor, and then only needs the map.

\paragraph{The search in every domain.} The search of \cref{sec:impl} runs in every domain on that domain's own validation pairs, with the same five prompts and bad views of the same texts (\cref{tab:pvs}; \cref{tab:views_main} shows the selected views). In all 14 domains, the selected views have a higher reward than bad views of the same texts. The gain is largest on photographs, chest X-rays, everyday sounds and proteins (44 to 65\%), and smallest on satellite images, wrist motion, lung sounds, and insect and frog calls (4 to 6\%). Imagining wins on photographs, satellite images, galaxies, lung sounds and bird calls. Captioning wins on AudioCaps, chest X-rays, proteins, physical simulations and EEG. Paraphrasing wins on videos, the terse report wins on molecules and wrist motion, and the literal rewrite wins on insect and frog calls. The rejection of bad views is stable. Over 50 random halves of the validation pairs, bad views have the highest reward in 6\% of draws on average, and in 10\% with only 25 pairs. Both rates are below the one in six of a random pick (\cref{tab:pvs}, right). The winning prompt is less stable (\cref{tab:pvs}, last row). The five prompts are often within a few hits of one another, most of all on templated reports, which repeat a few dozen texts and leave a view sampler little wording to change.

\subsection{Retrieval and Captioning}

\paragraph{Frontier models, fine-tuned on our pairs.} We compare with the latest open models. For images and signals, we use Qwen3.8-27B \citep{qwen2026qwen38} and Gemma-4-31B \citep{gemma2026gemma4}. For sounds, we use Qwen3-Omni-30B \citep{qwen2025qwen3omni} and Gemma-4-12B, the Gemma~4 model that hears audio. Each VLM reads the modality in a format it was built for. Sounds are given as waveforms, EEG and wrist motion as images of their traces, and simulations as images of their labeled fields. We LoRA-fine-tune each VLM on the same $n$ pairs as ours in three configurations: LoRA on the language model only, or also on the input encoder at two learning rates. We choose the configuration and the epoch on validation with the metric of the domain (\texttt{scripts/vlm\_finetune.py}, \texttt{scripts/vlm\_select.py}). Ours makes the same kind of choice between the rotation and the anchored alignment. Each VLM curve also follows our rule: when a fine-tune is worse on validation than a fine-tune on fewer pairs, the curve keeps the test score of the fine-tune on fewer pairs. The metrics score what a text states about its input: the sleep stage of an EEG epoch and the activity of a wrist recording (balanced accuracy), the lung findings (macro-F1), the calling species (accuracy), and the system and parameters of a simulation (\cref{app:domains}). Other methods also train on pairs. LiT trains the text side on pairs against a locked image encoder \citep{zhai2022lit}. Cross-modal distillation trains an encoder of an unlabeled modality to match a frozen encoder of a paired one \citep{gupta2016cross}. ImageBind binds modalities to images through their pairs \citep{girdhar2023imagebind}. Like the VLMs, these methods train a network on the pairs. Ours only scores candidate views and fits one map.

\paragraph{Ahead on signals and scientific images with ten pairs.} We draw the pairs three times, and for each draw we choose the map of \cref{sec:impl} on validation. For each comparison, we resample the test inputs with replacement, score both methods on the same resampled inputs, and take the central 95\% of the score differences as a confidence interval (\cref{tab:comparison}; \texttt{scripts/robustness.py}, \texttt{data/comparison.csv}). With 10 pairs, our captions beat both fine-tuned VLMs on three domains, and every 95\% interval lies above zero. The three domains are sleep EEG, read by CBraMod \citep{wang2025cbramod}, simulations of physical systems from The Well \citep{ohana2024well}, read by Walrus \citep{mccabe2025walrus}, and lung sounds. On EEG, sleep-stage balanced accuracy reaches .424, against at best .405 for the VLMs (Qwen3.8-27B on one draw). Our captions stay ahead with 30 and 100 pairs on EEG and on wrist motion, read by ssl-wearables \citep{yuan2024sslwearables}. On wrist motion, they also stay ahead with 300 pairs. On simulations, they are ahead or on par with 30 and 100 pairs. Conversely, the VLMs are ahead with 10 wrist-motion pairs, with 300 pairs on EEG and simulations, and from 30 pairs on lung sounds.

\paragraph{The frozen encoder is what reads the modality.} We also read each encoder without any text model, on the same pairs, draws, validation rows and metric (\texttt{scripts/baselines.py}). We use three baselines. The first copies the caption of the nearest pair in the encoder's feature space. The second picks the label that most of the nearest pairs name. The third is a logistic regression on the encoder's features, trained on the labels that the domain's own scorer parses from the pairs' captions. The second and third baselines write each predicted label as the caption of the nearest pair that names it. We choose the neighbors, class weights, regularization and thresholds of each baseline on validation (\cref{tab:comparison}). With 10 pairs, ours is ahead of all three baselines on EEG, simulations and lung sounds. With 100 and 300 pairs, ours is ahead of all three on bird calls and on insect and frog calls. From 30 pairs, the logistic regression leads on EEG and lung sounds, because it is trained on the labels that the metric reads. With 300 pairs, the logistic regression also leads on wrist motion, where it beats the fine-tuned VLMs while writing nothing but those labels. Copying the nearest pair's caption needs no label. This copy leads on simulations from 30 pairs, where the anchor finds a simulation of the same configuration among the pairs. Elsewhere, the copy is behind ours or on par. The text side is not what wins. We pass two off-the-shelf sentence encoders through the same rotation: all-mpnet-base-v2 \citep{reimers2019sentencebert,song2020mpnet} and gte-base \citep{li2023gte}. Both are trained contrastively on about a billion text pairs. Both are on par with our Platonic JEPAs or ahead of them in every domain. With 100 to 300 pairs, they lead by .05 to .07 on EEG and by .01 to .03 on simulations and wrist motion (\cref{tab:comparison}; \texttt{scripts/slurm/baselines\_rot.sh}). Hence, on signals and scientific images, the frozen encoder of the modality matters more than any text model. Frontier VLMs fine-tuned on the same pairs read these modalities less well with few pairs.

\paragraph{Far ahead of Gemma~4 on animal sounds.} We test on the calls of 150 bird species and of 85 insect and frog species. Gemma-4-12B fine-tuned on the same 10 to 300 pairs barely beats chance, i.e., one species in 150 or in 85 (\cref{tab:curves}). From 100 pairs, our captions name the right species 5 to 10 times as often as Gemma-4-12B. With 1{,}000 pairs, ours still leads Gemma-4-12B by far, and ours keeps improving with every pair added. Qwen3-Omni-30B names a few bird species out of the box and stays ahead. On insects and frogs, Qwen3-Omni-30B is close to ours with 10 and 30 pairs. On lung sounds, our captions are ahead of Gemma-4-12B and of the constant ``Normal breath sounds'' (\cref{tab:comparison}, floor), but behind Qwen3-Omni-30B (\cref{tab:curves}).

\paragraph{Natural images and everyday sounds.} On photographs, satellite images, chest X-rays and everyday sounds, the best fine-tuned VLM is far ahead from 10 pairs on (\cref{fig:vlm_curves,tab:curves}). On sounds, Gemma-4-12B is behind ours from 100 pairs, although on everyday sounds both stay below a caption that reads no input up to 100 pairs. Qwen3-Omni-30B captions sounds well out of the box when it is shown three corpus captions as style examples. Platonic views still beat bad views through the whole protocol in these domains (\cref{tab:domains}). On RSICD \citep{lu2018rsicd}, we measure this with a text-only decoder, Qwen2.5-3B fine-tuned on the benchmarks' training captions as text. On AudioCaps \citep{kim2019audiocaps}, we measure it by retrieval over three seeds. Both use one rotation fitted on the validation pairs. On chest X-rays, n-gram metrics reward the most common report. A normal report for every image scores above every method on IU X-ray \citep{demnerfushman2016preparing,chen2020r2gen} (\cref{tab:domains}). We thus score the findings with CheXbert \citep{smit2020chexbert}. Under CheXbert, Qwen3.8-27B leads at every number of pairs, and ours surpasses Gemma-4-31B from 1{,}000 pairs (.205 against .156).

\begin{figure*}[p]
\centering
\includegraphics{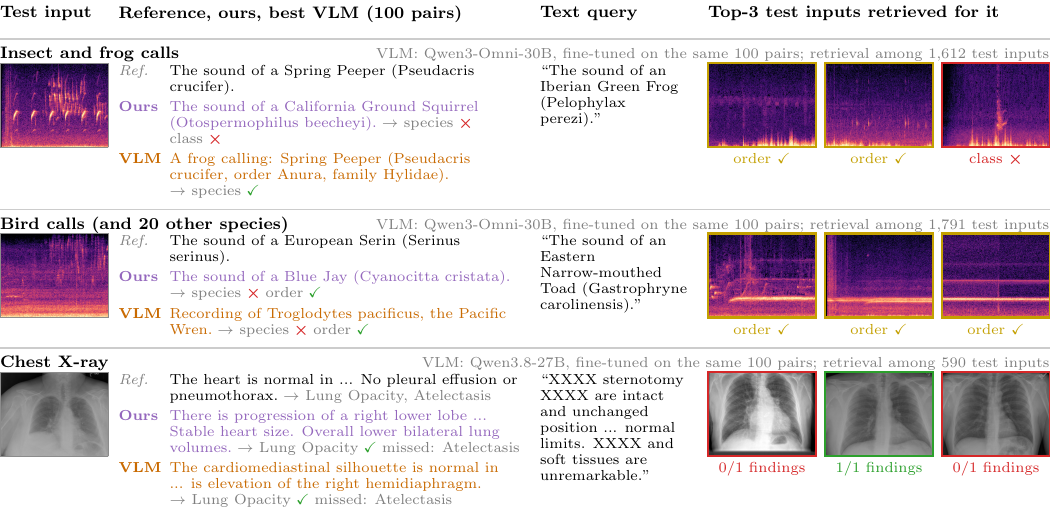}
\caption{\textbf{Held-out inputs of three more domains, read and searched with 100 pairs, next to the best fine-tuned VLM.} These are domains where VLMs lead (\cref{sec:vlm}). \Cref{fig:gallery_main} shows the four domains where our captions lead. Each row is one test input drawn at random (seed 0) among the test inputs whose reference names a label, never by whether an answer is right. We show its reference, our caption through the map chosen on validation with 100 pairs, and the caption of the same input by the VLM that ranks highest at 100 pairs in \cref{fig:vlm_curves}, fine-tuned on the same 100 pairs. Marks follow each domain's metric: the species, or else the deepest shared taxonomic level, and the CheXbert findings. Right: the first reference of another random test input serves as a text query. We show the three test inputs that our map ranks first for this query, scored against the query in the same way. Frames: green, all of the query's label; amber, part of it; red, none (\texttt{paper/gallery.py}). We observe that \emph{the VLM names the frog species where ours names another animal, both name a bird of the right order but the wrong species, and both read only part of the X-ray. Of the 9 retrieved inputs, 1 has all of the query's label and 5 have part of it.}}
\label{fig:gallery}
\end{figure*}

\paragraph{Reading individual inputs.} \Cref{fig:gallery_main} (four domains) and \cref{fig:gallery} (three more) show one random test input per domain, read and searched with 100 pairs, next to the best fine-tuned VLM on the same input. In \cref{fig:gallery_main}, both miss the stage of the EEG epoch, ours names the wrist activity where the VLM describes a person sitting, and the VLM names the simulation's pattern where ours names another one. In \cref{fig:gallery}, the VLM names the frog species where ours names another animal, and both find the lung opacity of the X-ray and miss its atelectasis. In retrieval, 15 of the 21 inputs that our map ranks first share all or part of the query's label.

\paragraph{Using every training pair.} The protocol uses the texts of the benchmarks' training pairs only as text. These training pairs can also fit the map (\texttt{demo/train\_pairs.py}, \texttt{demo/levels.sh}). We use the first $k$ pairs after a seed-0 shuffle and choose the map on validation. With the same frozen networks, AudioCaps retrieval and the satellite decoder on RSICD improve with the training pairs. AudioCaps reaches CIDEr 26.2 with all of them, and most of the gain comes by 3{,}000 pairs. The selection rule of \cref{sec:impl} gives every domain a curve from ten pairs to all of them (\cref{tab:curves}).

\paragraph{Retrieval.} The same maps also serve retrieval. We embed and map each test input, and compare it with the first caption of every test item. We score text-to-item recall at 10, which a constant answer cannot game (\texttt{scripts/retrieval\_ours.py}). We compare with open embedding models fine-tuned on the same pairs with their own losses. These are SigLIP~2 \citep{tschannen2025siglip2} and Qwen3-VL-Embedding \citep{qwen2026qwen3vlemb} on images and renderings, and PE-AV \citep{meta2025peav} on sounds. For each model, we choose on validation its learning rate and checkpoint, or the zero-shot model (\texttt{scripts/retrieval\_baselines.py}; \cref{tab:retrieval}). Ours retrieves better on chest X-rays from 100 pairs on. The contrastive models are far ahead on photographs, satellite images and everyday sounds. On simulations, they are slightly ahead with 100 pairs and far ahead with 3{,}000. Ours also retrieves better on wrist motion from 100 pairs on, and on insects and frogs with 100 pairs. On EEG, ours is on par, within the spread of the rotation's undetermined part (below). PE-AV is ahead on bird calls at every number of pairs. On lung sounds, the test reports take 13 distinct texts, and every method saturates.

\paragraph{Floors and significance.} Reading the corpus without looking at the input scores far below the curves. One random corpus text per test input, or the most frequent text, stays at the floor of every metric (\cref{tab:comparison}, last column; \texttt{scripts/robustness.py}). We draw the pairs two more times. We compare two methods by a paired bootstrap over test inputs (2{,}000 resamples; \texttt{scripts/robustness.py}). Each resample draws the test inputs with replacement within each draw of the pairs, and scores both methods on the same inputs. The central 95\% of the resampled differences in mean score form the interval. All three-draw comparisons of \cref{sec:vlm} use this procedure.

\paragraph{Pairs selected with the anchor.}\label{app:selection} Besides the random draws, we choose the 10 pairs from the anchor alone, before reading any caption, with TypiClust \citep{hacohen2022typiclust} (\texttt{scripts/select\_pairs.py}). We run $k$-means with ten clusters on the anchor's features of the whole pair set, centered and normalized, and take in each cluster its most typical member, i.e., the one with the highest inverse mean distance to its 20 nearest neighbors within the cluster. Three $k$-means seeds replace the three random draws. On sleep EEG, they select the same ten epochs, and on simulations they share seven of the ten pairs. The VLMs are fine-tuned on the same selected pairs, with the same recipes as for the random draws. On simulations, every selection labels all seven systems, whereas the first random draw labels only four of them, and the in-context reader scores far higher on open questions (\cref{tab:openqa}). The selected pairs also help the VLMs in most domains, and the decoder that reads the nearest pair's text often gains more than ours. These pairs lower our scores on sleep EEG, where our captions nevertheless stay ahead, and our answers to label questions on lung sounds, whose selections contain only normal breath sounds and fine crackles (\cref{tab:comparison,tab:labelqa}).

\paragraph{The rotation's undetermined part.} With fewer pairs than dimensions, the pairs determine the rotation only on their span. The singular value decomposition completes the rotation arbitrarily outside this span. We draw ten random completions (\texttt{scripts/rotation\_nullspace.py}). These completions move text-to-item recall at 10 by up to .03 on EEG, and by less on wrist motion. The completions move the captioning metric by at most .02 on chest X-rays, and by a fifth of that on EEG and wrist motion. The captioning and retrieval margins above clear this spread, except on EEG retrieval.

\paragraph{Cost.} \Cref{tab:cost} times one input on the same GPU. Each fine-tuned VLM writes as many tokens as its answers have on average. Captioning by retrieval runs on 16 CPU cores in under a second, except for physical simulations, whose encoder has 1.3B parameters. A new number of pairs costs one Procrustes fit, which takes milliseconds. In contrast, every point of a VLM curve is a fine-tuning run. However, our captioners for photographs and satellite images, which end with a 3B or 7B decoder, are not cheaper than a VLM (\cref{tab:cost}, last rows).

\clearpage
\section{Short and Open Questions: Details}\label{app:reasoning_details}

\begin{table}[t]
\centering\footnotesize
\caption{\textbf{Recognizing the system, and reading beyond it.} Test score of the in-context reader (gpt-oss-120b, 10 pairs) given the map's five hypotheses, or five random corpus descriptions of the input's own system, which tell it the system for free. Both are pooled over the three draws of the pairs, with the same seed for the control. The difference is ours minus the control, with a 95\% confidence interval from resampling the test inputs (all questions of one input together). When the map recognizes the system, the control does as well, except on questions that ask how one run will evolve (\texttt{scripts/factcheck.py system\_control}, \texttt{data/system\_control.csv}).}
\label{tab:system_control}
\setlength{\tabcolsep}{3pt}
\begin{tabular}{lrccc}
\toprule
 & answers (3 draws) & ours & same system, random & ours $-$ control\\
\midrule
\multicolumn{5}{l}{\emph{Simulations}}\\
\quad all test questions & 8{,}925 & 66.2 & 77.3 & -11.1 [-12.5, -9.8]\\
\quad the map's first hypothesis names the system & 7{,}005 & 76.3 & 77.3 & -1.0 [-1.9, -0.0]\\
\quad the map's first hypothesis names another system & 1{,}920 & 29.6 & 77.5 & -47.9 [-50.0, -45.7]\\
\multicolumn{5}{l}{\emph{Dynamics of one run}}\\
\quad all test questions & 8{,}925 & 43.7 & 46.9 & -3.2 [-4.0, -2.5]\\
\quad the map's first hypothesis names the system & 6{,}460 & 47.4 & 47.2 & +0.2 [-0.8, +1.2]\\
\quad \quad of these, questions on how the run evolves next & 1{,}292 & 26.3 & 21.8 & +4.5 [+2.7, +6.4]\\
\quad the map's first hypothesis names another system & 2{,}465 & 34.0 & 46.0 & -12.1 [-13.4, -10.6]\\
\bottomrule
\end{tabular}
\end{table}

\begin{figure*}[t]
\centering
\includegraphics[width=0.5\textwidth]{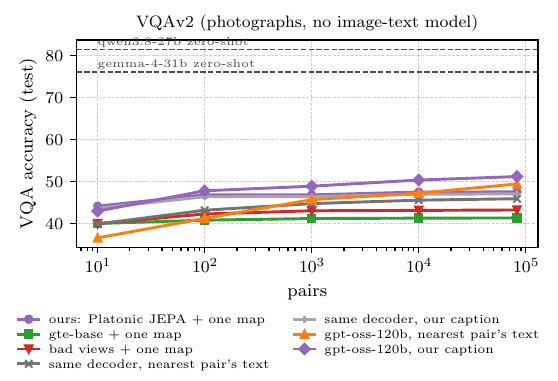}
\caption{\textbf{Question answering from a text-only decoder, against the number of pairs.} Test VQA accuracy on VQAv2 from 10 pairs to all of them, with no image--text model anywhere. Dashed: VLMs out of the box. All curves use the same decoder recipe trained on text alone, except gpt-oss-120b, a text-only reasoner that reads our caption or the text of the nearest pair. We choose the decoder's language model on validation (Qwen3-8B-Base for the Platonic JEPA, Qwen2.5-7B otherwise). The curves differ only in the text encoder (Platonic JEPA, bad views of the same sentences, or gte-base) or in the test-time input (the mapped input, or the embedding of the nearest pair's text) (\texttt{paper/figures.py reasoning}, \texttt{data/vqa.csv}). We observe that \emph{Platonic views carry more for reasoning than bad views or a sentence encoder at every number of pairs, and that the mapped input beats copying the text of the nearest pair}.}
\label{fig:reasoning}
\end{figure*}

This section gives the details of \cref{sec:reasoning}.

\paragraph{From captions to answers, on text alone.} The question-answering decoder is the decoder of the captioner of \cref{sec:scale}, with a question added. A text-only language model (Qwen2.5-7B-Instruct) writes five short questions and their answers for each of 139k COCO training captions, in the style of a visual question answering dataset (\texttt{platonic/generate.py}, prompt \texttt{vqa}). We drop non-answers and balance yes and no, which leaves 1.5M questions. A base language model reads a text embedding through ten soft-prompt tokens, then reads the question, and is trained to write the answer. We choose this language model on validation for each text encoder and benchmark, between Qwen2.5-7B and Qwen3-8B-Base. Qwen3-8B-Base serves the Platonic JEPA on VQAv2, and Qwen2.5-7B serves every other case. The decoder never reads the caption, only the frozen text encoder's embedding of the caption (\texttt{scripts/vqa.py}). At test time, the anchor features of a photograph replace this embedding, through the map that our COCO curve fits on $k$ pairs. Hence, no network sees an image with a question or an image with a caption. In contrast, VLMs train on pairs, and so do frozen language models that read images through an adapter \citep{tsimpoukelli2021frozen}.

\paragraph{Benchmarks.} We use VQAv2 \citep{goyal2017vqav2}, scored by its official accuracy and answer normalization. We take its validation questions about the 4{,}700 Karpathy test images (24{,}720 questions). The 1{,}560 questions on the 300 validation images choose every setting. These test and validation images are the only COCO images whose captions no decoder read.

\paragraph{Baselines.} The first baselines keep the same decoder and change only its test-time input. Instead of the mapped input, the decoder reads the embedding of the nearest pair's caption or report, or of the report that a per-statement probe on the $k$ pairs writes. Next, the strongest open language models read the question alone, or the question after those texts. We use Qwen3.8-27B as a text model, and gpt-oss-120b \citep{openai2025gptoss}, a text-only model. A pipeline through gpt-oss-120b thus uses no image--text model. Lastly, we include the VLMs of \cref{sec:vlm}, both out of the box and LoRA-fine-tuned on the questions of $k$ inputs, with the grid and the validation choice of \cref{sec:vlm}.

\paragraph{The decoder answers from what the embedding holds.} We feed the decoder the embedding of the input's own text instead of the mapped input. The decoder then reaches 47.1 on VQAv2, where a caption seldom holds the answer. With no input, the decoder's prior alone gives much less (\cref{tab:reasoning}). Hence, a language model trained on text alone answers questions about inputs it never saw, as well as the embedding it reads allows.

\paragraph{On photographs, 10 pairs beat all 82{,}783 through other text encoders.} Through the alignment anchored on 10 COCO pairs, the decoder reaches 44.1 on VQAv2, with no image--text model anywhere. This score is two thirds of the way from the decoder's prior to its score from the photo's own caption. With all of COCO's pairs, the mapped photo even surpasses its own caption, because a caption says less than the photo shows (\cref{tab:reasoning}). Through the same decoder, the mapped photo beats the nearest pair's caption at every number of pairs, and it even beats our own caption (\cref{tab:vqa_detail}). However, the views decide how far the decoder gets (\cref{fig:reasoning}). We train two other decoders with the same questions and the same two decoder recipes. One reads a bad-view JEPA of the same Wikipedia sentences, and the other reads gte-base, both through the rotation. We choose each text encoder's language model on validation: Qwen3-8B-Base for the Platonic JEPA, and Qwen2.5-7B for the other two, because through Qwen3-8B they read their input barely above their prior. These two decoders gain only a few points from 10 pairs to all 82{,}783 (\cref{tab:vqa_detail}). Our 10-pair score also uses the alignment anchored on unpaired samples, which we fit only into our encoder's space. Through the rotation alone, the three decoders are on par with 10 pairs. With 10 pairs, the Platonic decoder is ahead of both other decoders with all of COCO's pairs (44.1 against at most 43.2; \texttt{data/vqa.csv}). In short, on free text, Platonic views raise what an embedding carries for reasoning, as \cref{sec:phase} predicts. The sentence encoder that retrieves best with many pairs (\cref{sec:scale}) reasons worst.

\paragraph{Against VLMs and a text-only reasoner.} VLMs that read photographs answer far better out of the box (Qwen3.8-27B: 81.5 on the same questions; \cref{tab:reasoning}), because photographs are their regime (\cref{fig:phase}). A strong text-only reasoner also reads our captions well. Given our caption, gpt-oss-120b beats its own score from the question alone with 10 pairs. From 100 pairs on, gpt-oss-120b reading our caption also beats our decoder (47.8 with 100 pairs). With 10 pairs, the nearest pair's caption lowers the score of gpt-oss-120b (\cref{tab:reasoning}). Hence, with no image--text model anywhere, our decoder is the best pipeline we measured with 10 pairs. From 100 pairs on, the best pipeline is our captioner read by a text-only reasoner, because our caption adds what copying cannot. We also give gpt-oss-120b the map's five nearest captions in context, as for open questions. gpt-oss-120b then scores 43.9 with 10 pairs and 48.5 with 100 pairs. With 10 pairs, this score is above that of gpt-oss-120b reading our caption. On validation, however, the decoder stays ahead with 10 pairs, and our caption stays ahead with 100 pairs (\cref{tab:vqa_detail}). Short answers are thus best left to the decoder, and open questions to the in-context reader.

\paragraph{Label questions in six domains.} We follow the protocol of ECG-QA \citep{oh2023ecgqa}. We write verify, choose and query questions from the attributes that each input's annotation states: sleep stage and scored events, wrist activity, lung findings, the simulation's system and parameters, and species, family and order. We score answers by exact match on 3{,}000 to 6{,}000 test questions per domain (\cref{tab:labelqa}; \texttt{scripts/labelqa.py}, \texttt{data/labelqa.csv}). With 10 pairs, our decoder reading the mapped input is ahead of the VLMs on sleep EEG and on simulations (means over the three draws). Our decoder is ahead both of the VLMs out of the box and of the VLMs LoRA-fine-tuned on the questions of the same 10 pairs. The lead is narrow on EEG and clear on simulations. On EEG, Qwen3.8 reading the nearest pair's text scores below our decoder reading the mapped input. Our decoder reading that same text is on par with our decoder reading the mapped input. A model that already reads the modality is ahead, and further ahead once fine-tuned on 10 pairs, as \cref{sec:phase} predicts. Gemma-4-31B leads on wrist motion. Qwen3-Omni-30B, which hears audio, leads on lung sounds, birds, and insects and frogs. We break down the questions about simulations by attribute to show where the resolution of the map ends (\cref{tab:attributes}). The system and its regime come through the mapped input and improve with the pairs. The transport coefficients and the initial conditions vary within a single system. With 10 pairs, these attributes stay close to chance for every reader that goes through text, even when the corpus states them. In contrast, a VLM reads the number of initial shear layers from the rendering, and a classifier on the anchor reads it with 100 pairs.

\paragraph{Open questions about a simulation.} A scientist who trained a model on simulations also wants to ask open questions about what the model sees. We thus follow the CaReSound protocol of CaReAQA \citep{wang2025careaqa}. We take 595 test windows of The Well (85 per system). For each window, gpt-oss-120b reads the ground-truth description and writes five \emph{blind} questions with reference answers. Blind questions are those that an expert who has not seen the input would ask: what the input shows, its regime and what the regime means, how it evolves, how it would change under another condition, and what it implies. The same model writes such questions on the texts of the unpaired corpus. Our decoder (Qwen3-8B-Base, text only) is trained on these corpus questions. Two judges from different model families, gpt-oss-120b and Qwen3.8-27B used as a text model, grade every answer from 0 to 10 against the reference. The grades of the two judges correlate closely, and both judges rank the methods alike (\texttt{scripts/openqa.py}, \texttt{data/openqa\_blind.csv}). \Cref{tab:openqa} gives the results with 10 pairs, and \cref{tab:openqa_detail} gives every method, draw and interval. The in-context reader gives gpt-oss-120b the five corpus descriptions nearest to the mapped simulation, ranked and without scores. This reader scores far above the VLMs out of the box, and far above gpt-oss-120b reading the question alone. The language model of a VLM could also serve as the decoder in the scientific domains. We leave this variant for future work. \Cref{fig:reasoning_examples} shows random questions from this benchmark, with the answers of our in-context reader and of the fine-tuned VLM.

\paragraph{What the input adds.} Much of what these questions ask is generic to a system. Our decoder with no input already scores on par with the fine-tuned VLMs, and the map adds 7 to 9 points to this score. gpt-oss-120b gains less when it reads one caption instead of five hypotheses, or the description of the nearest pair (\cref{tab:openqa_detail}). Some questions have a wording that leaks no fact of the input. On these questions, gpt-oss-120b with the hypotheses still leads our decoder and the VLMs, out of the box or fine-tuned (\cref{tab:openqa_detail}, subset).

\paragraph{Other readers, other prompts.} As the matched control with no input, we put five corpus descriptions drawn at random for each input in the same prompt. With them, gpt-oss-120b scores 36.4 on simulations, against 66.2 with the map's hypotheses (\cref{tab:openqa_detail}; \texttt{scripts/slurm/rand\_hypotheses.sh}). \Cref{tab:icl_readers} compares each open model as an in-context reader of the map's hypotheses with the same model fine-tuned on the pairs. We also tried three prompts from in-context visual question answering on the first draw of the simulations (\texttt{scripts/slurm/icl\_variants.sh}, \cref{tab:icl_variants}). One variant gives each hypothesis its similarity as a confidence, as Prophet does with its answer candidates \citep{shao2023prompting}. This variant gains about one point on validation and on test, also over three draws. Another variant adds a worked example from the nearest labeled pair, with its hypotheses and its true description. PICa and Prophet select in-context examples in the same way \citep{yang2022empirical,shao2023prompting}. This variant loses about three points. A higher reasoning effort does not help. We keep the plain prompt, because the gain from confidences stays within the confidence intervals.

\begin{table}[tbp]
\centering\footnotesize
\caption{\textbf{What the hypotheses state, checked against each input.} We take the first of the five corpus descriptions that the in-context reader receives. We check whether it states the same physical system as the test simulation's own description. The same fixed patterns read the system from both texts (number of systems in parentheses). We report accuracy / balanced accuracy, in percent. Majority: always the most frequent class, whose balanced accuracy is chance. The map: 10 pairs on each of our three draws, and 100 pairs. Nearest pairs: the description of the nearest labelled pair (\texttt{scripts/factcheck.py}, \texttt{data/factcheck.csv}; \texttt{paper/tables\_tex.py}).}
\label{tab:factcheck}
\setlength{\tabcolsep}{3pt}
\begin{tabular}{lccccccc}
\toprule
 & & \multicolumn{3}{c}{the map, 10 pairs} & the map & \multicolumn{2}{c}{nearest pair}\\
\cmidrule(lr){3-5}\cmidrule(lr){7-8}
Domain: what is checked & majority & draw 1 & draw 2 & draw 3 & 100 pairs & 10 pairs & 100 pairs\\
\midrule
Simulations: system (7) & 14 / 14 & 63 / 63 & 84 / 84 & 88 / 88 & 100 / 100 & 57 / 57 & 100 / 100\\
\bottomrule
\end{tabular}
\end{table}

\begin{table}[tbp]
\centering\footnotesize
\caption{\textbf{The same model, in context or fine-tuned.} Test score on blind open questions with 10 pairs, averaged over our three draws of the pairs. Scores run from 0 to 100 and average two judges. The 95\% confidence interval comes from resampling the test inputs, keeping all questions of one input together. In context, each language model reads the map's five nearest corpus descriptions and the question as text, with no training. Fine-tuned, the same model reads the input as a VLM after LoRA fine-tuning on the questions of the same pairs. Our decoder with no input gives the corpus prior (\texttt{scripts/slurm/incontext\_readers.sh}, \texttt{data/openqa\_blind.csv}; \texttt{paper/tables\_tex.py}).}
\label{tab:icl_readers}
\setlength{\tabcolsep}{4pt}
\begin{tabular}{lc}
\toprule
Reader & Simulations\\
\midrule
gpt-oss-120b, in context (\textbf{ours}) & 66.2 [64.8, 67.6]\\
Qwen3.8-27B, in context (\textbf{ours}) & 56.1 [54.5, 57.6]\\
\quad Qwen3.8-27B reading the input, fine-tuned on the pairs & 48.5 [47.1, 49.8]\\
Gemma-4-31B, in context (\textbf{ours}) & 53.7 [52.0, 55.3]\\
\quad Gemma-4-31B reading the input, fine-tuned on the pairs & 51.3 [49.9, 52.6]\\
our decoder, no input & 51.9 [50.7, 53.1]\\
\bottomrule
\end{tabular}
\end{table}

\begin{table}[tbp]
\centering\footnotesize
\caption{\textbf{In-context variants.} gpt-oss-120b answers the blind open questions about simulations with 10 pairs, reading the map's five nearest corpus descriptions with one of four prompts. We report the validation and test scores (0--100, mean of two judges) on the first draw of the pairs and, where run, over three draws. The 95\% confidence interval comes from resampling the test inputs, keeping all questions of one input together (\texttt{scripts/slurm/icl\_variants.sh}, \texttt{data/openqa\_blind.csv}; \texttt{paper/tables\_tex.py}).}
\label{tab:icl_variants}
\setlength{\tabcolsep}{3pt}
\begin{tabular}{lcccc}
\toprule
 & \multicolumn{2}{c}{first draw} & \multicolumn{2}{c}{three draws}\\
\cmidrule(lr){2-3}\cmidrule(lr){4-5}
Prompt & val & test [95\%] & val & test [95\%]\\
\midrule
the map's five nearest descriptions, ranked (\textbf{ours}) & 60.0 & 59.7 [57.3, 62.1] & 67.1 & 66.2 [64.8, 67.6]\\
\quad with each hypothesis's similarity as a confidence & 61.0 & 60.7 [58.3, 63.1] & 67.9 & 67.3 [65.8, 68.6]\\
\quad with one worked example from the nearest pair & 56.5 & 57.0 [54.4, 59.5] & -- & --\\
\quad at high reasoning effort & 59.0 & 59.4 [56.9, 62.0] & -- & --\\
\bottomrule
\end{tabular}
\end{table}

\paragraph{Other text sides, more pairs.} On the simulations' corpus, the text side need not be Platonic. The same reader scores better through gte-base than through our JEPA, on validation as well (\cref{tab:openqa_detail}). The corpus states a few factors of a system in templates. This corpus is thus close to the closed vocabularies of \cref{prop:closed}, where any encoder whose values are affinely independent serves. On the free text of captions, the Platonic JEPA is what lets the decoder read the input (VQAv2, above). With 100 pairs, the pairs' own descriptions suffice. gpt-oss-120b scores higher when it reads the descriptions of the five nearest pairs, or of the nearest one, than when it reads the map's hypotheses (\cref{tab:openqa_detail}).

\paragraph{Open questions about one run.} A scientist also asks about the run in front of them: where the activity is, how it is organized, how it evolves, why, and what comes next. We thus describe each window from its own fields, as an expert annotator would (\texttt{platonic/modalities/physics\_dynamics.py}). The description gives the vorticity magnitude on a $3\times3$ grid of regions (for Gray--Scott, the gradient of its pattern), with banding, coherent structures, trends and drift. gpt-oss-120b writes five blind questions of these kinds per window, with references, on 595 test windows (\texttt{scripts/dynqa.py}, \texttt{data/dynqa.csv}). The text corpus describes 20{,}604 windows of trajectories disjoint from the pool, and no model reads their fields. Our decoder reads 13 text-JEPA embeddings, its \emph{slots}: one for the system, three for whole-field texts and one phrase per region. We map Walrus's mean token and its $3\times3$ region tokens into these slots. The nine regions share one map, fitted on nine region pairs per pair. With 10 pairs (\cref{tab:openqa}), gpt-oss-120b reads the five corpus windows nearest to the mapped window and scores 43.7 over three draws. Our decoder is on par with gpt-oss-120b. The noise added to the decoder's inputs during training is chosen on validation. Qwen3.8-27B and Gemma-4-31B read the rendered window and score at most 32.6, out of the box or fine-tuned on the questions of the same 10 pairs.

\paragraph{What the map adds for one run.} Most of our decoder's lead comes from what it learned on the corpus. With no input, the decoder already holds most of its lead over the VLMs. When the decoder reads the slots of the nearest pair instead of the mapped window, it is nearly as good. The map thus adds 0.5 points over copying the nearest pair with 10 pairs, and 1.1 points with 100 pairs. For gpt-oss-120b, the map's hypotheses add more over the description of the nearest pair (\cref{tab:openqa_detail}). On the questions that ask where the activity is, our decoder leads the same decoder with no input, gpt-oss-120b and the VLMs (\cref{tab:openqa_detail}, subset). With 100 pairs, gpt-oss-120b and a longer-trained decoder, chosen on validation, both improve. When this decoder reads each window's true descriptions, it does far better still. Hence the map, not the reader, limits the decoder. Without its region slots, the decoder loses most on the questions of where (\cref{tab:openqa_detail}).

\clearpage

\clearpage
\section{When Each Method Leads}\label{sec:phase}

\begin{table*}[t]
\centering\footnotesize
\caption{\textbf{Which method to use, and when.} Two quantities place a task: the number of distinct answers it needs, and the number of pairs at hand (\cref{sec:phase} gives the regimes). Each row gives one piece of evidence and the table that holds it. Our photograph results show reasoning without any image--text model, but do not replace such a model (\texttt{paper/tables\_tex.py}).}
\label{tab:guide}
\setlength{\tabcolsep}{4pt}
\begin{tabular}{@{}>{\raggedright\arraybackslash}p{4.3cm}>{\raggedright\arraybackslash}p{5.3cm}>{\raggedright\arraybackslash}p{5.6cm}l@{}}
\toprule
Task and pairs at hand & Method & Evidence & Table\\
\midrule
A VLM already reads the modality (photographs, satellite images, chest X-rays, sounds) & fine-tune that VLM, after checking on a few validation questions & VQAv2: Qwen3.8-27B out of the box 81.5, ours at most 47.5 & \cref{tab:reasoning}\\
Closed vocabulary (labels $\times$ templates at most a text encoder's width plus one), few pairs & skip the search: any text encoder, the frozen anchor and one map & EEG captions, 10 pairs: .424 against .405 for the best fine-tuned VLM & \cref{tab:comparison}\\
Closed vocabulary, more than a few pairs per label & a classifier on the frozen anchor & EEG captions, 300 pairs: logistic regression .599 against .513 for ours & \cref{tab:comparison}\\
Questions on a label set & a decoder trained on the corpus's questions, reading the mapped input & simulations, 10 pairs: 59.4 against 55.1 for the best fine-tuned VLM & \cref{tab:labelqa}\\
An attribute that varies within one system (initial conditions, transport coefficients) & a VLM, if one reads a rendering of the input; otherwise a classifier on the anchor, with about a hundred pairs & initial shear layers: Gemma-4-31B out of the box 91.6, the anchor's probe with 100 pairs 83.1, ours (gpt-oss + map, 100 pairs) 51.4 & \cref{tab:attributes}\\
Free text, a short answer, about ten pairs & search views, then the text-only decoder reading the mapped input & VQAv2: 44.1 with 10 pairs, other text encoders at most 43.2 with all, gpt-oss with the map's five nearest captions in context 43.9 & \cref{tab:vqa_detail}\\
Free text, a short answer, about a hundred pairs & a strong text-only reasoner reading our caption & VQAv2, 100 pairs: 47.8 against 46.8 for the decoder & \cref{tab:vqa_detail}\\
Any task, first: give the in-context reader random corpus descriptions in the same prompt & where the map's hypotheses do no better, the map adds nothing yet: label about a hundred pairs & open questions, simulations: the map's top 5 66.2, five random descriptions 36.4; mass spectra, which the anchor barely aligns: the map's top 5 46.5, five random descriptions 53.6, the question alone 57.0 & \cref{tab:openqa_detail}\\
About ten pairs to label, and a few distinct systems or classes (simulations, wrist activities) & choose the pairs by clustering the frozen anchor, before reading any caption; keep random draws, or more pairs, when some labels are rare (sleep EEG, lung sounds) & open questions on simulations: 75.9 with selected pairs against 66.2 with random draws; captions (\cref{tab:comparison}): wrist motion .332 against .217, sleep EEG .397 against .424; lung-sound label questions (\cref{tab:labelqa}): 50.7 against 53.6 & \cref{tab:openqa}\\
Open questions, about ten pairs & a strong language model (we test three open-weight ones) given the map's five nearest corpus descriptions in context, with no fine-tuning; on templated corpora, any good text encoder as text side & simulations: 66.2 over three draws, against at most 51.3 for VLMs fine-tuned on the same pairs; Qwen3.8-27B: 56.1 in context, 48.5 fine-tuned & \cref{tab:icl_readers}\\
Open questions, about thirty pairs & the same in-context reader: with thirty pairs, the map recognizes the system almost as well as being told it & simulations: 76.3 with 30 pairs and 78.8 with 100, against 77.3 when told the system & \cref{tab:system_control}\\
Open questions, about a hundred pairs & the map's hypotheses; on simulations, the five nearest pairs' descriptions & simulations: 81.0 against 78.5 with the map's hypotheses & \cref{tab:openqa_detail}\\
Tens of thousands of pairs & methods that read only the pairs catch up & COCO, all pairs: the nearest pairs' consensus 97.8 against 92.0 & \cref{tab:coco}\\
\bottomrule
\end{tabular}
\end{table*}

\begin{figure*}[t]
\centering
\includegraphics[width=0.78\textwidth]{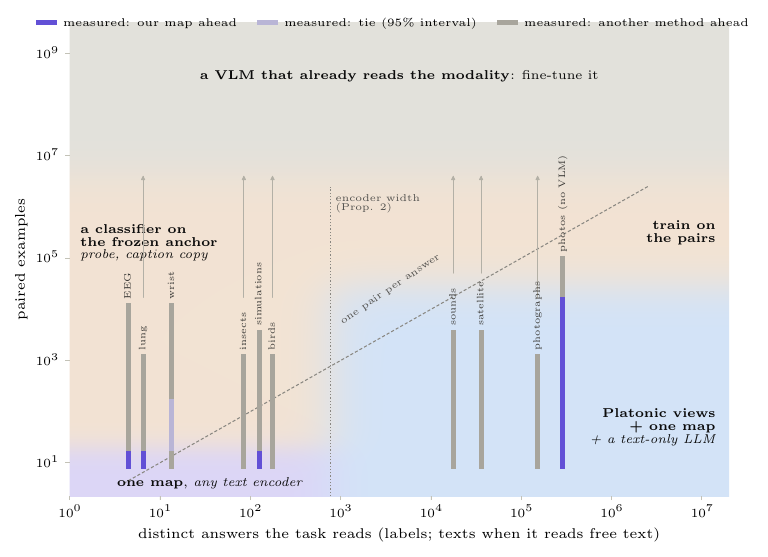}
\caption{\textbf{Which method reads a modality.} The horizontal axis gives the distinct answers a task reads, from a label set said in templates to free text. The vertical axis gives the pairs at hand. On closed vocabularies, every text encoder with affinely independent values is Platonic (\cref{prop:closed}). There, one map fitted on the pairs does the reading until a classifier on the frozen anchor takes over. On free text, Platonic views and one map lead until the pairs suffice to train on. When a VLM already reads the modality, it is the method to fine-tune. Bars: each domain of \cref{fig:phase_evidence} except chest X-rays, over the pairs we measured, in violet where our map was ahead of every other method on test, and in light violet where no method was ahead beyond the 95\% interval (\texttt{paper/phase.py}, \texttt{data/phase\_cells.csv}). Arrows: modalities that existing VLMs were pretrained to read. We observe that \emph{the bars fall where the regimes say}.}
\label{fig:phase}
\end{figure*}

\begin{figure*}[t]
\centering
\includegraphics[width=\textwidth]{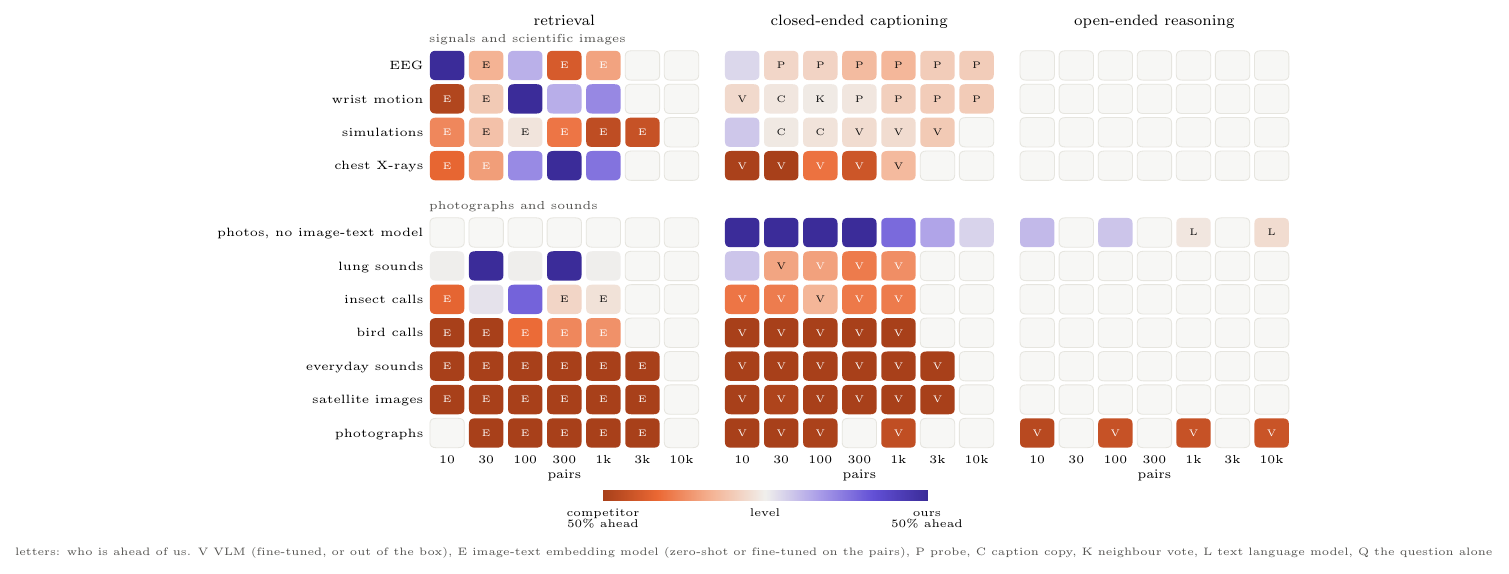}
\caption{\textbf{Who is ahead, task by task.} Our relative margin over the best competitor on test, per domain (rows) and number of pairs (columns). Violet: ours is ahead. The letter names the competitor that is ahead of us (\texttt{data/phase\_margins.csv}). Retrieval competitors: image--text embedding models, zero-shot or fine-tuned on the pairs (\texttt{scripts/retrieval\_baselines.py}). Captioning competitors: fine-tuned VLMs and the methods that read only the pairs (\cref{sec:vlm}). Reasoning: VQAv2 on photographs only, against the competitors of \cref{sec:reasoning} (\cref{tab:labelqa,tab:openqa} give the other domains). We observe that \emph{our map leads where no VLM reads the modality out of the box and pairs are few, and on free text without an image--text model}.}
\label{fig:phase_evidence}
\end{figure*}

Two quantities place a task (\cref{fig:phase}), namely the distinct answers it reads and the pairs at hand; \cref{prop:closed} splits the first, and our measurements the second (\cref{fig:phase_evidence}).

\paragraph{On label sets, one map and any text encoder suffice.} Every label-scored domain of \cref{sec:exp} reads at most 150 answers: 5 sleep stages, 8 lung findings, 10 activities, 85 and 150 species, and 147 simulation configurations. If the paired texts took at most $k+1$ distinct values, \cref{prop:closed} would make the text side one encoder up to a rotation. Our label domains fall in this regime by the answers they read (\cref{fig:phase}). Reports and templates make the distinct texts outnumber the labels, yet four of these corpora also fall in this regime by their distinct texts: lung sounds, EEG, wrist motion, and insect and frog calls hold fewer distinct texts than the 384 dimensions of the search's BERT-S, while bird calls and simulations hold more (\texttt{data/phase\_domains.csv}). Indeed, the search's reward gains only a few percent over bad views on three of the six label domains (lung sounds, wrist motion, and insect and frog calls), against tens of percent on most free-text corpora (\cref{tab:pvs}, gain). Hence, the frozen anchor reads these modalities through one map. With 10 pairs, this map is ahead of the fine-tuned VLMs on EEG, simulations and lung sounds. Other text encoders through the same map are on par or ahead (\cref{tab:comparison}; \texttt{data/comparison.csv}). From a few pairs per label, a classifier on the same anchor takes over (from 30 pairs on EEG and lung sounds, \cref{sec:vlm}). Where a VLM already reads the modality, sounds included, fine-tuning the VLM wins, except on lung sounds with 10 pairs.

\paragraph{On free text, Platonic views come first, then one map.} When the answer is free text, $x_1$ takes more values than the encoder has dimensions, and views decide what the encoder keeps. For example, on Flickr30k, Platonic views roughly double the retrieval of bad views of the same sentences from 30 pairs on (\cref{tab:flickr_pairs}; \texttt{scripts/flickr\_sweep.py}, \texttt{data/flickr\_pairs.csv}). A 109M JEPA trained on Wikipedia alone also aligns with DINOv2 better than every language model we tested (\cref{sec:scale}). With no image--text model anywhere, our captioner is ahead of the strongest baseline that reads only the pairs, from 10 to 10{,}000 COCO pairs (CIDEr 43.0 against 9.6 with 10 pairs). This baseline returns the consensus of the nearest pairs' captions \citep{devlin2015exploring}. Our captioner is on par with it at 30{,}000 pairs and behind with all 82{,}783 (\cref{tab:coco}; \texttt{scripts/coco\_nn.py}, \texttt{data/coco\_nn.csv}). Sentence encoders are text JEPAs trained contrastively on a billion human-paired texts. For all-mpnet-base-v2, these positives include the other captions of the same Flickr30k and COCO photos. These encoders are on par with ours on Flickr30k retrieval up to 100 pairs and ahead from 300 pairs. Without a pair, ours leads on WIT (\cref{fig:models}).

\section{Limitations}\label{sec:limits}

\paragraph{Open questions mostly test what a system implies.} Given the system of an input, five random descriptions of that system answer our open questions as well as the map's hypotheses, except when predicting how one run evolves (\cref{tab:system_control}). Benchmarks of input-specific reasoning remain to be built.

\paragraph{The search needs an anchor and a few hundred pairs.} PVS scores views against a frozen encoder of the source modality, trained without text, on a few hundred paired items. With 25 pairs, bad views still win 10\% of the draws on average and up to 20\% (\cref{tab:pvs}). Within the generic family, rewards are often within noise, except on photographs. The search thus mostly rejects bad views.

\paragraph{Views inherit the anchor's invariances.} An anchor trained to ignore a factor cannot reward views that keep this factor. For example, DINOv2 is trained to be invariant to color jitter and grayscale. On chest X-rays, the findings of the reports align only weakly with RAD-DINO. The language model that writes the views also limits them.

\paragraph{Imagined views follow the language model's prior.} An imagined view keeps what its text states and draws the other details from the language model's prior, which is often stereotypical. For example, the views of one Flickr30k caption describe a typical tennis court, with a yellow shirt in one view and an orange outfit in another, while the photo shows a wall with graffiti (\cref{fig:teaser,tab:gallery}). The Platonic JEPA thus learns to ignore the details that a text leaves open. When the anchor encodes such a detail and a task needs it, the mapped input cannot carry it. When the prior differs from the scenes that the anchor sees, the views can also pull the JEPA towards typical scenes. The reward measures the net effect of both against the anchor, but it does not separate them, and we did not measure each effect on its own.

\paragraph{Caption metrics reward templates.} One fixed report scores above every method on IU X-ray. We thus score chest X-rays by their findings. On AudioCaps, a caption that reads nothing beats ours up to 100 pairs (\cref{tab:curves}). Scored by findings, Platonic views do not help X-ray captions. These views align better with an anchor that barely encodes what the reports describe.

\paragraph{Labels beat captions once pairs reach a few dozen.} Some metrics read a few labels from each text. For these metrics, we train a logistic regression on the frozen encoder, with the labels that the metric parses from the pairs' captions. This classifier leads ours from 30 pairs on EEG and lung sounds. Copying the nearest pair's caption leads on simulations (\cref{sec:vlm}). Captioning by retrieval needs no label schema and returns free text. When the schema is known and pairs number a few dozen, a classifier on the anchor is the stronger method. On these modalities, an off-the-shelf sentence encoder serves as the text side as well as a Platonic JEPA does. The reason is that on a closed vocabulary, every encoder with affinely independent values is Platonic (\cref{prop:closed}).

\paragraph{A billion human pairs still retrieve better.} Without a pair, our Wikipedia JEPA aligns with DINOv2 better than sentence encoders trained on a billion human-made text pairs. Through a rotation fitted on hundreds of image--caption pairs or more, however, these encoders retrieve better (\cref{sec:scale}). The natural next step is thus to start the Platonic JEPA from such an encoder rather than from scratch.

\paragraph{A few pairs are needed.} Without any pair, distribution matching alone barely aligns the unlabeled samples of the two modalities. Retrieval R@1 reaches .012 to .017 on COCO photographs and at most .003 in every other domain (\texttt{data/zeropair.csv}). Every map we report is thus fitted on at least ten pairs. We leave regularized few-pair fits, e.g., bagged or low-rank Procrustes, to future work.

\paragraph{Reasoning is measured by language models.} gpt-oss-120b writes the open questions and their references. gpt-oss-120b is also one of the two judges and our best reader. The other judge, Qwen3.8-27B, grades independently and ranks the methods in the same order. A judge of another kind, e.g., a domain expert, could rank them differently. The references often omit the parameters of an input. The judges can thus reward a fluent answer that names the right system with the wrong parameters. For this reason, \cref{tab:factcheck} also checks the hypotheses against each input's own description. Our decoders owe much to what they learned from the corpus. With no input, the decoder already scores on par with the fine-tuned VLMs on simulation questions. For the per-run decoder, the map adds little over the description of the nearest pair (\cref{tab:openqa,tab:openqa_detail,tab:reasoning}). Lastly, on photographs, VLMs that read them stay far ahead (VQAv2 81.5 against 47.5 for our decoder with all pairs).

\paragraph{Frontier models remain ahead on natural images and sounds.} On COCO, LLaVA-OneVision-7B reaches 136.1 zero-shot. Our captioner reaches 93.0 and needs a rotation fitted on in-domain pairs (\cref{tab:captioner} gives the score with a rotation fitted on Flickr30k pairs). Fine-tuned VLMs also stay ahead on satellite images and chest X-rays. Qwen3-Omni-30B stays ahead on every sound domain, except on lung sounds with 10 pairs.

\paragraph{Assumptions.} The theory assumes finitely many observations and spectral JEPAs with a gap, and we interpret it in a world with injective renderings. The experiments use LeJEPA instead. PVS certifies the world that the modalities share. A nuisance common to the anchor and the target is thus certified as world. What the anchor cannot see, e.g., the sentiment of a review against an image anchor, is treated as nuisance. A language JEPA anchored on images thus learns what texts and images share. Lastly, each candidate costs one JEPA training. Generated views also inherit the generator's knowledge of the world, here that of a text-only 7B language model.

\section{Further Related Work}\label{sec:related}

\paragraph{Multi-view theory.} \Cref{thm:ceiling}(i) is the canonical-correlation bound \citep{hotelling1936relations}, written for whitened encoders aligned by a rotation \citep{schonemann1966generalized}. \Cref{thm:ceiling}(ii) rests on the conditional independence of co-training \citep{blum1998cotraining} and multi-view redundancy \citep{tosh2021contrastive}. JEPAs learn the spectral embedding of their views \citep{haochen2021provable,balestriero2022contrastive}, and good views on both sides align two JEPAs \citep{balestriero2026prh}. Our results run the argument the other way: from a fixed anchor of the source modality back to the views of the target modality.

\paragraph{Language models that write views.} Language models already propose prompts from scores \citep{yang2024opro}. In PVS, they write the positives themselves, and the reward selects them.

\clearpage
\section{Detailed Results}\label{app:results}

The tables below hold the numbers that the main text cites without stating them. \texttt{paper/tables\_tex.py} writes each table from the tracked results. The tables cover the Flickr30k search (\cref{tab:flickr_views}); the Wikipedia JEPAs, every text model on WIT-1024 and their retrieval against the number of pairs (\cref{tab:wiki_scale,tab:wit_models,tab:flickr_pairs}); the COCO captioner (\cref{tab:coco}); the domains, from every curve to the three-draw comparisons and retrieval (\cref{tab:curves,tab:comparison,tab:retrieval}); and question answering (\cref{tab:vqa_detail,tab:labelqa,tab:attributes,tab:openqa_detail}).

\begin{table}[tbp]
\centering\footnotesize
\caption{\textbf{Flickr30k text JEPAs, every measure.} BERT-S text JEPAs trained on the Flickr30k training captions with one LeJEPA recipe (\cref{sec:flickr}). We report the reward (validation halves) and, with DINOv2, the test R@1 (mean, then each seed), R@5 and mutual 10-NN. Means are over the seeds that did not collapse (\texttt{data/flickr.csv}). For the encoders scored, we also report the scene-explained fraction $\rho_f$ and the theory's reward $R$ of \cref{eq:reward}. Both use the whitened top 32 directions of the validation and test pairs (\texttt{data/nuisance.csv}). We also give the Spearman correlation of $R$ with the search's reward and with the test alignment (\texttt{paper/analysis.py theory\_reward}, \texttt{data/theory\_reward.csv}). Below: how often the rewrites copy Flickr30k (\texttt{scripts/recall.py}, \texttt{data/recall.csv}; \texttt{paper/tables\_tex.py}).}
\label{tab:flickr_views}
\setlength{\tabcolsep}{3pt}
\begin{tabular}{lccccccc}
\toprule
Positives (all $\uparrow$) & reward & R@1 & R@1 per seed & R@5 & mNN & $\rho_f$ & $R$\\
\midrule
spans of one caption (bad views) & .157 & .133 & .127 / .140 / .133 & .356 & .132 & .649 & 12.3\\
imagine's texts, bad views & .219 & .187 & .187 & .451 & .169 & .725 & 13.9\\
neighbors, no pair & .164 & .138 & .138 & .379 & .149 & -- & --\\
certified neighbors, bad-view JEPA & .200 & .170 & .170 & .427 & .176 & .721 & 14.0\\
certified neighbors, imagine JEPA & .292 & .265 & .265 / .261 / .269 & .573 & .233 & -- & --\\
LLM: literal rewrite & .234 & .187 & .187 & .445 & .167 & -- & --\\
LLM: paraphrase & .223 & .191 & .191 & .476 & .169 & .724 & 13.5\\
other captions of the photo (oracle) & .277 & .246 & .248 / .260 / .230 & .541 & .232 & .805 & 15.2\\
LLM: another observer & .288 & .272 & .275 / .263 / .279 & .563 & .202 & .778 & 14.7\\
LLM: imagine the photo & .325 & .287 & .285 / .286 / .291 & .609 & .240 & .821 & 15.8\\
LLM: imagine, with details & .343 & .309 & .308 / .320 / .298 & .626 & .246 & .831 & 16.0\\
\midrule
Spearman with $R$ (8 encoders) & .905 & .881 & & & .976 & & \\
\bottomrule
\end{tabular}
\par\medskip
\begin{tabular}{lc}
\toprule
How often the rewrites copy Flickr30k & share\\
\midrule
rewrite: its nearest human training caption (TF-IDF) is another caption of its photo & 7.8\%\\
original caption: its nearest other human training caption is another caption of its photo & 15.7\%\\
rewrites equal to a human caption & 0.008\%\\
\bottomrule
\end{tabular}
\end{table}

\begin{table}[tbp]
\centering\footnotesize
\caption{\textbf{Platonic JEPAs of Wikipedia: views, data, capacity, seeds and anchors.} Top: BERT JEPAs trained on Wikipedia sentences with the recipe of \cref{sec:scale}. Rows vary the positives, the sentences, the model and the epoch. We report WIT-1024 mutual 10-NN at [CLS] with DINOv2-B and DINOv2-g. We also report Flickr30k test R@1 through a rotation fitted on the Flickr30k training pairs (\texttt{data/wiki\_scale.csv}). Bottom: WIT-1024 mutual 10-NN with DINOv2-g and with DINOv3-7B. Each model is read at the token its objective trains (\cref{app:scale_details}; \texttt{data/wit\_anchors.csv}; \texttt{paper/tables\_tex.py}).}
\label{tab:wiki_scale}
\setlength{\tabcolsep}{3pt}
\begin{tabular}{lllcccc}
\toprule
Positives of a sentence & sentences & model & epoch & WIT, DINOv2-B $\uparrow$ & WIT, DINOv2-g $\uparrow$ & Flickr30k R@1 $\uparrow$\\
\midrule
next-token prediction & 10M & BERT-B & 29 & .010 & .012 & .196\\
bad views & 10M & BERT-B & 29 & .137 & .149 & .184\\
one imagined photograph & 10M & BERT-B & 29 & .166 & .182 & .330\\
two imagined photographs & 10M & BERT-B & 29 & .168 & .188 & .355\\
six texts (sentence + five) & 10M & BERT-B & 29 & .164 & .187 & .350\\
one imagined photograph & 90M & BERT-B & 51 & .166 & .187 & .356\\
two, from Google's BERT-B & 10M & BERT-B & 19 & .168 & .188 & .340\\
two, from Google's BERT-L & 10M & BERT-L & 13 & .173 & .192 & .323\\
two, from Google's BERT-L & 10M & BERT-L & 21 & .174 & .193 & .346\\
two, from Google's BERT-L & 10M & BERT-L & 29 & .176 & .195 & .343\\
two imagined photographs, seed 1 & 10M & BERT-B & 29 & .170 & .187 & .352\\
two imagined photographs, seed 2 & 10M & BERT-B & 29 & .167 & .189 & .349\\
two, from Google's BERT-L, seed 1 & 10M & BERT-L & 29 & .177 & .195 & .355\\
two, from Google's BERT-L, seed 2 & 10M & BERT-L & 29 & .169 & .195 & .348\\
\bottomrule
\end{tabular}
\par\medskip
\begin{tabular}{lcc}
\toprule
Text model & WIT, DINOv2-g $\uparrow$ & WIT, DINOv3-7B $\uparrow$\\
\midrule
Platonic views, BERT-L & .195 & .196\\
Platonic views, BERT-B & .188 & .192\\
oracle s0 & .207 & .207\\
oracle s1 & .214 & .210\\
oracle s2 & .213 & .211\\
bad views & .149 & .152\\
next-token & .123 & .130\\
Qwen2.5-72B & .169 & .179\\
Llama-3-70B & .166 & .172\\
Qwen3.5-27B & .164 & .172\\
Gemma-4-31B & .163 & .172\\
\bottomrule
\end{tabular}
\end{table}

\begin{table}[tbp]
\centering\scriptsize
\caption{\textbf{Every text model of \cref{fig:models}.} WIT-1024 mutual 10-NN \citep{huh2024platonic} with DINOv2-B and DINOv2-g, best layer pair: each language model read at the mean of its token states, its best readout (the last token loses .052 on average), each JEPA at its first token, the one its loss trains (\texttt{data/wit\_models.csv}; \texttt{paper/tables\_tex.py}).}
\label{tab:wit_models}
\setlength{\tabcolsep}{3pt}
\begin{tabular}{llccc}
\toprule
Text model & size & readout & DINOv2-B $\uparrow$ & DINOv2-g $\uparrow$\\
\midrule
\multicolumn{5}{l}{\emph{Language models}}\\
\quad Qwen2.5-0.5B & 0.5B & token mean & .102 & .115\\
\quad Qwen2.5-1.5B & 1.5B & token mean & .108 & .124\\
\quad Qwen2.5-3B & 3B & token mean & .125 & .141\\
\quad Qwen2.5-7B & 7B & token mean & .118 & .133\\
\quad Mistral-7B & 7B & token mean & .140 & .157\\
\quad OLMo-3-7B & 7B & token mean & .132 & .149\\
\quad Llama-3-8B & 8B & token mean & .145 & .164\\
\quad Qwen3-8B & 8B & token mean & .129 & .151\\
\quad Llama-3.1-8B & 8B & token mean & .143 & .162\\
\quad Qwen3-14B & 14B & token mean & .134 & .155\\
\quad Qwen2.5-14B & 14B & token mean & .130 & .148\\
\quad Phi-4 & 14B & token mean & .129 & .148\\
\quad gpt-oss-20b & 20B & token mean & .119 & .126\\
\quad GLM-4.7-Flash & 30B & token mean & .074 & .086\\
\quad Qwen3-30B-A3B & 30B & token mean & .128 & .145\\
\quad Qwen2.5-32B & 32B & token mean & .128 & .144\\
\quad OLMo-3-32B & 32B & token mean & .139 & .153\\
\quad Llama-3.1-70B & 70B & token mean & .146 & .162\\
\quad Llama-3-70B & 70B & token mean & .151 & .166\\
\quad Qwen2.5-72B & 72B & token mean & .154 & .169\\
\quad gpt-oss-120b & 120B & token mean & .123 & .133\\
\quad DeepSeek-V4-Flash & 284B & token mean & .085 & .096\\
\multicolumn{5}{l}{\emph{Language models that also read images}}\\
\quad Qwen3.5-0.8B & 0.8B & token mean & .110 & .121\\
\quad Qwen3.5-9B & 9B & token mean & .136 & .156\\
\quad Gemma-4-12B & 12B & token mean & .135 & .153\\
\quad Ministral-3-14B & 14B & token mean & .134 & .154\\
\quad Qwen3.8-27B & 27B & token mean & .139 & .157\\
\quad Qwen3.5-27B & 27B & token mean & .147 & .164\\
\quad Gemma-4-31B & 31B & token mean & .143 & .163\\
\quad Mistral-Small-4 & 119B & token mean & .108 & .115\\
\multicolumn{5}{l}{\emph{Sentence encoders}}\\
\quad gte-base & 109M & token mean & .111 & .126\\
\quad all-mpnet-base-v2 & 109M & token mean & .118 & .131\\
\multicolumn{5}{l}{\emph{\textbf{Ours}: Platonic views, 10M Wikipedia sentences, no image}}\\
\quad BERT-B & 109M & first token ([CLS]) & .168 & .188\\
\quad BERT-L (Google's BERT) & 335M & first token ([CLS]) & .176 & .195\\
\multicolumn{5}{l}{\emph{Oracle: LLaVA recaptions of 940M images}}\\
\quad BERT-B, seed 0 & 109M & first token ([CLS]) & .195 & .207\\
\quad BERT-B, seed 1 & 109M & first token ([CLS]) & .198 & .214\\
\quad BERT-B, seed 2 & 109M & first token ([CLS]) & .200 & .213\\
\multicolumn{5}{l}{\emph{Controls on the same 10M sentences}}\\
\quad bad views (masking) & 109M & first token ([CLS]) & .137 & .149\\
\quad next-token prediction & 109M & token mean & .114 & .123\\
\multicolumn{5}{l}{\emph{Controls on the oracle's 940M texts}}\\
\quad bad views (masking) & 109M & first token ([CLS]) & .103 & .112\\
\quad next-token prediction & 109M & token mean & .109 & .122\\
\bottomrule
\end{tabular}
\end{table}

\begin{table}[tbp]
\centering\footnotesize
\caption{\textbf{Retrieval against the number of pairs, Flickr30k.} Test R@1 through one rotation fitted on $n$ Flickr30k training pairs. We average over three draws of the pairs, and use one draw when all pairs are used. We compare our Wikipedia JEPA, the bad-view JEPA of the same sentences, and two contrastive sentence encoders trained on about a billion human-made text pairs (\texttt{scripts/flickr\_sweep.py}, \texttt{data/flickr\_pairs.csv}; \texttt{paper/tables\_tex.py}).}
\label{tab:flickr_pairs}
\setlength{\tabcolsep}{3pt}
\begin{tabular}{lcccccccc}
\toprule
Text encoder $\uparrow$ & 10 & 30 & 100 & 300 & 1{,}000 & 3{,}000 & 10{,}000 & all (30{,}014)\\
\midrule
Platonic views, BERT-B (ours) & .006 & .018 & .073 & .143 & .234 & .291 & .337 & .355\\
bad views, same sentences & .004 & .009 & .028 & .065 & .103 & .142 & .178 & .184\\
gte-base & .005 & .019 & .073 & .166 & .283 & .388 & .445 & .476\\
all-mpnet-base-v2 & .007 & .019 & .084 & .177 & .308 & .387 & .446 & .481\\
\bottomrule
\end{tabular}
\end{table}

\begin{table}[tbp]
\centering\footnotesize
\caption{\textbf{The COCO captioner, against the pairs and by every metric.} Top: COCO Karpathy CIDEr ($\times$100) against the number of image--caption pairs, with the frozen networks of \cref{fig:captioner_pairs}. Ours: for each number of pairs, the map chosen on validation (\cref{sec:impl}). When the map of fewer pairs is better on validation, we keep it and give its number of pairs in parentheses. We also report the rotation fitted on the pairs only (mean of three draws), and the alignment anchored on the pairs, where each pair weighs $w$ pseudo-pairs. Another baseline is the consensus of the captions of the nearest pairs \citep{devlin2015exploring} (mean of three draws; \texttt{scripts/coco\_nn.py}, \texttt{data/coco\_nn.csv}). VLMs are LoRA-fine-tuned on the same pairs (\texttt{data/results/domains/curves.json}, \texttt{data/domain\_captions.csv}, \texttt{data/domain\_vlm.csv}). Bottom: the stages of \cref{tab:captioner}. Decoding is by consensus over three beams and 32 samples unless stated. We report validation CIDEr, and test CIDEr, BLEU-4 and METEOR ($\times$100) (\texttt{paper/tables\_tex.py}).}
\label{tab:coco}
\setlength{\tabcolsep}{3pt}
\begin{tabular}{rlccccccccc}
\toprule
 & \multicolumn{3}{c}{ours, map chosen on validation} & rotation only & \multicolumn{2}{c}{anchored ($w=10$)} & nearest & \multicolumn{3}{c}{VLM fine-tuned on the pairs}\\
\cmidrule(lr){2-4}\cmidrule(lr){6-7}\cmidrule(lr){9-11}
Pairs & map & val & test & test, 3 draws & val & test & pairs, test & Qwen2.5-VL-7B & Qwen3.8-27B & Gemma-4-31B\\
\midrule
10 & anchored & 45.0 & 43.0 & 12.5 & 45.0 & 43.0 & 9.6 & 130.7 & 128.1 & 127.4\\
30 & anchored & 63.5 & 59.7 & 26.4 & 63.5 & 59.7 & 20.9 & 134.3 & 133.2 & 128.4\\
100 & anchored & 72.0 & 69.1 & 46.8 & 72.0 & 69.1 & 33.8 & 135.2 & 135.5 & 128.7\\
300 & anchored & 75.0 & 73.5 & 64.1 & 75.1 & 73.5 & 48.0 & 135.2 & -- & --\\
1{,}000 & rotation & 81.2 & 79.0 & 78.1 & 80.5 & -- & 61.9 & 139.2 & 140.9 & 134.1\\
3{,}000 & rotation & 91.5 & 85.0 & 84.7 & 85.3 & -- & 73.8 & 137.2 & -- & --\\
10{,}000 & rotation & 94.5 & 87.9 & 89.0 & -- & -- & 83.2 & 139.1 & -- & --\\
30{,}000 & rotation & -- & 91.0 & 90.9 & -- & -- & 91.5 & -- & -- & --\\
82{,}783 & rotation & 99.0 & 92.0 & 92.0 & -- & -- & 97.8 & -- & -- & --\\
\bottomrule
\end{tabular}
\par\medskip
\begin{tabular}{lcccc}
\toprule
Captioner (Qwen2.5-7B decoder, all $\uparrow$) & val CIDEr & test CIDEr & BLEU-4 & METEOR\\
\midrule
DINOv2-g + DINOv3-L, rotation from 29k Flickr30k pairs & 67.0 & 66.4 & 19.3 & 21.4\\
\quad rotation from COCO's 82{,}783 training pairs instead & 93.6 & 89.6 & 26.1 & 24.8\\
\quad\quad + DINOv3-7B read with the other two & 99.0 & 92.0 & 26.7 & 25.2\\
\quad\quad\quad 64 consensus samples instead of 32 & 98.6 & -- & -- & --\\
\quad\quad\quad 128 consensus samples instead of 32 & 100.9 & 93.0 & 27.0 & 25.3\\
\quad\quad DINOv3-7B in place of DINOv3-L & 98.8 & 91.0 & 26.6 & 25.1\\
\bottomrule
\end{tabular}
\end{table}

\begin{table}[tbp]
\centering\scriptsize
\caption{\textbf{Every curve of \cref{fig:vlm_curves}.} Test score of each domain's metric against the number of pairs. Column 0 is out of the box. Column all uses every training pair, and the last column gives their number. Ours: one draw, with the map chosen on validation for each number of pairs, or the map of fewer pairs when it is better on validation. VLMs: LoRA-fine-tuned on the same pairs, with configuration and epoch chosen on validation, under the same rule. No input: the best caption that reads no input. On domains scored by labels, it is the better of one random corpus text per input and the most frequent text. On the other domains, it is the corpus caption most similar to 2{,}000 others (\texttt{data/noinput\_floors.json}) (\texttt{data/results/domains/curves.json}). Satellite images: the four benchmarks pooled, and RSICD alone through the same maps (\texttt{data/domain\_captions.csv}; \texttt{paper/tables\_tex.py}).}
\label{tab:curves}
\setlength{\tabcolsep}{2.5pt}
\begin{tabular}{llcccccccccr}
\toprule
Domain, metric & method & 0 & 10 & 30 & 100 & 300 & 1{,}000 & 3{,}000 & 10{,}000 & all & (all pairs)\\
\midrule
EEG, sleep-stage balanced accuracy & ours & -- & .429 & .477 & .477 & .489 & .539 & .539 & .539 & .539 & 11{,}964\\
 & Qwen3.8-27B & .203 & .405 & .421 & .474 & .596 & -- & -- & -- & -- & \\
 & Gemma-4-31B & .188 & .372 & .430 & -- & -- & -- & -- & -- & -- & \\
& no input & .200 & -- & -- & -- & -- & -- & -- & -- & -- & \\
\midrule
Wrist motion, activity balanced accuracy & ours & -- & .206 & .265 & .343 & .343 & .362 & .433 & .421 & .421 & 81{,}750\\
 & Qwen3.8-27B & .117 & .213 & .245 & .268 & .297 & .404 & -- & -- & -- & \\
 & Gemma-4-31B & .105 & .214 & .249 & .252 & .284 & .319 & -- & -- & -- & \\
& no input & .100 & -- & -- & -- & -- & -- & -- & -- & -- & \\
\midrule
Simulations, system and parameters & ours & -- & .447 & .673 & .720 & .733 & .738 & .746 & -- & .746 & 8{,}400\\
 & Qwen3.8-27B & .267 & .377 & .655 & .708 & .757 & -- & .864 & -- & -- & \\
 & Gemma-4-31B & .220 & .376 & .664 & .722 & .781 & .809 & .864 & -- & -- & \\
& no input & .095 & -- & -- & -- & -- & -- & -- & -- & -- & \\
\midrule
Insect and frog calls, species accuracy & ours & -- & .047 & .088 & .141 & .229 & .314 & -- & -- & .361 & 2{,}270\\
 & Gemma-4-12B & .000 & .014 & .019 & .019 & .022 & .132 & -- & -- & -- & \\
 & Qwen3-Omni-30B & .025 & .058 & .096 & .184 & .393 & .649 & -- & -- & -- & \\
& no input & .012 & -- & -- & -- & -- & -- & -- & -- & -- & \\
\midrule
Bird calls, species accuracy & ours & -- & .010 & .020 & .054 & .094 & .143 & .174 & -- & .174 & 3{,}597\\
 & Gemma-4-12B & .002 & .009 & .011 & .011 & .010 & .017 & -- & -- & -- & \\
 & Qwen3-Omni-30B & .066 & .075 & .104 & .142 & .268 & .481 & -- & -- & -- & \\
& no input & .007 & -- & -- & -- & -- & -- & -- & -- & -- & \\
\midrule
Lung sounds, findings macro-F1 & ours & -- & .242 & .245 & .245 & .245 & .245 & -- & -- & .245 & 2{,}440\\
 & Gemma-4-12B & .011 & .228 & .230 & .230 & .230 & .230 & -- & -- & -- & \\
 & Qwen3-Omni-30B & .118 & .251 & .359 & .462 & .518 & .530 & -- & -- & -- & \\
& no input & .184 & -- & -- & -- & -- & -- & -- & -- & -- & \\
\midrule
Satellite images, four sets pooled, CIDEr & ours & -- & 23.8 & 32.9 & 38.7 & 47.7 & 55.8 & 61.7 & 64.8 & 64.8 & 19{,}643\\
 & Qwen3.8-27B & 8.2 & 58.6 & 60.1 & 84.4 & 114.3 & 142.9 & 163.2 & -- & -- & \\
 & Gemma-4-31B & 27.5 & 59.6 & 63.4 & 89.6 & 123.5 & 145.0 & 155.6 & -- & -- & \\
 & Qwen2.5-VL-7B & 8.1 & 51.7 & 59.0 & 77.9 & 108.6 & 133.9 & 154.6 & -- & 173.6 & \\
& no input & 10.5 & -- & -- & -- & -- & -- & -- & -- & -- & \\
& ours, RSICD only & -- & 11.4 & 16.7 & 23.2 & 32.0 & 41.1 & 46.5 & 46.8 & 46.5 & \\
\midrule
Chest X-rays, CheXbert findings & ours & .055 & .112 & .077 & .147 & .147 & .205 & -- & -- & .225 & 2{,}063\\
 & Qwen3.8-27B & .170 & .221 & .229 & .213 & .248 & .240 & -- & -- & .268 & \\
 & Gemma-4-31B & .198 & .198 & .234 & .140 & .242 & .156 & -- & -- & .208 & \\
 & Qwen2.5-VL-7B & .054 & .038 & .047 & .068 & .072 & .031 & -- & -- & .062 & \\
& no input & .000 & -- & -- & -- & -- & -- & -- & -- & -- & \\
\midrule
Sounds (AudioCaps), CIDEr & ours & -- & 5.7 & 10.2 & 12.5 & 16.1 & 19.7 & 24.0 & 24.5 & 26.2 & 45{,}178\\
 & Gemma-4-12B & 2.6 & 11.5 & 11.5 & 11.5 & 13.7 & 14.7 & 17.7 & -- & -- & \\
 & Qwen3-Omni-30B & 25.7 & 60.7 & 60.7 & 60.7 & 60.7 & 60.7 & 60.7 & -- & -- & \\
& no input & 14.1 & -- & -- & -- & -- & -- & -- & -- & -- & \\
\midrule
Photographs (COCO), CIDEr & ours & -- & 43.0 & 59.7 & 69.1 & 73.5 & 79.0 & 85.0 & 87.9 & 92.0 & 82{,}783\\
 & Qwen3.8-27B & 56.7 & 128.1 & 133.2 & 135.5 & -- & 140.9 & -- & -- & -- & \\
 & Gemma-4-31B & 25.6 & 127.4 & 128.4 & 128.7 & -- & 134.1 & -- & -- & -- & \\
 & Qwen2.5-VL-7B & 19.6 & 130.7 & 134.3 & 135.2 & 135.2 & 139.2 & 137.2 & 139.1 & -- & \\
& no input & 1.0 & -- & -- & -- & -- & -- & -- & -- & -- & \\
\bottomrule
\end{tabular}
\end{table}

\begin{table}[tbp]
\centering\scriptsize
\caption{\textbf{Captioning over three draws of the pairs, every method.} Test score of each domain's metric, averaged over three draws of $k$ pairs. The metrics are sleep-stage balanced accuracy for EEG, balanced accuracy for wrist motion, findings macro-F1 for lung sounds, system and parameters for simulations, and species accuracy for calls. Every choice is made on validation (\texttt{scripts/robustness.py}, \texttt{data/comparison.csv}). Ours: the map of \cref{sec:impl} chosen on validation, or the rotation alone. gte and mpnet: gte-base and all-mpnet-base-v2 as the text side, through the rotation. The anchor without text (\texttt{scripts/baselines.py}) gives three baselines and the best of them on validation. The first copies the caption of the nearest pair. The second takes the label that most of the nearest pairs name. The third is a logistic regression on the labels that the domain's scorer parses from the captions of the pairs. Fine-tuned VLMs: Qwen3.8-27B (sounds: Qwen3-Omni-30B) and Gemma-4-31B ($^\dagger$: one draw). Ours $-$ best VLM: the difference between our mean score and that of the best VLM, over the draws that both ran. Its 95\% confidence interval comes from resampling the test inputs, with both methods scored on the same inputs. Floor: one random corpus text per input, and the most frequent text. Rows \emph{sel.}: the 10 pairs selected with the anchor, ours and the VLMs fine-tuned on these pairs, averaged over three selections (\cref{app:selection}, \texttt{data/comparison\_sel.csv}; \texttt{paper/tables\_tex.py}).}
\label{tab:comparison}
\setlength{\tabcolsep}{2.2pt}
\begin{tabular}{lrcccccccccccc}
\toprule
 & & \multicolumn{2}{c}{\textbf{ours}} & \multicolumn{2}{c}{other text side} & \multicolumn{4}{c}{the anchor without text} & \multicolumn{2}{c}{fine-tuned VLM} & ours $-$ best VLM & floor\\
\cmidrule(lr){3-4}\cmidrule(lr){5-6}\cmidrule(lr){7-10}\cmidrule(lr){11-12}
Domain & pairs & map & rotation & gte & mpnet & nearest & $k$-NN & logistic & best & Qwen & Gemma & [95\% interval] & random / constant\\
\midrule
EEG & 10 & .424 & .375 & .425 & .426 & .389 & .385 & .373 & .382 & .405$^\dagger$ & .331 & +.044 [+.018, +.069] & .200 / .200\\
 & 10 sel. & .397 &  &  &  &  &  &  &  & -- & .366 & +.030 [+.014, +.046] & \\
 & 30 & .475 & .444 & .500 & .494 & .449 & .466 & .509 & .509 & .421$^\dagger$ & .399 & +.078 [+.053, +.105] & \\
 & 100 & .530 & .468 & .528 & .540 & .484 & .525 & .573 & .573 & .457 & -- & +.073 [+.057, +.088] & \\
 & 300 & .513 & .476 & .540 & .540 & .493 & .554 & .599 & .599 & .554 & -- & -.041 [-.056, -.025] & \\
\midrule
Simulations & 10 & .543 & .504 & .510 & .534 & .478 & .478 & .473 & .479 & .477 & .503 & +.040 [+.033, +.046] & .095 / .079\\
 & 10 sel. & .622 &  &  &  &  &  &  &  & .549 & .602 & +.020 [+.016, +.025] & \\
 & 30 & .659 & .662 & .658 & .658 & .672 & .672 & .667 & .672 & .655$^\dagger$ & .639 & +.018 [+.009, +.027] & \\
 & 100 & .718 & .714 & .723 & .725 & .743 & .743 & .739 & .740 & .708$^\dagger$ & .721 & -.004 [-.009, +.002] & \\
 & 300 & .739 & .733 & .745 & .746 & .780 & .778 & .773 & .776 & -- & .780 & -.040 [-.046, -.035] & \\
\midrule
Lung sounds & 10 & .231 & .251 & .256 & .253 & .193 & .202 & .213 & .214 & .180 & -- & +.051 [+.037, +.069] & .184 / .118\\
 & 10 sel. & .236 &  &  &  &  &  &  &  & .182 & -- & +.054 [+.040, +.068] & \\
 & 30 & .232 & .248 & .257 & .253 & .238 & .229 & .280 & .280 & .289 & -- & -.058 [-.076, -.041] & \\
 & 100 & .304 & .253 & .267 & .264 & .293 & .290 & .365 & .365 & .383 & -- & -.079 [-.100, -.058] & \\
 & 300 & .356 & .265 & .276 & .263 & .332 & .328 & .411 & .411 & .502 & -- & -.146 [-.170, -.122] & \\
\midrule
Wrist motion & 10 & .217 & .211 & .209 & .212 & .215 & .215 & .200 & .215 & .231 & .214$^\dagger$ & -.014 [-.022, -.006] & .099 / .100\\
 & 10 sel. & .332 &  &  &  &  &  &  &  & .234 & -- & +.097 [+.088, +.108] & \\
 & 30 & .282 & .296 & .297 & .300 & .289 & .289 & .279 & .292 & .244$^\dagger$ & .254 & +.028 [+.019, +.038] & \\
 & 100 & .338 & .341 & .362 & .361 & .342 & .343 & .330 & .343 & .277 & .268 & +.061 [+.051, +.072] & \\
 & 300 & .394 & .374 & .394 & .390 & .356 & .369 & .405 & .405 & .341 & .283$^\dagger$ & +.053 [+.043, +.064] & \\
\midrule
Bird calls & 10 & .014 & .015 & .015 & .016 & .014 & .014 & .015 & .014 & .074 & -- & -.061 [-.068, -.053] & .006 / .007\\
 & 10 sel. & .020 &  &  &  &  &  &  &  & .098 & -- & -.078 [-.086, -.070] & \\
 & 30 & .025 & .027 & .027 & .027 & .021 & .021 & .024 & .024 & .108 & -- & -.084 [-.092, -.075] & \\
 & 100 & .057 & .055 & .054 & .053 & .036 & .038 & .049 & .047 & .143 & -- & -.086 [-.097, -.076] & \\
 & 300 & .119 & .087 & .091 & .080 & .071 & .067 & .093 & .089 & .264 & -- & -.145 [-.158, -.133] & \\
\midrule
Insects, frogs & 10 & .044 & .041 & .044 & .044 & .039 & .038 & .041 & .039 & .063 & -- & -.019 [-.026, -.012] & .012 / .012\\
 & 10 sel. & .049 &  &  &  &  &  &  &  & .055 & -- & -.006 [-.012, -.001] & \\
 & 30 & .077 & .080 & .080 & .078 & .074 & .074 & .071 & .073 & .108 & -- & -.031 [-.040, -.022] & \\
 & 100 & .160 & .142 & .141 & .137 & .123 & .124 & .134 & .127 & .190 & -- & -.030 [-.043, -.018] & \\
 & 300 & .275 & .219 & .226 & .209 & .187 & .186 & .245 & .245 & .391 & -- & -.116 [-.132, -.101] & \\
\bottomrule
\end{tabular}
\end{table}

\begin{table}[tbp]
\centering\scriptsize
\caption{\textbf{Retrieval against the pairs.} Text-to-item recall at 10 on the test items. Each query is the first caption of a test item (\texttt{scripts/retrieval\_ours.py}). Ours goes through the map chosen on validation. Open embedding models are zero-shot (0) or fine-tuned on the same pairs with their own losses. Their learning rate and checkpoint are chosen on validation (\texttt{scripts/retrieval\_baselines.py}; \texttt{data/results/domains/retrieval\_curves.json}; \texttt{paper/tables\_tex.py}).}
\label{tab:retrieval}
\setlength{\tabcolsep}{3pt}
\begin{tabular}{lclccccccccr}
\toprule
Domain & chance & model & 0 & 10 & 30 & 100 & 300 & 1{,}000 & 3{,}000 & all & (all pairs)\\
\midrule
EEG & .142 & \textbf{ours} & -- & .244 & .244 & .220 & .195 & .195 & .146 & -- & --\\
 &  & Qwen3-VL-Embedding-8B & .195 & .098 & .146 & .195 & .268 & .146 & -- & -- & \\
 &  & SigLIP 2 g & .146 & .122 & .293 & .146 & .317 & .244 & -- & -- & \\
\midrule
Wrist motion & .079 & \textbf{ours} & -- & .095 & .162 & .248 & .248 & .276 & .305 & -- & --\\
 &  & Qwen3-VL-Embedding-8B & .067 & .181 & .181 & .143 & .219 & .229 & -- & -- & \\
 &  & SigLIP 2 g & .010 & .067 & .095 & .133 & .143 & .190 & -- & -- & \\
\midrule
Simulations & .062 & \textbf{ours} & -- & .201 & .357 & .448 & .416 & .409 & .474 & -- & --\\
 &  & Qwen3-VL-Embedding-8B & .221 & .273 & .409 & .461 & .591 & .734 & .825 & -- & \\
 &  & SigLIP 2 g & .136 & .201 & .364 & .422 & .597 & .695 & .753 & -- & \\
\midrule
Chest X-rays & .025 & \textbf{ours} & -- & .053 & .055 & .088 & .151 & .173 & -- & .214 & 2{,}063\\
 &  & Qwen3-VL-Embedding-8B & .055 & .053 & .070 & .073 & .098 & .138 & -- & -- & \\
 &  & SigLIP 2 g & .070 & .080 & .065 & .058 & .063 & .045 & -- & -- & \\
\midrule
Lung sounds & .232 & \textbf{ours} & -- & .462 & .462 & .462 & .462 & .462 & -- & .462 & 2{,}440\\
 &  & PE-AV-L & .308 & .462 & .231 & .462 & .308 & .462 & -- & -- & \\
\midrule
Bird calls & .065 & \textbf{ours} & -- & .047 & .160 & .273 & .467 & .660 & .767 & -- & --\\
 &  & PE-AV-L & .207 & .213 & .393 & .407 & .633 & .867 & -- & -- & \\
\midrule
Insects, frogs & .112 & \textbf{ours} & -- & .200 & .447 & .729 & .741 & .953 & -- & .965 & 2{,}270\\
 &  & PE-AV-L & .247 & .306 & .435 & .565 & .800 & .988 & -- & -- & \\
\midrule
Satellite images & .003 & \textbf{ours} & -- & .027 & .039 & .070 & .101 & .127 & .156 & -- & --\\
 &  & Qwen3-VL-Embedding-8B & .352 & -- & .374 & .394 & .464 & .530 & .580 & -- & \\
 &  & SigLIP 2 g & .302 & .399 & .409 & .416 & .437 & .475 & .516 & -- & \\
\midrule
Photographs & .002 & \textbf{ours} & -- & .107 & .216 & .274 & .293 & .327 & .365 & -- & --\\
 &  & Qwen3-VL-Embedding-8B & .861 & -- & .884 & .876 & .895 & -- & .896 & -- & \\
 &  & SigLIP 2 g & .844 & -- & -- & .895 & .895 & .892 & .883 & -- & \\
\midrule
Sounds & .011 & \textbf{ours} & -- & .031 & .076 & .114 & .225 & .260 & .313 & -- & --\\
 &  & PE-AV-L & .841 & .841 & .845 & .839 & .847 & .896 & .906 & -- & \\
\bottomrule
\end{tabular}
\end{table}

\begin{table}[tbp]
\centering\footnotesize
\caption{\textbf{VQAv2 through every text encoder and input.} Test VQA accuracy, with no image--text model anywhere. Rows use the question-answering decoder of \cref{sec:reasoning}, trained on text alone through each text encoder. The row's title names its language model. At test time, the decoder reads the input mapped through $k$ pairs, or the embedding of a text. No input: the decoder's prior. Own caption: the embedding of the photo's own caption. We also show text-only language models that read the question alone (no input) or after a caption (\texttt{scripts/vqa.py}, \texttt{data/vqa.csv}; \texttt{paper/tables\_tex.py}).}
\label{tab:vqa_detail}
\setlength{\tabcolsep}{3pt}
\begin{tabular}{lccccccc}
\toprule
Decoder input (all $\uparrow$) & no input & 10 & 100 & 1{,}000 & 10{,}000 & 82{,}783 & own caption\\
\midrule
\multicolumn{8}{l}{\emph{Platonic JEPA, Qwen3-8B-Base (\textbf{ours})}}\\
\quad mapped input (map chosen on validation) & 37.8 & 44.1 & 46.8 & 46.8 & 47.5 & 47.5 & 47.1\\
\quad mapped input, rotation only &  & 40.5 & 45.1 & 46.8 & 47.5 & 47.5 & \\
\quad the nearest pair's caption &  & 39.9 & 43.2 & 44.7 & 45.6 & 45.9 & \\
\quad our caption &  & 43.6 & 46.4 & 46.4 & 47.0 & 47.1 & \\
\multicolumn{8}{l}{\emph{bad views of the same sentences, Qwen2.5-7B}}\\
\quad mapped input (map chosen on validation) & 37.5 & 40.0 & 42.3 & 43.0 & 43.1 & 43.2 & 42.7\\
\quad the nearest pair's caption &  & 39.8 & 41.4 & 41.9 & 42.3 & 42.5 & \\
\quad our caption &  & 41.6 & 42.3 & 42.5 & 42.6 & 42.4 & \\
\multicolumn{8}{l}{\emph{gte-base, Qwen2.5-7B}}\\
\quad mapped input (map chosen on validation) & 37.8 & 40.0 & 40.8 & 41.2 & 41.3 & 41.3 & 41.5\\
\quad the nearest pair's caption &  & 39.4 & 40.2 & 40.6 & 40.8 & 40.7 & \\
\quad our caption &  & 40.5 & 40.8 & 41.0 & 41.0 & 41.0 & \\
\multicolumn{8}{l}{\emph{Platonic JEPA, Qwen2.5-7B}}\\
\quad mapped input (map chosen on validation) & 37.2 & 43.0 & 45.1 & 45.3 & 45.5 & 45.5 & 45.0\\
\multicolumn{8}{l}{\emph{bad views, Qwen3-8B-Base}}\\
\quad mapped input (map chosen on validation) & 37.8 & 39.4 & 40.0 & 40.1 & 40.1 & 40.2 & --\\
\multicolumn{8}{l}{\emph{gte-base, Qwen3-8B-Base}}\\
\quad mapped input (map chosen on validation) & 38.8 & 39.4 & 39.4 & 39.6 & 39.5 & -- & --\\
\multicolumn{8}{l}{\emph{text-only language models, no image--text model}}\\
\quad gpt-oss-120b, our caption & 41.1 & 43.0 & 47.8 & 48.9 & 50.4 & 51.2 & \\
\quad gpt-oss-120b, the nearest pair's caption & 41.1 & 36.6 & 41.2 & 45.7 & 47.2 & 49.4 & \\
\quad gpt-oss-120b, the map's five nearest captions, in context & 41.1 & 43.9 & 48.5 & -- & -- & -- & \\
\quad Qwen3.8-27B as a text model, our caption & 42.8 & 45.2 & 50.6 & 51.2 & 53.0 & 53.8 & \\
\quad Qwen3.8-27B as a text model, the nearest pair's caption & 42.8 & 38.2 & 43.2 & 48.1 & 50.1 & 52.0 & \\
\bottomrule
\end{tabular}
\end{table}

\begin{table}[tbp]
\centering\scriptsize
\caption{\textbf{Label questions in six domains.} Test exact match on verify, choose and query questions written from each input's annotation (\texttt{scripts/labelqa.py}, \texttt{data/labelqa.csv}). We use 10 pairs unless stated, and average over the draws run. No input: gpt-oss-120b answers from the question alone, and our decoder answers from its prior. Text-only readers: Qwen3.8-27B reads the text of the nearest pair (NN), and gpt-oss-120b reads our caption. Our decoder reads the text of the nearest pair or the mapped input (10 or 100 pairs). VLMs read the input, out of the box (zs) or LoRA-fine-tuned on the questions of our 10 pairs (ft, mean over the three draws; $^\dagger$: on one draw of 10 pairs other than ours). Rows \emph{selected pairs}: the 10 pairs selected with the anchor, averaged over three selections, with the rotation or the ridge map, and the VLMs fine-tuned on these pairs (\cref{app:selection}); --: not run (\texttt{paper/tables\_tex.py}).}
\label{tab:labelqa}
\setlength{\tabcolsep}{2pt}
\begin{tabular}{lcccccccccccl}
\toprule
 & \multicolumn{2}{c}{no input} & \multicolumn{2}{c}{text-only reader} & \multicolumn{3}{c}{our decoder} & \multicolumn{2}{c}{Qwen VLM} & \multicolumn{2}{c}{Gemma VLM} & \\
\cmidrule(lr){2-3}\cmidrule(lr){4-5}\cmidrule(lr){6-8}\cmidrule(lr){9-10}\cmidrule(lr){11-12}
Domain & gpt-oss & decoder & Qwen3.8 + NN & gpt-oss + caption & NN & \textbf{ours} (10) & \textbf{ours} (100) & zs & ft & zs & ft & VLMs\\
\midrule
Sleep EEG & 48.0 & 45.9 & 57.9 & 51.8 & 64.0 & 63.9 & 72.9 & 49.4 & 58.5 & 53.8 & 62.3 & Qwen3.8-27B, Gemma-4-31B\\
\quad selected pairs & & & -- & -- & 65.1 & 61.7 & & & -- & & -- & \\
Wrist motion & 27.7 & 29.0 & 35.6 & 37.7 & 36.7 & 36.0 & 40.3 & 37.3 & 42.3$^\dagger$ & 41.5 & 44.1$^\dagger$ & Qwen3.8-27B, Gemma-4-31B\\
\quad selected pairs & & & -- & -- & 44.9 & 39.6 & & & -- & & -- & \\
Simulations & 28.2 & 32.4 & 46.6 & 52.0 & 55.5 & 59.4 & 70.3 & 51.8 & 55.1 & 46.9 & 53.4 & Qwen3.8-27B, Gemma-4-31B\\
\quad selected pairs & & & -- & -- & 62.8 & 62.5 & & & -- & & -- & \\
Lung sounds & 23.2 & 24.8 & 50.7 & 50.6 & 68.2 & 53.6 & 62.8 & 67.3 & 76.1$^\dagger$ & 23.0 & 76.2$^\dagger$ & Qwen3-Omni-30B, Gemma-4-12B\\
\quad selected pairs & & & -- & -- & 63.0 & 50.7 & & & -- & & -- & \\
Bird calls & 33.4 & 42.5 & 37.8 & 37.0 & 35.6 & 39.6 & 46.1 & 58.5 & 59.0$^\dagger$ & 35.2 & -- & Qwen3-Omni-30B, Gemma-4-12B\\
\quad selected pairs & & & -- & -- & 41.1 & 41.8 & & & -- & & -- & \\
Insects, frogs & 37.3 & 43.6 & 45.6 & 46.4 & 43.3 & 46.7 & 59.1 & 56.7 & 62.2$^\dagger$ & 40.8 & -- & Qwen3-Omni-30B, Gemma-4-12B\\
\quad selected pairs & & & -- & -- & 48.4 & 50.5 & & & -- & & -- & \\
\bottomrule
\end{tabular}
\end{table}

\begin{table}[tbp]
\centering\scriptsize
\caption{\textbf{What questions about a simulation recover, attribute by attribute.} Test exact match on the choose questions of each attribute (one of three values, or none). Our decoder reads the system, the Gray-Scott pattern and the viscoelastic state from the mapped input, and these scores improve with the pairs. The transport coefficients stay near chance for every reader with 10 pairs, the VLMs included. These coefficients come only partly within reach with 100 pairs. The corpus states the number of initial shear layers, but this number varies within one system. A VLM reads it from the rendering, and the anchor's probe reads it with 100 pairs. The readers that go through the map gain little on it. The corpus never states the shear-layer width or the tracer blobs, and no reader recovers them. We average over the draws of the pairs. VLMs are out of the box (zs) or LoRA-fine-tuned on the questions of our 10 pairs (ft) (\texttt{scripts/labelqa.py breakdown}, \texttt{data/labelqa\_attributes.csv}).}
\label{tab:attributes}
\setlength{\tabcolsep}{3pt}
\begin{tabular}{llccccccccc}
\toprule
 & & & \multicolumn{2}{c}{our decoder} & \multicolumn{2}{c}{gpt-oss + top 5} & \multicolumn{2}{c}{anchor probe} & \multicolumn{2}{c}{best VLM} \\
\cmidrule(lr){4-5}\cmidrule(lr){6-7}\cmidrule(lr){8-9}\cmidrule(lr){10-11}
Kind & Attribute & chance & 10 & 100 & 10 & 100 & 10 & 100 & zs & ft (10) \\
\midrule
System & physical system & 33.3 & 87.8 & 99.0 & 66.3 & 92.2 & 78.8 & 100.0 & 71.3 & 77.9\\
\addlinespace
Regime & Gray-Scott pattern & 33.3 & 35.2 & 76.7 & 30.1 & 32.2 & 35.4 & 92.3 & 56.6 & 59.7\\
 & viscoelastic state & 33.3 & 43.8 & 84.7 & 20.9 & 57.5 & 35.5 & 86.3 & 41.8 & 36.4\\
 & boundary condition & 33.3 & 49.7 & 47.3 & 22.7 & 46.4 & 57.6 & 98.2 & 78.2 & 62.1\\
\addlinespace
Transport coefficient & Reynolds number & 33.3 & 36.1 & 39.1 & 33.7 & 33.7 & 35.0 & 47.6 & 36.7 & 32.0\\
 & Rayleigh number & 33.3 & 44.8 & 44.1 & 29.3 & 42.4 & 28.3 & 64.6 & 41.4 & 30.0\\
 & Prandtl number & 33.3 & 43.1 & 44.1 & 29.2 & 22.9 & 21.5 & 31.3 & 42.7 & 37.8\\
 & Schmidt number & 33.3 & 43.8 & 44.9 & 38.2 & 27.0 & 30.7 & 40.1 & 33.7 & 34.5\\
 & adiabatic index & 33.3 & 44.4 & 39.1 & 16.1 & 20.4 & 20.1 & 19.7 & 44.1 & 39.4\\
 & cooling time & 33.3 & 38.1 & 60.0 & 27.1 & 31.4 & 13.6 & 37.9 & 43.6 & 34.5\\
\addlinespace
Initial condition & initial shear layers & 33.3 & 42.2 & 51.3 & 42.3 & 51.4 & 54.7 & 83.1 & 91.6 & 72.0\\
 & initial shear-layer width & 33.3 & 39.0 & 45.5 & 19.5 & 21.3 & 29.5 & 50.1 & 43.7 & 43.7\\
 & initial tracer blobs & 33.3 & 43.7 & 45.2 & 22.7 & 30.8 & 26.7 & 46.5 & 37.8 & 38.5\\
\bottomrule
\end{tabular}
\end{table}

\begin{table}[tbp]
\centering\scriptsize
\caption{\textbf{Open questions, every method.} Test score on blind questions about each input (\texttt{scripts/openqa.py}, \texttt{data/openqa\_blind.csv}; \texttt{scripts/dynqa.py}, \texttt{data/dynqa.csv}). Scores run from 0 to 100 and average two judges. Columns give the score on the first draw of the pairs, on each of the other two draws, and averaged over the three draws. The first-draw and three-draw scores come with a 95\% confidence interval from resampling the test inputs, keeping all questions of one input together. Further columns give a subset of the questions, the validation score (both averaged over the draws when there are three), and the correlation of the two judges' grades. VLMs are LoRA-fine-tuned on the questions of the same 10 pairs as ours, on each of our three draws, unless marked. Our decoder uses the map chosen on validation over the three draws, either the rotation or the ridge map (\cref{tab:openqa}; \texttt{paper/tables\_tex.py}).}
\label{tab:openqa_detail}
\setlength{\tabcolsep}{2.5pt}
\begin{tabularx}{\linewidth}{@{}>{\raggedright\arraybackslash}Xccccccc@{}}
\toprule
Method & pairs &
\shortstack{first draw\\{}[95\%]} &
\shortstack{draws\\2 / 3} &
\shortstack{mean of 3\\{}[95\%]} &
subset & val & judges\\
\midrule
\multicolumn{8}{@{}l@{}}{\emph{Simulations} (subset: questions whose wording leaks no fact of the input)}\\
\quad Qwen3.8-27B, out of the box & 0 & 35.8 [34.3, 37.3] & -- & -- & 26.8 & -- & 0.91\\
\quad Gemma-4-31B, out of the box & 0 & 36.6 [35.3, 37.8] & -- & -- & 27.5 & -- & 0.91\\
\quad Qwen3.8-Flash-Next (Qwen4 architecture, 177B), out of the box & 0 & 34.3 [33.0, 35.7] & -- & -- & 26.0 & -- & 0.91\\
\quad gpt-oss-120b, question alone & 0 & 37.1 [35.7, 38.4] & -- & -- & 18.2 & 38.1 & 0.93\\
\quad our decoder, no input & 0 & 51.9 [50.7, 53.1] & -- & -- & 32.8 & 52.9 & 0.93\\
\quad Qwen3.8-27B, fine-tuned on the same 10 pairs & 10 & 44.3 [42.1, 46.4] & 50.1 / 51.0 & 48.5 [47.1, 49.8] & 43.4 & -- & 0.93\\
\quad Gemma-4-31B, fine-tuned on the same 10 pairs & 10 & 44.6 [42.4, 46.7] & 52.3 / 57.0 & 51.3 [49.9, 52.6] & 46.3 & -- & 0.94\\
\quad Qwen3.8-Flash-Next (Qwen4 architecture, 177B), fine-tuned on the same 10 pairs & 10 & 47.2 [45.2, 49.0] & 49.8 / 53.9 & 50.3 [48.9, 51.6] & 44.6 & -- & 0.94\\
\quad our decoder, mapped input (rotation) & 10 & 59.1 [57.5, 60.5] & 61.1 / 62.9 & 61.0 [59.8, 62.3] & 49.8 & 62.5 & 0.92\\
\quad our decoder, the nearest pair's description & 10 & 57.6 [56.1, 59.0] & 61.0 / 60.9 & 59.9 [58.7, 61.1] & 47.7 & 60.8 & 0.92\\
\quad gpt-oss-120b + our caption & 10 & 55.4 [53.1, 57.8] & -- & -- & 52.0 & 57.7 & 0.93\\
\quad gpt-oss-120b + the nearest pair's description & 10 & 52.7 [50.1, 55.2] & -- & -- & 48.7 & 53.7 & 0.93\\
\quad gpt-oss-120b + the five nearest pairs' descriptions & 10 & 47.7 [45.8, 49.8] & -- & -- & 36.5 & 48.9 & 0.93\\
\quad gpt-oss-120b + the map's five nearest descriptions (\textbf{ours}) & 10 & 59.7 [57.3, 62.1] & 68.6 / 70.4 & 66.2 [64.8, 67.6] & 63.2 & 67.1 & 0.92\\
\quad \quad the same through gte-base & 10 & 61.1 [58.7, 63.3] & 74.0 / 72.0 & 69.0 [67.7, 70.2] & 66.5 & 71.2 & 0.91\\
\quad \quad five random corpus descriptions instead (no input) & 10 & 36.8 [34.8, 38.8] & 36.5 / 36.0 & 36.4 [34.7, 38.1] & 21.1 & 37.0 & 0.95\\
\quad Qwen3.8-27B, fine-tuned on the same 30 pairs & 30 & 65.3 [64.0, 66.4] & 62.3 / 64.9 & 64.2 [63.3, 65.0] & 59.8 & -- & 0.93\\
\quad Gemma-4-31B, fine-tuned on the same 30 pairs & 30 & 62.3 [60.9, 63.7] & 65.0 / 62.4 & 63.2 [62.2, 64.3] & 59.2 & -- & 0.92\\
\quad our decoder, mapped input (rotation) & 30 & 62.4 [61.0, 63.7] & -- & -- & 52.7 & 63.5 & 0.90\\
\quad our decoder, the nearest pair's description & 30 & 64.0 [62.8, 65.3] & -- & -- & 55.0 & 65.3 & 0.89\\
\quad gpt-oss-120b + the map's five nearest descriptions (\textbf{ours}) & 30 & 77.1 [75.8, 78.4] & 76.1 / 75.7 & 76.3 [75.3, 77.4] & 73.3 & 77.9 & 0.90\\
\quad Qwen3.8-27B, fine-tuned on the same 100 pairs & 100 & 67.3 [66.1, 68.5] & 70.4 / 69.3 & 69.0 [68.1, 69.9] & 65.3 & -- & 0.92\\
\quad Gemma-4-31B, fine-tuned on the same 100 pairs & 100 & 69.9 [68.8, 71.0] & 69.5 / 70.7 & 70.0 [69.0, 71.0] & 67.1 & -- & 0.92\\
\quad our decoder, mapped input (rotation) & 100 & 63.6 [62.3, 64.8] & 63.3 / 63.5 & 63.5 [62.2, 64.7] & 54.1 & 64.5 & 0.91\\
\quad our decoder, the nearest pair's description & 100 & 64.0 [62.7, 65.3] & 64.2 / 63.9 & 64.0 [62.8, 65.2] & 55.1 & 65.2 & 0.90\\
\quad gpt-oss-120b + our caption & 100 & 77.1 [75.7, 78.4] & -- & -- & 73.2 & 80.2 & 0.88\\
\quad gpt-oss-120b + the nearest pair's description & 100 & 79.8 [78.6, 81.1] & -- & -- & 77.2 & 82.1 & 0.88\\
\quad gpt-oss-120b + the five nearest pairs' descriptions & 100 & 81.0 [79.9, 82.0] & -- & -- & 79.3 & 83.8 & 0.87\\
\quad gpt-oss-120b + the map's five nearest descriptions (\textbf{ours}) & 100 & 78.5 [77.4, 79.7] & 79.2 / 78.7 & 78.8 [77.9, 79.7] & 76.5 & 78.7 & 0.89\\
\multicolumn{8}{@{}l@{}}{\emph{Mass spectra} (the anchor barely aligns with text: retrieval R@1 2 to 3\%)}\\
\quad gpt-oss-120b, question alone & 0 & 57.0 [54.6, 59.4] & -- & -- & 52.3 & 56.6 & 0.88\\
\quad gpt-oss-120b + the nearest pair's description & 10 & 38.5 [35.3, 41.3] & -- & -- & 28.7 & 40.2 & 0.88\\
\quad gpt-oss-120b + the map's five nearest descriptions (\textbf{ours}) & 10 & 45.7 [42.9, 48.5] & 46.7 / 47.0 & 46.5 [43.9, 49.1] & 38.6 & 47.3 & 0.91\\
\quad gpt-oss-120b + the map's five nearest descriptions (\textbf{ours}) & 100 & 49.2 [46.5, 52.0] & -- & -- & 41.8 & 48.4 & 0.89\\
\quad \quad five random corpus descriptions instead (no input) & 10 & 54.1 [51.7, 56.7] & 53.4 / 53.2 & 53.6 [51.3, 55.9] & 47.9 & 53.8 & 0.91\\
\multicolumn{8}{@{}l@{}}{\emph{One run's dynamics} (subset: questions of where the activity is)}\\
\quad Qwen3.8-27B, out of the box & 0 & 31.2 [30.0, 32.5] & -- & -- & 16.9 & -- & 0.89\\
\quad Gemma-4-31B, out of the box & 0 & 29.6 [28.5, 30.6] & -- & -- & 13.7 & -- & 0.89\\
\quad Qwen3.8-27B, fine-tuned on the same 10 pairs & 10 & 29.7 [28.5, 30.9] & 34.3 / 33.9 & 32.6 [31.8, 33.5] & 26.2 & -- & 0.88\\
\quad Gemma-4-31B, fine-tuned on the same 10 pairs & 10 & 28.4 [27.3, 29.5] & 30.1 / 31.1 & 29.9 [29.2, 30.7] & 24.2 & -- & 0.85\\
\quad gpt-oss-120b, question alone & 0 & 25.2 [24.4, 26.0] & -- & -- & 10.2 & 26.2 & 0.92\\
\quad our decoder, no input & 0 & 34.5 [33.3, 35.7] & -- & -- & 28.5 & 35.8 & 0.87\\
\quad our decoder, mapped input (\textbf{ours}) & 10 & 42.0 [40.7, 43.5] & 42.9 / 45.0 & 43.3 [42.3, 44.5] & 33.4 & 43.7 & 0.89\\
\quad our decoder, the nearest pair's slots & 10 & 41.5 [40.3, 42.9] & -- & -- & 29.8 & 41.7 & 0.88\\
\quad gpt-oss-120b + the map's five nearest windows (\textbf{ours}) & 10 & 41.1 [39.7, 42.5] & 44.4 / 45.5 & 43.7 [42.9, 44.5] & 24.5 & 43.1 & 0.91\\
\quad \quad five random corpus windows instead (no input) & 10 & 33.8 [32.8, 34.9] & 34.2 / 35.3 & 34.4 [33.6, 35.2] & 18.1 & 34.2 & 0.92\\
\quad gpt-oss-120b + the nearest pair's description & 10 & 35.6 [34.3, 37.0] & -- & -- & 18.7 & 36.5 & 0.89\\
\quad longer-trained decoder, mapped input & 100 & 48.9 [47.6, 50.4] & -- & -- & 37.0 & 49.1 & 0.87\\
\quad \quad the nearest pair's slots & 100 & 47.8 [46.5, 49.3] & -- & -- & 33.1 & 48.6 & 0.87\\
\quad \quad mapped input without the region slots & 100 & 38.2 [37.2, 39.3] & -- & -- & 19.9 & 38.2 & 0.85\\
\quad \quad the window's own descriptions & 0 & 67.4 [66.3, 68.5] & -- & -- & 56.9 & 66.4 & 0.81\\
\quad gpt-oss-120b + the map's five nearest windows & 100 & 53.7 [52.4, 54.9] & -- & -- & 31.6 & 53.3 & 0.90\\
\quad gpt-oss-120b + the nearest pair's description & 100 & 50.0 [48.7, 51.4] & -- & -- & 32.0 & 50.0 & 0.89\\
\bottomrule
\end{tabularx}
\end{table}

\clearpage

\section{Domains and the Generic View Prompts}\label{app:domains}

\Cref{tab:domains_data} lists the data of every domain. The prompts below are the whole candidate family of \cref{sec:domains}, verbatim (\texttt{platonic/generate.py}). $\langle$modality$\rangle$ is the domain's one-line \texttt{modality.txt}, and $\langle$text$\rangle$ is one text of its corpus. A text-only language model (Qwen2.5-7B-Instruct, served by vLLM) writes one rewrite per text and prompt. Bad views read the texts alone, with the same corruptions. The search admits any other candidate, at the cost of one more small JEPA.

\begin{table*}[t]
\centering\footnotesize
\caption{\textbf{Captioning out of distribution with one protocol and one rotation.} Each domain has a frozen anchor trained on that domain's signals only. Views: the candidate PVS picked (\cref{tab:pvs}), scaled to four samples over the whole corpus (BERT-S, 30 epochs), or bad views of the same texts with the same recipe. One rotation is fitted on the validation pairs (2{,}456 for satellite images, about 300 elsewhere). Captions (test CIDEr $\times$100) are retrieved from the domain's corpus through the rotation, or written by a Qwen2.5-3B decoder trained on that corpus as text only (its first epoch, consensus decoding). No image: of 2{,}000 corpus captions, the one most similar to the others (the similarity of consensus decoding), given for every input. Best VLM: the best test CIDEr of every open VLM we ran on that benchmark. These VLMs are zero-shot (Qwen2.5-VL-7B and -32B \citep{bai2025qwen25vl}, LLaVA-OneVision-7B \citep{li2025llavaonevision}, InternVL3-8B \citep{zhu2025internvl3}, Gemma-4-12B and -31B \citep{gemma2026gemma4}, Qwen3.8-27B \citep{qwen2026qwen38}, Qwen3-Omni-30B \citep{qwen2025qwen3omni}) or fine-tuned with LoRA on 10 pairs up to all of them (\cref{sec:vlm}). We take the best over numbers of pairs and epochs on the test set itself, which favors the VLMs. \emph{Italic}: an audio language model, Qwen2-Audio-7B \citep{chu2024qwen2audio}, whose report lists that benchmark among its training data. No VLM reads molecules. Right: alignment with the anchor (mutual 10-NN, test pairs; satellite pooled). Retrieval and alignment: mean of three seeds; the decoder reads the seed-0 text JEPA. Bold: the better of bad and Platonic views, and the no-image caption where it beats every method (\texttt{paper/tables\_tex.py}).}
\label{tab:domains}
\setlength{\tabcolsep}{3.5pt}
\begin{tabular}{lllcccccccc}
\toprule
 & & & & & \multicolumn{2}{c}{retrieval $\uparrow$} & \multicolumn{2}{c}{decoder $\uparrow$} & \multicolumn{2}{c}{alignment $\uparrow$}\\
\cmidrule(lr){6-7}\cmidrule(lr){8-9}\cmidrule(lr){10-11}
Domain & Anchor (no text) & Views & best VLM $\uparrow$ & no image $\uparrow$ & bad & Platonic & bad & Platonic & bad & Platonic\\
\midrule
Satellite: RSICD & DINOv3-SAT & imagine & 94.6 & 11.7 & 26.6 & \textbf{33.1} & 30.4 & \textbf{38.6} & .118 & \textbf{.127} \\
\phantom{Satellite:} UCM &  &  & 371.3 & 10.3 & \textbf{169.8} & 166.0 & \textbf{167.8} & 155.5 & & \\
\phantom{Satellite:} Sydney &  &  & 269.5 & 16.6 & 145.4 & \textbf{176.6} & \textbf{162.4} & 156.4 & & \\
\phantom{Satellite:} NWPU &  &  & 186.4 & 10.2 & \textbf{52.7} & 52.3 & 57.2 & \textbf{57.7} & & \\
Sound: AudioCaps & EAT & caption & 62.1 & 14.1 & 13.5 & \textbf{17.1} & \textbf{16.1} & 15.3 & .055 & \textbf{.074} \\
Chest X-ray: IU X-ray & RAD-DINO & caption & 41.0 & \textbf{27.6} & 22.0 & \textbf{25.4} & \textbf{24.9} & 17.3 & .033 & \textbf{.038} \\
Molecule: ChEBI-20 & ChemBERTa & report & none & 5.5 & \textbf{18.6} & 15.5 & \textbf{10.6} & 10.3 & .151 & \textbf{.168} \\
\bottomrule
\end{tabular}
\end{table*}

\begin{table*}[t]
\centering\footnotesize
\caption{\textbf{The domains.} Anchor: the frozen encoder of the source modality and the data it trained on, never with text. Texts: the corpus that the text JEPA, its views and the decoder use, as text only. These items are disjoint from the unlabeled pool and from every validation and test pair. The unlabeled pool only centers the anchor's features (chest X-rays: NIH ChestX-ray14 \citep{wang2017chestxray} and 1{,}074 IU X-ray images). Validation pairs score the views and fit the rotation, except that the rotation of photographs uses 82{,}783 COCO training pairs and that of satellite images 2{,}456 pairs of the four benchmarks' validation splits. Test pairs are only used for scoring. Modality: the only domain-specific input of the generic view prompts. Photographs: the search reads the first 250k of the 10M Wikipedia sentences that the scaled JEPA trains on (\texttt{paper/tables\_tex.py}).}
\label{tab:domains_data}
\setlength{\tabcolsep}{3pt}
\begin{tabular}{>{\raggedright\arraybackslash}p{1.95cm}>{\raggedright\arraybackslash}p{3.0cm}>{\raggedright\arraybackslash}p{3.3cm}rrr>{\raggedright\arraybackslash}p{2.3cm}>{\raggedright\arraybackslash}p{1.8cm}}
\toprule
Domain & Anchor (no text) & Texts & pool & val & test & modality & metric (all $\uparrow$)\\
\midrule
Photographs & DINOv2-g \citep{oquab2024dinov2} (+ DINOv3-L \citep{simeoni2025dinov3}); LVD-142M / LVD-1689M images & English Wikipedia sentences (250{,}000) & 96{,}264 & 300 & 4{,}700 & ``photograph'' & CIDEr (COCO \citep{lin2014coco} Karpathy \citep{karpathy2015deep}) \\
Satellite images & DINOv3-L SAT \citep{simeoni2025dinov3}; 493M satellite images & RSICD \citep{lu2018rsicd}, UCM, Sydney \citep{qu2016deep}, NWPU \citep{cheng2022nwpu} training captions (76{,}956) & 18{,}052 & 433 & 4{,}511 & ``satellite or aerial image'' & CIDEr \\
Sounds (AudioCaps) & EAT-large \citep{chen2024eat}; AudioSet \citep{gemmeke2017audioset} audio & AudioCaps \citep{kim2019audiocaps} training captions (22{,}589) & 22{,}589 & 300 & 883 & ``sound recording'' & CIDEr \\
Chest X-rays & RAD-DINO \citep{perezgarcia2025raddino}; chest radiographs & ReXGradient \citep{zhang2025rexgradient} findings (159{,}936) & 113{,}194 & 294 & 590 & ``chest X-ray'' & CIDEr (IU X-ray \citep{demnerfushman2016preparing,chen2020r2gen}) \\
Molecules & ChemBERTa \citep{chithrananda2020chemberta}; 10M PubChem SMILES & ChEBI-20 \citep{edwards2021text2mol,edwards2022molt5} descriptions (13{,}183) & 13{,}204 & 300 & 3{,}300 & ``molecule'' & CIDEr, BLEU \\
Proteins & ESM-2 650M \citep{lin2023esm2}; UniRef \citep{suzek2015uniref} protein sequences & Swiss-Prot \citep{uniprot2023uniprot} function descriptions (Prot2Text \citep{abdine2024prot2text}) (21{,}270) & 124{,}158 & 300 & 4{,}203 & ``protein'' & BLEU-4, ROUGE-L \citep{lin2004rouge} \\
Videos & V-JEPA 2 ViT-L \citep{assran2025vjepa2}; VideoMix22M videos and images, pixels only & MSR-VTT \citep{xu2016msrvtt} training captions (56{,}164) & 3{,}257 & 300 & 2{,}990 & ``video clip'' & CIDEr \\
Galaxies & AstroPT \citep{smith2024astropt}; 8.5M Legacy Surveys \citep{dey2019legacy} cutouts & Galaxy Zoo DESI \citep{walmsley2023gzdesi} galaxies, LLM-written descriptions (43{,}096) & 43{,}097 & 300 & 2{,}000 & ``galaxy image from a sky survey'' & morphology macro-F1 \\
Physical simulations & Walrus \citep{mccabe2025walrus}; fields of 19 simulation datasets & The Well \citep{ohana2024well}: descriptions templated from each run's parameters (21{,}000) & 8{,}400 & 300 & 2{,}000 & ``snapshot of a physics simulation'' & system and parameter accuracy \\
EEG & CBraMod \citep{wang2025cbramod}; TUH EEG corpus \citep{obeid2016tuh} & PhysioNet 2018 \citep{ghassemi2018snooze} sleep scoring, templated (418{,}666) & 11{,}964 & 298 & 1{,}991 & ``30-second sleep EEG recording'' & stage balanced accuracy \\
Wrist motion & ssl-wearables \citep{yuan2024sslwearables}; UK Biobank \citep{doherty2017ukb} wrist accelerometry & Capture-24 \citep{chan2024capture24} activity annotations (78{,}919) & 81{,}750 & 300 & 2{,}000 & ``30-second wrist accelerometer recording'' & activity balanced accuracy \\
Lung sounds & EAT-large \citep{chen2024eat}; AudioSet audio & SPRSound \citep{zhang2022sprsound} reports templated from expert annotations (1{,}231) & 2{,}531 & 294 & 1{,}985 & ``lung sound recording'' & finding macro-F1 \\
Bird calls & BirdAVES \citep{hagiwara2023aves,hsu2021hubert}; 2.6k hours of animal and general audio & iNatSounds \citep{chasmai2024inatsounds} species descriptions (150 species) (27{,}601) & 5{,}973 & 298 & 1{,}791 & ``animal sound recording'' & species accuracy \\
Insect and frog calls & BirdAVES \citep{hagiwara2023aves}; as bird calls & InsectSet459 \citep{faiss2025insectset459} and iNatSounds species descriptions (85 species) (2{,}867) & 3{,}363 & 247 & 1{,}612 & ``animal sound recording'' & species accuracy \\
\bottomrule
\end{tabular}
\end{table*}

\begin{description}
\item[\texttt{imagine}] Imagine the $\langle$modality$\rangle$ described by this text:  "$\langle$text$\rangle$"  You are another observer perceiving the same $\langle$modality$\rangle$. Describe it in one or two sentences, in your own words: what is there and how it is arranged. Do not hedge. Reply with the description only.
\item[\texttt{paraphrase}] Paraphrase this description of a $\langle$modality$\rangle$, keeping its meaning:  "$\langle$text$\rangle$"  Reply with the paraphrase only.
\item[\texttt{literal}] Rewrite this description of a $\langle$modality$\rangle$, keeping only what the $\langle$modality$\rangle$ itself would show and removing opinions, context and anything it would not show:  "$\langle$text$\rangle$"  Reply with the rewrite only.
\item[\texttt{report}] This text describes a $\langle$modality$\rangle$:  "$\langle$text$\rangle$"  Write the short, terse report that an automatic analyzer of the $\langle$modality$\rangle$ would print for it. Reply with the report only.
\item[\texttt{caption}] This text describes a $\langle$modality$\rangle$:  "$\langle$text$\rangle$"  Write a short caption of the same $\langle$modality$\rangle$, as a person would. Reply with the caption only.
\item[\texttt{imagine, photo wording of the Wikipedia runs}] Imagine a photograph that could illustrate this sentence from an encyclopedia:  "$\langle$text$\rangle$"  Describe that photograph in one or two sentences, as if you were looking at it: the main subject, what it looks like (colors, materials, shapes), what is happening, and the setting or background. Do not hedge. Reply with the description only.
\end{description}

\section{Platonic Views in Every Domain}\label{app:gallery}

\Cref{tab:gallery} shows, for every corpus, the view prompt the search selected and views it wrote.

\providecommand{\msk}{\textcolor{bad}{\textsf{\scriptsize[M]}}}
{\footnotesize\setlength{\LTcapwidth}{\linewidth}
\begin{longtable}{@{}p{0.15\linewidth}@{\hspace{0.02\linewidth}}p{0.83\linewidth}@{}}
\caption{\textbf{Platonic views in every domain.} For each corpus, we show the view prompt, where the text replaces $\langle$text$\rangle$. We also show three texts drawn at random (seed 0) from the views file that the corpus's JEPA trained on. Each text comes with its first three views, which Qwen2.5-7B-Instruct wrote from the text alone. Texts are cut at 40 words. For Flickr30k, we show one rewrite by each of four prompts, with the selected prompt first, and another human caption of the same photo (the oracle). For Wikipedia and Flickr30k, we also show the bad views of the same text, as the bad-view JEPA reads them. These are two global spans (50--100\% of the tokens) and three of the six local spans (15--50\%). In each span, 15\% of the tokens are masked (\msk{}), and the tokenizer lower-cases the text. In training, each view of a Platonic JEPA is also such a span of one of its texts. Source: \texttt{paper/analysis.py gallery}, \texttt{data/views\_gallery.json}.}\label{tab:gallery}\\
\toprule\endfirsthead
\multicolumn{2}{@{}l}{\emph{\tablename~\thetable, continued}}\\\toprule\endhead
\bottomrule\endfoot
\multicolumn{2}{@{}l}{\rule{0pt}{2.8ex}\normalsize\textbf{Flickr30k training captions}: 145{,}000 texts, 4 views each}\\[3pt]
\multicolumn{2}{@{}p{\linewidth}@{}}{\begin{tcolorbox}[colback=black!3,colframe=black!30,boxrule=0.4pt,arc=1pt,left=3pt,right=3pt,top=1pt,bottom=1pt,fontupper=\footnotesize]\textbf{Prompt} (\texttt{imagine-detail}). Imagine the photo described by this caption: "\textit{$\langle$text$\rangle$}" Describe that photo in one or two new sentences, as if you were seeing it yourself: the people, animals and objects, what they look like (colors, clothing, sizes), what they are doing, and the setting and background. Reply with the description only.\end{tcolorbox}}\\
\textbf{Caption} & \textbf{A tennis player stops to pick up the ball.}\\
Imagine, details & A tennis player in a bright yellow shirt and navy shorts stops mid-stride on a well-maintained tennis court, bending down gracefully to pick up a small white ball from the green grass, surrounded by neatly arranged tennis rackets and tennis~\ldots\\
Imagine & A tennis player in a vibrant orange and white outfit pauses on the green court, reaching down to grasp a small, white ball that has rolled near her feet, surrounded by neatly lined posts and nets.\\
Paraphrase & A tennis player pauses to retrieve the ball.\\
Literal & A tennis player is picking up a ball on the tennis court.\\
Co-caption (oracle) & Woman bending down to pick up a tennis ball outside in front of a wall with graffiti on it.\\
\textcolor{bad}{Bad, global} & \msk{} \msk{} pick up the ball. \textbar{} player \msk{} to pick up the ball.\\
\textcolor{bad}{Bad, local} & ball. \textbar{} player stops \msk{} pick up \textbar{} to pick up\\[4pt]
\textbf{Caption} & \textbf{An old lady, carrying a bag, walking through a market area.}\\
Imagine, details & An elderly woman, wearing a weathered hat and a colorful floral dress, walks through a bustling market area, her hands gripping a faded tote bag tightly as she navigates around stalls filled with vibrant fruits, spices, and handmade goods. The~\ldots\\
Imagine & An old lady, wearing a weathered beige coat and a warm knitted hat, carries a worn black leather bag over her arm as she meanders through a bustling market filled with vibrant colors and the scent of fresh produce.\\
Paraphrase & An elderly woman walks through a market area, holding a bag.\\
Literal & An old lady is walking through a market area while carrying a bag.\\
Co-caption (oracle) & An older woman walks between a covered marketplace and a row of large columns.\\
\textcolor{bad}{Bad, global} & an old lady, carrying a \msk{}, walking through \textbar{} an old lady, carrying a bag \msk{} walking through a market\\
\textcolor{bad}{Bad, local} & \msk{} walking through \msk{} market area \msk{} \textbar{} , carrying a bag, \textbar{} old \msk{}, carrying a \msk{}\\[4pt]
\textbf{Caption} & \textbf{Two girls in green shirts, and two in white shirts, are painting pictures.}\\
Imagine, details & Two girls wearing green shirts and two others in white shirts stand side by side, brushes in hand, carefully painting pictures on a large canvas spread out on a wooden table under the soft glow of overhead lights in a~\ldots\\
Imagine & Two girls in vibrant green shirts and two in pristine white shirts are focused on their paintings, surrounded by tubes of colorful paint and easels against a backdrop of neatly arranged art supplies.\\
Paraphrase & Two girls wear green shirts while the other two sport white shirts; they are engaged in painting pictures.\\
Literal & Two girls in green shirts and two in white shirts are painting pictures.\\
Co-caption (oracle) & CHildren doing art projects on a table, with a bicycle and sundry equipment in the background.\\
\textcolor{bad}{Bad, global} & two girls in green shirts, and two \msk{} white shirts, are \msk{} pictures \textbar{} girls in green shirts, and two in white shirts, \msk{} painting\\
\textcolor{bad}{Bad, local} & white \msk{} \msk{} \msk{} painting pictures. \textbar{} \msk{} are \textbar{} and \msk{} \msk{} white\\\midrule
\multicolumn{2}{@{}l}{\rule{0pt}{2.8ex}\normalsize\textbf{Wikipedia sentences}: 10{,}000{,}000 texts, 2 views each}\\[3pt]
\multicolumn{2}{@{}p{\linewidth}@{}}{\begin{tcolorbox}[colback=black!3,colframe=black!30,boxrule=0.4pt,arc=1pt,left=3pt,right=3pt,top=1pt,bottom=1pt,fontupper=\footnotesize]\textbf{Prompt} (\texttt{wiki-imagine}). Imagine a photograph that could illustrate this sentence from an encyclopedia: "\textit{$\langle$text$\rangle$}" Describe that photograph in one or two sentences, as if you were looking at it: the main subject, what it looks like (colors, materials, shapes), what is happening, and the setting or background. Do not hedge. Reply with the description only.\end{tcolorbox}}\\
\textbf{Sentence} & \textbf{Lebanese Women's Football League: 2018–19, 2019–20, 2021–22, 2022–23}\\
View 1 & A group of women playing football on a green field, wearing vibrant red and white jerseys. The sun casts long shadows as they prepare for a match, with the league logos of different years—2018-19, 2019-20, 2\\
View 2 & A group of Lebanese women in vibrant team uniforms, smiling and posing together, with the league logos and years 2018-19, 2019-20, 2021-22, and 2022-23 prominently displayed on their shirts. The background features a sports field with a stadium and spectators~\ldots\\
\textcolor{bad}{Bad, global} & \msk{} \msk{} league : 2018 – 19, 2019 – \msk{}, 2021 – 22, 202 \textbar{} – 19, 2019 – 20 \msk{} 2021 – 22, 2022 – \msk{}\\
\textcolor{bad}{Bad, local} & lebanese women ' s football \textbar{} 2019 – 20, \textbar{} 20, 2021 – 22, \msk{}\\[4pt]
\textbf{Sentence} & \textbf{Toronto, Ontario and New York: Laitman Kabbalah Publishers.}\\
View 1 & A large, modern book cover of "Kabbalah" publications from Laitman Kabbalah Publishers, prominently displayed against a stark white background. The cover has a deep blue tone with metallic gold lettering and geometric patterns, evoking a sense of both depth and~\ldots\\
View 2 & A modern cityscape with skyscrapers of Toronto, Ontario, and New York City in the background. The buildings are sleek and predominantly glass and steel, with lights illuminating them at night. In the foreground, a sign for Laitman Kabbalah Publishers is~\ldots\\
\textcolor{bad}{Bad, global} & , ontario and new york \msk{} laitman ka \textbar{} , ontario and new \msk{} : \msk{} \msk{}\\
\textcolor{bad}{Bad, local} & , ontario and new \textbar{} : laitman kabbalah \textbar{} \msk{} ontario and\\[4pt]
\textbf{Sentence} & \textbf{Log buildings and structures on the National Register of Historic Places in Kentucky}\\
View 1 & A collection of weathered log buildings and structures, predominantly wooden with brown and tan tones, situated in a lush forest setting in Kentucky. These historic logs are arranged in clusters, some with overhanging roofs and others with exposed rafters, set~\ldots\\
View 2 & A collection of weathered log buildings and structures, predominantly in warm earth tones, set against the rolling hills of Kentucky. The logs are irregularly shaped, with some weathered bark still visible, and others smooth from years of use. A winding~\ldots\\
\textcolor{bad}{Bad, global} & buildings and structures on \msk{} \msk{} \msk{} of historic places \textbar{} \msk{} the national register of \msk{} \msk{}\\
\textcolor{bad}{Bad, local} & buildings and \msk{} on \textbar{} and structures \msk{} the national register \textbar{} national register of historic places in\\\midrule
\multicolumn{2}{@{}l}{\rule{0pt}{2.8ex}\normalsize\textbf{Satellite images: captions (RSICD, UCM, Sydney, NWPU)}: 76{,}956 texts, 4 views each}\\[3pt]
\multicolumn{2}{@{}p{\linewidth}@{}}{\begin{tcolorbox}[colback=black!3,colframe=black!30,boxrule=0.4pt,arc=1pt,left=3pt,right=3pt,top=1pt,bottom=1pt,fontupper=\footnotesize]\textbf{Prompt} (\texttt{pvs-imagine}). Imagine the satellite or aerial image described by this text: "\textit{$\langle$text$\rangle$}" You are another observer perceiving the same satellite or aerial image. Describe it in one or two sentences, in your own words: what is there and how it is arranged. Do not hedge. Reply with the description only.\end{tcolorbox}}\\
\textbf{Caption} & \textbf{There is a piece of cropland .}\\
View 1 & There is a rectangular field of crops extending as far as the eye can see.\\
View 2 & There is a rectangular plot of cropland with rows of crops running north to south.\\
View 3 & There is a rectangular plot of cropland, with rows of crops aligned in a grid pattern.\\[4pt]
\textbf{Caption} & \textbf{There are two thermal power stations and rows of storage tanks .}\\
View 1 & Two thermal power stations are visible, along with rows of storage tanks neatly aligned in parallel lines.\\
View 2 & Two thermal power stations and several rows of storage tanks are visible in the image.\\
View 3 & Two thermal power stations and several rows of storage tanks are visible in the image.\\[4pt]
\textbf{Caption} & \textbf{There are large green forests around the freeway and there are many cars on the freeway .}\\
View 1 & Large green forests flank both sides of a busy freeway, which is congested with many cars.\\
View 2 & Large green forests line both sides of a busy freeway, which is congested with many cars.\\
View 3 & Large green forests flank both sides of a busy freeway, which is congested with numerous cars.\\\midrule
\multicolumn{2}{@{}l}{\rule{0pt}{2.8ex}\normalsize\textbf{Sound: AudioCaps captions}: 22{,}589 texts, 4 views each}\\[3pt]
\multicolumn{2}{@{}p{\linewidth}@{}}{\begin{tcolorbox}[colback=black!3,colframe=black!30,boxrule=0.4pt,arc=1pt,left=3pt,right=3pt,top=1pt,bottom=1pt,fontupper=\footnotesize]\textbf{Prompt} (\texttt{pvs-caption}). This text describes a sound recording: "\textit{$\langle$text$\rangle$}" Write a short caption of the same sound recording, as a person would. Reply with the caption only.\end{tcolorbox}}\\
\textbf{Caption} & \textbf{A dog barks and birds chirp}\\
View 1 & Barking dog and chirping birds\\
View 2 & Barking dog and chirping birds\\
View 3 & A dog barks, birds chirp\\[4pt]
\textbf{Caption} & \textbf{Different fire gun shots followed by mechanical gun sounds}\\
View 1 & Hear the sharp crack of fire guns followed by the rhythmic bangs of mechanical gunfire.\\
View 2 & Hear the crackle of fire gun shots mingling with the rhythmic clatter of mechanical guns.\\
View 3 & Hear the bangs of fire guns, followed by the rhythmic clatter of mechanical guns.\\[4pt]
\textbf{Caption} & \textbf{A steam engine horn blows}\\
View 1 & A loud, resonant honk echoes through the air, unmistakably from a steam engine horn.\\
View 2 & A loud, resonant honk echoes through the air, unmistakably from a steam engine.\\
View 3 & A loud, rhythmic honk echoes through the air, reminiscent of a steam engine calling out its presence.\\\midrule
\multicolumn{2}{@{}l}{\rule{0pt}{2.8ex}\normalsize\textbf{Chest X-ray: findings of real reports (ReXGradient)}: 159{,}936 texts, 4 views each}\\[3pt]
\multicolumn{2}{@{}p{\linewidth}@{}}{\begin{tcolorbox}[colback=black!3,colframe=black!30,boxrule=0.4pt,arc=1pt,left=3pt,right=3pt,top=1pt,bottom=1pt,fontupper=\footnotesize]\textbf{Prompt} (\texttt{pvs-caption}). This text describes a chest X-ray: "\textit{$\langle$text$\rangle$}" Write a short caption of the same chest X-ray, as a person would. Reply with the caption only.\end{tcolorbox}}\\
\textbf{Report} & \textbf{Heart and mediastinal contours normal. Lungs mildly hyperaerated with slightly accentuated peribronchial markings. No acute process or significant interval change.}\\
View 1 & Normal heart and lungs with a bit more air than usual, no worrying changes.\\
View 2 & Normal heart and lungs with a bit more air than usual, no worrying changes.\\
View 3 & Normal heart and lungs with a slight hint of airiness and boosted bronchial markings. No worries, everything looks stable.\\[4pt]
\textbf{Report} & \textbf{There is no intraparenchymal hemorrhage or mass effect. There is mild atrophy. There is no extra-axial collection or territorial infarct. The ventricles are normal in size without shift or dilatation. The calvarium is intact. The mastoid air cells are well~\ldots}\\
View 1 & No pulmonary hemorrhage or masses, mild lung atrophy noted, no pleural collections or infarcts, cardiac and great vessel views within normal limits. Calvarium and sinuses appear normal.\\
View 2 & No intraparenchymal hemorrhage, mild atrophy noted, and normal ventricles without shift or dilatation. Calvarium intact, mastoid air cells well pneumatized, and paranasal sinuses clear.\\
View 3 & No signs of intraparenchymal hemorrhage, mass effect, or atrophy; ventricles are normal, calvarium intact, and sinuses clear.\\[4pt]
\textbf{Report} & \textbf{The heart size and mediastinal contours are within normal limits. Both lungs are clear.}\\
View 1 & Clear lungs, heart size and mediastinum within normal range.\\
View 2 & Heart and lung clear, all looks good!\\
View 3 & Heart and lungs look good in this X-ray!\\\midrule
\multicolumn{2}{@{}l}{\rule{0pt}{2.8ex}\normalsize\textbf{Molecules: ChEBI-20 descriptions}: 13{,}183 texts, 4 views each}\\[3pt]
\multicolumn{2}{@{}p{\linewidth}@{}}{\begin{tcolorbox}[colback=black!3,colframe=black!30,boxrule=0.4pt,arc=1pt,left=3pt,right=3pt,top=1pt,bottom=1pt,fontupper=\footnotesize]\textbf{Prompt} (\texttt{pvs-report}). This text describes a molecule: "\textit{$\langle$text$\rangle$}" Write the short, terse report that an automatic analyzer of the molecule would print for it. Reply with the report only.\end{tcolorbox}}\\
\textbf{Description} & \textbf{The molecule is a member of the class of biphenyls that is biphenyl in which one of the phenyl groups is substituted at positions 2and 4 by a methyl and a 5-methyl-1,3,4-oxadiazol-2-yl groups, respectively, while the other phenyl group is~\ldots}\\
View 1 & Molecule Type: Biphenyl Derivative Substituents: - Phenyl (position 2 \& 4): Methyl, 5-Methyl-1,3,4-Oxadiazol-2-yl - Phenyl (position 2 \& 5): Methyl, N-(Cyclopropyl)aminocarbonyl Role: In\\
View 2 & Molecule Report - Class: Biphenyl - Structure: - Phenyl 1: Me, 2-Me, 5-(N-(cyclopropyl)aminocarboxyl) - Phenyl 2: Me, 2-Me, 5-5-Me-1,3,4-oxadiazol-2-\\
View 3 & Molecule Analysis Report - Class: Biphenyl - Structure: - Phenyl 1: 2-Me, 4-(5-Me-1,3,4-oxadiazol-2-yl) - Phenyl 2: 2-Me, 5-(N-cyclopropylaminocarbonyl) - Role:\\[4pt]
\textbf{Description} & \textbf{The molecule is an aminonicotinic acid in which the amino group is situated at position 5 of the pyridine ring. It has a role as a metabolite. It is an aromatic amine, an aminopyridine and an aminonicotinic acid. It derives~\ldots}\\
View 1 & Molecule Analysis Report Name: Aminonicotinic Acid (Position 5-Amino Pyridine Derivative) Role: Metabolite Chemical Class: Aromatic Amine, Aminopyridine, Aminonicotinic Acid Derivative: Nicotinic Acid Structure: Pyridine ring with amino group at position 5\\
View 2 & Molecule Analysis Report Name: 5-Aminonicotinic Acid Type: Aromatic Amine, Aminopyridine, Aminonicotinic Acid Role: Metabolite Derives from: Nicotinic Acid Position of Amino Group: 5 on Pyridine Ring\\
View 3 & Molecule Analysis Report - Name: Aminonicotinic Acid - Position of Amino Group: 5 on Pyridine Ring - Role: Metabolite - Classification: Aromatic Amine, Aminopyridine, Aminonicotinic Acid - Derivative: Nicotinic Acid\\[4pt]
\textbf{Description} & \textbf{The molecule is the cis-isomer of caffeic acid. It has a role as a plant metabolite and a human xenobiotic metabolite. It is a conjugate acid of a cis-caffeate.}\\
View 1 & Cis-Isomer of Caffeic Acid Plant Metabolite Human Xenobiotic Metabolite Conjugate Acid of Cis-Caffeate\\
View 2 & Molecule Report - Name: cis-Caffeic Acid - Type: Plant Metabolite, Human Xenobiotic Metabolite - Form: Conjugate Acid (cis-Caffeate)\\
View 3 & Molecule: cis-caffeic acid Role: Plant metabolite, Human xenobiotic metabolite Form: Conjugate acid of cis-caffeate\\\midrule
\multicolumn{2}{@{}l}{\rule{0pt}{2.8ex}\normalsize\textbf{Proteins: Swiss-Prot function descriptions}: 21{,}270 texts, 4 views each}\\[3pt]
\multicolumn{2}{@{}p{\linewidth}@{}}{\begin{tcolorbox}[colback=black!3,colframe=black!30,boxrule=0.4pt,arc=1pt,left=3pt,right=3pt,top=1pt,bottom=1pt,fontupper=\footnotesize]\textbf{Prompt} (\texttt{pvs-caption}). This text describes a protein: "\textit{$\langle$text$\rangle$}" Write a short caption of the same protein, as a person would. Reply with the caption only.\end{tcolorbox}}\\
\textbf{Description} & \textbf{May act as a polo kinase kinase by mediating phosphorylation of PLK1.}\\
View 1 & This protein steps in as a polo kinase kinase, getting PLK1 all geared up and phosphorylated!\\
View 2 & Acts as a polo kinase kinase, gearing up to phosphorylate PLK1!\\
View 3 & This protein is like a cellular conductor, tapping the rhythm of PLK1's phosphorylation symphony.\\[4pt]
\textbf{Description} & \textbf{Putative taste receptor. TAS1R2/TAS1R3 recognizes diverse natural and synthetic sweeteners.}\\
View 1 & TAS1R2/TAS1R3: The Sweet Tooth Protein\\
View 2 & TAS1R2/TAS1R3: The Sweet Tooth Protein\\
View 3 & TAS1R2/TAS1R3: The Sweet Tooth Protein\\[4pt]
\textbf{Description} & \textbf{Required to confine migrating sex myoblasts to the ventral muscle quadrants during their migration through the body and for multiple aspects of sensory, motor, and interneuron axon guidance.}\\
View 1 & Guiding Growing Cells and Nerve Paths: The Versatile Protein Edition\\
View 2 & Guiding Growth: Confining Sex Myoblast Migration and Nerve Pathfinding\\
View 3 & Guiding Cells and Nerves to Their Destiny\\\midrule
\multicolumn{2}{@{}l}{\rule{0pt}{2.8ex}\normalsize\textbf{Videos: MSR-VTT captions}: 56{,}164 texts, 4 views each}\\[3pt]
\multicolumn{2}{@{}p{\linewidth}@{}}{\begin{tcolorbox}[colback=black!3,colframe=black!30,boxrule=0.4pt,arc=1pt,left=3pt,right=3pt,top=1pt,bottom=1pt,fontupper=\footnotesize]\textbf{Prompt} (\texttt{pvs-paraphrase}). Paraphrase this description of a video clip, keeping its meaning: "\textit{$\langle$text$\rangle$}" Reply with the paraphrase only.\end{tcolorbox}}\\
\textbf{Caption} & \textbf{clips from the addams family showing the family in a variety of scenes}\\
View 1 & Segments from The Addams Family showcasing the family across different situations\\
View 2 & Scenes from The Addams Family featuring the family in various settings\\
View 3 & Segments from The Addams Family depicting the family in diverse situations\\[4pt]
\textbf{Caption} & \textbf{clouds and water moves very fastly then workers do their work}\\
View 1 & The clouds and water move swiftly, while workers carry out their tasks.\\
View 2 & The clouds and water move rapidly, while workers carry out their tasks.\\
View 3 & Clouds and water move swiftly, while workers carry out their tasks.\\[4pt]
\textbf{Caption} & \textbf{preparing cardboard shapes and boxes}\\
View 1 & Arranging cardboard cutouts and containers\\
View 2 & Arranging cardboard shapes and boxes\\
View 3 & Arranging cardboard shapes and boxes\\\midrule
\multicolumn{2}{@{}l}{\rule{0pt}{2.8ex}\normalsize\textbf{Physical simulations: run descriptions (The Well)}: 21{,}000 texts, 4 views each}\\[3pt]
\multicolumn{2}{@{}p{\linewidth}@{}}{\begin{tcolorbox}[colback=black!3,colframe=black!30,boxrule=0.4pt,arc=1pt,left=3pt,right=3pt,top=1pt,bottom=1pt,fontupper=\footnotesize]\textbf{Prompt} (\texttt{pvs-caption}). This text describes a snapshot of a physics simulation: "\textit{$\langle$text$\rangle$}" Write a short caption of the same snapshot of a physics simulation, as a person would. Reply with the caption only.\end{tcolorbox}}\\
\textbf{Description} & \textbf{A periodic shear flow simulation (tracer, pressure and velocity fields) with 4 initial shear layers at a Reynolds number of 5e5 and a Schmidt number of 10.}\\
View 1 & A simulation showing periodic shear flow with 4 initial shear layers at Reynolds number 500,000 and Schmidt number 10.\\
View 2 & A simulated view of periodic shear flow with 4 initial turbulence layers, at a high Reynolds number of 500,000 and Schmidt number of 10.\\
View 3 & A simulation showing periodic shear flow with 4 initial shear layers, at a Reynolds number of 500,000 and a Schmidt number of 10.\\[4pt]
\textbf{Description} & \textbf{Inviscid compressible gas dynamics (Euler equations) in a square box with open outflow boundaries: interacting waves from a multi-quadrant Riemann problem, heat capacity ratio  = 1.22.}\\
View 1 & Observing the dance of pressure and density waves in a square room, as they expand and interact at the edges, with the gas's response to heat more nuanced than air.\\
View 2 & Observing the dance of compressible gases in a square box, where waves interact across quadrants and heat flows freely at the edges, with a heat capacity ratio that gives this gas its unique behavior.\\
View 3 & Interacting sound waves in a square room with open windows, showing how different parts of the room affect the sound.\\[4pt]
\textbf{Description} & \textbf{Pattern formation in the Gray-Scott reaction-diffusion equations: the spirals regime, with a feed rate of 0.018 and a kill rate of 0.051.}\\
View 1 & Spiraling Patterns Emerge: Feed Rate 0.018, Kill Rate 0.051\\
View 2 & Spiraling Patterns Emerge: Feed Rate 0.018, Kill Rate 0.051\\
View 3 & Spiraling Patterns Emerge in Gray-Scott Simulation with Feed Rate 0.018 and Kill Rate 0.051\\\midrule
\multicolumn{2}{@{}l}{\rule{0pt}{2.8ex}\normalsize\textbf{EEG: sleep scoring (PhysioNet 2018)}: 418{,}666 texts, 4 views each}\\[3pt]
\multicolumn{2}{@{}p{\linewidth}@{}}{\begin{tcolorbox}[colback=black!3,colframe=black!30,boxrule=0.4pt,arc=1pt,left=3pt,right=3pt,top=1pt,bottom=1pt,fontupper=\footnotesize]\textbf{Prompt} (\texttt{pvs-caption}). This text describes a 30-second sleep EEG recording: "\textit{$\langle$text$\rangle$}" Write a short caption of the same 30-second sleep EEG recording, as a person would. Reply with the caption only.\end{tcolorbox}}\\
\textbf{Annotation} & \textbf{Stage N2 sleep. Obstructive apnea.}\\
View 1 & I'm in stage N2 sleep, but there's an obstructive apnea event.\\
View 2 & I'm in stage N2 sleep, and it looks like I'm experiencing an obstructive apnea event.\\
View 3 & I'm in stage N2 sleep, and it looks like I had an obstructive apnea event.\\[4pt]
\textbf{Annotation} & \textbf{Stage N2 sleep.}\\
View 1 & Feeling peaceful, in stage N2 sleep now.\\
View 2 & I'm in stage N2 sleep, where dreams are short and muscles relax.\\
View 3 & I'm in stage N2 sleep, where dreams start to become more vivid but I'm still fairly rested.\\[4pt]
\textbf{Annotation} & \textbf{Stage N1 sleep. Hypopnea.}\\
View 1 & Just drifted into a light sleep with some shallow breathing.\\
View 2 & Just drifting in Stage N1 sleep, with a hypopnea event.\\
View 3 & Light dozing in stage N1 with a hint of shallow breathing.\\\midrule
\multicolumn{2}{@{}l}{\rule{0pt}{2.8ex}\normalsize\textbf{Wrist motion: activity annotations (Capture-24)}: 78{,}919 texts, 4 views each}\\[3pt]
\multicolumn{2}{@{}p{\linewidth}@{}}{\begin{tcolorbox}[colback=black!3,colframe=black!30,boxrule=0.4pt,arc=1pt,left=3pt,right=3pt,top=1pt,bottom=1pt,fontupper=\footnotesize]\textbf{Prompt} (\texttt{pvs-report}). This text describes a 30-second wrist accelerometer recording: "\textit{$\langle$text$\rangle$}" Write the short, terse report that an automatic analyzer of the 30-second wrist accelerometer recording would print for it. Reply with the report only.\end{tcolorbox}}\\
\textbf{Annotation} & \textbf{sitting; leisure; eating; eating sitting indoor/outdoor}\\
View 1 & - Activity: Sitting, Leisure, Eating - Location: Indoor/Outdoor - Duration: 30 seconds\\
View 2 & Timestamp: 0:00-30s Activity: Sitting (10s) + Leisure (5s) + Eating (10s) + Eating Sitting (5s) Indoor/Outdoor (variable)\\
View 3 & Sitting (leisure), eating (sitting, indoor/outdoor)\\[4pt]
\textbf{Annotation} & \textbf{sitting; home activity; miscellaneous; sitting; sitting desk entertainment/hobby (with or without eating at the same time)}\\
View 1 & 30-second wrist accelerometer recording: - 60\% Sitting - 40\% Sitting (Desk Entertainment/Hobby, possible eating)\\
View 2 & 30-second Wrist Activity Report: - 50\% Sitting - 50\% Desk Entertainment/Hobbies (with possible eating)\\
View 3 & 30-second Wrist Accelerometer Recording Report: - Primary Activity: Sitting - Secondary Activities: Home miscellaneous, Desk entertainment/hobby (possibly with eating)\\[4pt]
\textbf{Annotation} & \textbf{manual work; occupation; miscellaneous; manual labour}\\
View 1 & 30-second wrist accelerometer recording: Activity Type: Manual Work Occupation: Manual Labour Miscellaneous: Yes\\
View 2 & 30s Wrist Accelerometer Data Analysis Report - Activity: Manual Work - Context: Occupation - Classification: Miscellaneous Manual Labour\\
View 3 & 30-second wrist accelerometer recording analysis: Activity: Manual Work Category: Occupation Type: Miscellaneous Manual Labour\\\midrule
\multicolumn{2}{@{}l}{\rule{0pt}{2.8ex}\normalsize\textbf{Lung sounds: auscultation reports (SPRSound)}: 1{,}231 texts, 4 views each}\\[3pt]
\multicolumn{2}{@{}p{\linewidth}@{}}{\begin{tcolorbox}[colback=black!3,colframe=black!30,boxrule=0.4pt,arc=1pt,left=3pt,right=3pt,top=1pt,bottom=1pt,fontupper=\footnotesize]\textbf{Prompt} (\texttt{pvs-imagine}). Imagine the lung sound recording described by this text: "\textit{$\langle$text$\rangle$}" You are another observer perceiving the same lung sound recording. Describe it in one or two sentences, in your own words: what is there and how it is arranged. Do not hedge. Reply with the description only.\end{tcolorbox}}\\
\textbf{Report} & \textbf{Normal breath sounds.}\\
View 1 & The recording shows clear, even breath sounds without any adventitious sounds or abnormalities.\\
View 2 & The recording shows clear, even breath sounds with no abnormalities, distributed evenly throughout the lungs.\\
View 3 & The recording shows clear, even breath sounds with no additional noises or abnormalities.\\[4pt]
\textbf{Report} & \textbf{Normal breath sounds.}\\
View 1 & Clear, even breath sounds without any abnormalities.\\
View 2 & There are clear, even breath sounds evenly distributed throughout the lungs.\\
View 3 & There are even, clear sounds evenly distributed throughout the lungs, indicating normal breath sounds.\\[4pt]
\textbf{Report} & \textbf{Normal breath sounds.}\\
View 1 & There are even, clear breath sounds distributed evenly throughout the lungs.\\
View 2 & Clear, even breath sounds with no abnormality noted.\\
View 3 & The lung sounds are clear and even, with no abnormalities, indicating normal inspiration and expiration.\\\midrule
\multicolumn{2}{@{}l}{\rule{0pt}{2.8ex}\normalsize\textbf{Bird calls: species descriptions (iNatSounds)}: 27{,}601 texts, 4 views each}\\[3pt]
\multicolumn{2}{@{}p{\linewidth}@{}}{\begin{tcolorbox}[colback=black!3,colframe=black!30,boxrule=0.4pt,arc=1pt,left=3pt,right=3pt,top=1pt,bottom=1pt,fontupper=\footnotesize]\textbf{Prompt} (\texttt{pvs-imagine}). Imagine the animal sound recording described by this text: "\textit{$\langle$text$\rangle$}" You are another observer perceiving the same animal sound recording. Describe it in one or two sentences, in your own words: what is there and how it is arranged. Do not hedge. Reply with the description only.\end{tcolorbox}}\\
\textbf{Description} & \textbf{A bird calling: Blue Jay (Cyanocitta cristata, order Passeriformes, family Corvidae).}\\
View 1 & A Blue Jay is chirping loudly, with a distinctive loud echoic call that rises and falls in pitch.\\
View 2 & A Blue Jay is chirping loudly, with a distinctive loud piercing call followed by a series of sharper notes.\\
View 3 & A blue jay making its distinctive loud, squawking call.\\[4pt]
\textbf{Description} & \textbf{The sound of a Black-capped Chickadee (Poecile atricapillus).}\\
View 1 & A clear, high-pitched "chick-a-dee-dee-dee" sequence, starting with a lower "chick" note followed by rapid, repeated "dees.\\
View 2 & A clear, high-pitched "chick-a-dee-dee-dee" call, distinctive and memorable, echoing in the background.\\
View 3 & A clear, high-pitched "chick-a-dee-dee-dee" call, starting with a deep-throated syllable followed by a rapid series of notes.\\[4pt]
\textbf{Description} & \textbf{A bird calling: Mourning Dove (Zenaida macroura, order Columbiformes, family Columbidae).}\\
View 1 & A soft cooing sound, characteristic of a Mourning Dove.\\
View 2 & A mournful cooing sound from a mourning dove.\\
View 3 & A mourning dove is cooing softly, its distinctive repetitive coo coo sounding in the background.\\\midrule
\multicolumn{2}{@{}l}{\rule{0pt}{2.8ex}\normalsize\textbf{Insect and frog calls: species descriptions}: 2{,}867 texts, 4 views each}\\[3pt]
\multicolumn{2}{@{}p{\linewidth}@{}}{\begin{tcolorbox}[colback=black!3,colframe=black!30,boxrule=0.4pt,arc=1pt,left=3pt,right=3pt,top=1pt,bottom=1pt,fontupper=\footnotesize]\textbf{Prompt} (\texttt{pvs-literal}). Rewrite this description of a animal sound recording, keeping only what the animal sound recording itself would show and removing opinions, context and anything it would not show: "\textit{$\langle$text$\rangle$}" Reply with the rewrite only.\end{tcolorbox}}\\
\textbf{Description} & \textbf{A Dark Bush-cricket, Pholidoptera griseoaptera, calling: a katydid of the family Tettigoniidae.}\\
View 1 & A Dark Bush-cricket, Pholidoptera griseoaptera, calling.\\
View 2 & A Dark Bush-cricket, Pholidoptera griseoaptera, calling.\\
View 3 & Dark Bush-cricket, Pholidoptera griseoaptera, calling.\\[4pt]
\textbf{Description} & \textbf{The sound of an American Bullfrog (Lithobates catesbeianus).}\\
View 1 & The sound of an American Bullfrog (Lithobates catesbeianus).\\
View 2 & The sound of an American Bullfrog (Lithobates catesbeianus).\\
View 3 & The sound of an American Bullfrog (Lithobates catesbeianus).\\[4pt]
\textbf{Description} & \textbf{A Swamp Cicada, Neotibicen tibicen, calling: a cicada of the family Cicadidae.}\\
View 1 & A Swamp Cicada, Neotibicen tibicen, calling.\\
View 2 & A Swamp Cicada, Neotibicen tibicen, calling.\\
View 3 & A Swamp Cicada, Neotibicen tibicen, calling.\\
\end{longtable}}

\end{document}